\documentclass[runningheads]{llncs}

\usepackage[mobile]{eccv}

\usepackage{eccvabbrv}

\usepackage{booktabs}

\usepackage[accsupp]{axessibility}  

\usepackage[pagebackref,breaklinks,colorlinks,citecolor=eccvblue]{hyperref}

\usepackage{orcidlink}

\usepackage{array}
\usepackage{multirow}
\usepackage{rotating} 
\usepackage{graphicx}
\usepackage{amsmath,amssymb} 
\usepackage{color}
\usepackage{pifont}
\usepackage{subcaption}
\usepackage{comment}
\usepackage{tabularx}
\usepackage{xcolor}

\definecolor{darkgreen}{RGB}{0,100,0}
\definecolor{darkyellow}{RGB}{150,150,0}

\newcommand{\TODO}[1]{\textcolor{red}{#1}}

\newcommand{\ASE}{ASE}
\newcommand{\ADT}{ADT}
\newcommand{\AOThree}{AEO}
\newcommand{\AEO}{AEO}
\newcommand{\EFMmodel}{EVL}
\newcommand{\EVL}{EVL}
\newcommand{\OCCmodel}{OpenOcc}

\newcommand{\cmark}{\ding{51}}
\newcommand{\xmark}{\ding{55}}

\newcommand{\specificthanks}[1]{\@fnsymbol{#1}}

\begin{document}


\title{EFM3D: A Benchmark for Measuring Progress Towards 3D Egocentric Foundation Models} 

\titlerunning{EFM3D}
\authorrunning{J.~Straub~\etal}
\author{Julian Straub\thanks{Project lead.}, Daniel DeTone\thanks{Equal contribution in alphabetic order.}, Tianwei Shen$^{\star\star}$, Nan Yang$^{\star\star}$, \\
Chris Sweeney, Richard Newcombe}
\institute{Meta Reality Labs Research}

\maketitle

\begin{abstract}




The advent of wearable computers enables a new source of context for AI that is embedded in egocentric sensor data. This new egocentric data comes equipped with fine-grained 3D location information and thus presents the opportunity for a novel class of spatial foundation models that are rooted in 3D space. 
To measure progress on what we term Egocentric Foundation Models (EFMs) we establish EFM3D, a benchmark with two core 3D egocentric perception tasks. EFM3D is the first benchmark for 3D object detection and surface regression on high quality annotated egocentric data of Project Aria. We propose Egocentric Voxel Lifting (\EVL{}), a baseline for 3D EFMs. \EVL{} leverages all available egocentric modalities and inherits foundational capabilities from 2D foundation models. This model, trained on a large simulated dataset, outperforms existing methods on the EFM3D benchmark.



\end{abstract}

\begin{figure}
    \centering
    \vspace{-5mm}
    \includegraphics[width=\textwidth]{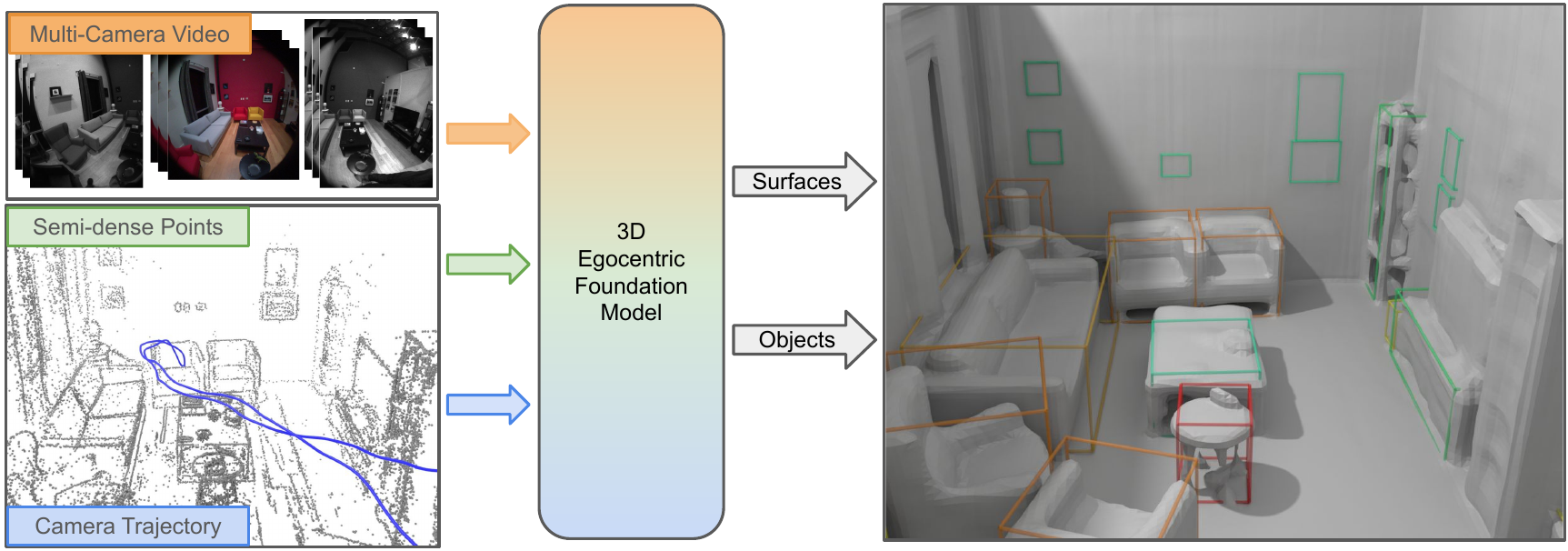}
    \caption{3D Egocentric Foundation Models leverage spatial priors from egocentric data to power core 3D tasks such as 3D object detection and reconstruction.}
    \label{fig:efm3d_teaser}
    \vspace{-5mm}
\end{figure}

\section{Introduction}
Foundation models trained on Internet-scale text, image and video datasets have demonstrated the potential in using large-scale self-supervised learning approaches to build backbones that are useful for numerous downstream tasks, through both fine-tuning and zero-shot learning. 
The advent of wearable spatial computers enables a new source of context from egocentric sensor data. Key to unlocking this context is understanding the environment of the wearer. 
This new egocentric data source comes equipped with fine-grained 3D location information~\cite{engel2023project,grauman2023egoexo} and thus presents the opportunity for a novel class of spatial foundation models that are rooted in 3D space.
This class of 3D egocentric foundation models (EFMs) can leverage strong priors from egocentric data like camera poses, calibration and semi-dense point information. 
%
As with 2D image foundation models~\cite{he2022masked,oquab2023dinov2,assran2023ijepa,kirillov2023segany}, the availability of large amounts of data is critical for training such models and high quality annotations to measuring their performance. While there are now significant amounts of 2D data with 2D annotations~\cite{deng2009imagenet,lin2014coco,gupta2019lvis}, and a large body of 3D scene dataset~\cite{dai2017scannet,yeshwanthliu2023scannetpp,replica19arxiv,dehghan2021arkitscenes}, and autonomous vehicles (AV) datasets~\cite{caesar2020nuscenes,Geiger2012CVPR}, the equivalent for egocentric data from wearable devices with 3D annotations is only just starting to become available~\cite{engel2023project,grauman2023egoexo,ase2024web}.  
%
%
%

To enable measuring progress towards EFMs we propose the EFM3D benchmark which contains two tasks: 3D bounding box detection and surface regression.
%
%
To set up the first baseline model on EFM3D, we introduce the Egocentric Voxel Lifting (\EFMmodel{}) model which relies on frozen 2D foundation features to set a competitive performance. \EFMmodel{} leverages all egocentric modalities from Aria including posed and calibrated RGB and greyscale video streams and semidense points. 
When trained on our new large-scale simulated dataset \EFMmodel{} generalizes well to the real-world EFM3D benchmark and significantly outperforms the current state-of-the-art 3D scene understanding models even when they are retrained on the same dataset. To summarize our contributions:

\noindent \textbf{Dataset.} We release more annotations including 3D object bounding boxes (OBBs) and groundtruth (GT) meshes built on top of Aria Synthetic Environments dataset~\cite{ase2024web} and real Project~Aria sequences to enable research on the two foundational tasks of 3D object detection and surface reconstruction. 

\noindent \textbf{Benchmark.} We set up the EFM3D benchmark with the first two tasks, namely 3D object detection and surface reconstruction, to advance continuous research in the areas of egocentric machine perception.

\noindent \textbf{Method.} We introduce a baseline model, \EFMmodel{}, to solve both tasks at the state-of-the-art level by leveraging explicit volumetric representation, a full suite of egocentric signals, and 2D features from vision foundation models.



\section{Related Work}
We categorize the related work into datasets, 3D foundation models, object detection, and surface reconstruction methods using video and point cloud inputs. 
\begin{table}[]
    \centering
    \resizebox{\textwidth}{!}{%
    \begin{tabular}{c|c|c|c|c|c|c|c|c|c}
                                                  & sim/real & motion       & video  & points& poses  & \# scenes & \# sequences & \#3D OBBs   & surface  \\ \hline \hline
        Ego4D~\cite{grauman2022ego4d}             &real      & egocentric & \cmark &       &        & 74        & 3k hrs       & \xmark      &    \xmark   \\
        Ego-Exo4D~\cite{grauman2023egoexo}        &real      & egocentric & \cmark &\cmark & \cmark &   131     & 5,625        &  \xmark     &     \xmark \\
        AEA~\cite{lv2024aria}                     &real      & egocentric & \cmark &\cmark & \cmark & 5         & 143          & \xmark      &  \xmark \\   
        ScanNet~\cite{dai2017scannet}             &real      & scanning   & \cmark &       & \cmark & 707       & 1,513        & 36k         &  fused depth \\
        ARKITScenes~\cite{dehghan2021arkitscenes} &real      & scanning   & \cmark &       & \cmark & 1,661     & 5,048        & 52k        &  fused depth \\
        Hypersim~\cite{roberts2021hypersim}       &sim       & random     & \cmark &       & \cmark & 461       & 774          & 54k        & CAD mesh \\ \hline
        \ASE{}~\cite{ase2024web}                  &sim       & egocentric & \cmark &\cmark &\cmark  & 100k      & 100k         & 3M          & CAD mesh\\
        \ADT{}~\cite{Pan_2023_ICCV}               &real      & egocentric &\cmark &\cmark &\cmark   &  1        & 6            & 281         & CAD mesh \\
        \AOThree{}                                &real      & egocentric &\cmark &\cmark &\cmark   &  20       & 26           & 584        & \xmark \\
    \end{tabular}
    }
    \caption{Related datasets either are egocentric or have 3D OBB and surface annotations. We fill this gap and contribute 3D OBB annotations for~\ASE{}, CAD meshes for a validation subset of~\ASE{} and~\ADT{} as well as \AOThree{}---an egocentric validation dataset with high quality 3D OBB annotations.}
    \label{tab:datasets}
    \vspace{-5mm}
\end{table}

\noindent\textbf{Datasets}. 
As shown in Table~\ref{tab:datasets}, egocentric datasets~\cite{grauman2023egoexo,grauman2022ego4d,lee2012discovering,Damen2018EPICKITCHENS,northcutt2020egocom} have been typically designed to enable activity recognition, video narration and 2D understanding tasks. The most recent egocentric datasets, through the use of Project Aria~\cite{engel2023project}, are starting to enable 3D perception by releasing accurate camera calibrations, pose and semi-dense points information. Examples are Ego-Exo4D~\cite{grauman2023egoexo}, which also contains localized third-person perspective information, the Aria Everyday Activities (AEA)~\cite{lv2024aria} dataset as well as the Aria Digital Twin (ADT)~\cite{Pan_2023_ICCV} dataset. Only the ADT dataset contains 3D OBBs and CAD mesh GT annotations for real-world Project Aria data. However all data is recorded in a single space, making diversity too low for OBB tasks.
Thus far 3D object detection and surface reconstruction datasets used by the community are either real RGB-D scanning sequences of indoor spaces (e.g. ScanNet~\cite{dai2017scannet,Avetisyan_2019_CVPR}, ScanNet++~\cite{yeshwanthliu2023scannetpp}, and  ARKitScenes~\cite{dehghan2021arkitscenes}) or simulations (e.g. Hypersim~\cite{roberts2021hypersim}, InterioNet~\cite{InteriorNet18}, and 3D-Front~\cite{fu20213dfront}).   
%
%
%
While these RGB-D datasets have sufficient scale for training, there is substantial modality difference from RGB-D datasets to egocentric data from Project Aria. 
Project Aria provides one high-resolution RGB and two grey-scale video streams whereas RGB-D datasets come with just a single RGB stream. RGB-D datasets contain dense, uniformly sampled depth and surface information whereas egocentric data from Project Aria comes only with sparse depth via semi-dense point clouds from the SLAM system.
Additionally, the motion in scanning datasets is different than egocentric motion because the aim is to ``cover'' and observe all surfaces of the scene, which is not the case in typical egocentric data.
We find these differences in modalities and motion patterns are challenging and necessitate substantial differences in model design which opens up new research.
To close these gaps we provide new object and mesh annotations for existing Project Aria datasets~\cite{ase2024web,Pan_2023_ICCV} and a small high-quality object detection benchmark.

\noindent\textbf{3D Foundation Models}. Likely hindered by the availability of sufficient 3D training data, there are few related 3D foundation models and none designed for egocentric data. 3D-LFM~\cite{dabhi202333dlfm} generates 3D skeletons from 2D point inputs for more than 30 categories of 2D-3D point sets. 
Most relevant is Ponder~\cite{huang2023ponder,zhu2023ponderv2} which uses a point-cloud encoder on unprojected RGB-D frames from ScanNet to produce a dense feature volume that is trained via a NERF~\cite{mildenhall2021nerf} rendering loss. The architecture shares the lifting into a feature volume~\cite{harley2019learning} but does so via a sparse point encoder instead of leveraging 2D foundation models and dense lifting. As we find in our experiments 3D point-encoder-based methods like 3DETR~\cite{misra2021end} struggle with the non-uniform egocentric point clouds.

\noindent\textbf{3D Object Detection}. 
The most similar method to the proposed \EFMmodel{} model is ImVoxelNet~\cite{rukhovich2022imvoxelnet}. After unprojecting 2D image features into a 3D feature volume, a 3D CNN head regresses 3D bounding box parameters at multi-scales. Similarly, Raytran~\cite{tyszkiewicz2022raytran} lifts 2D features into a 3D grid via a sparse attention mechanism to detect 3D OBBs. Neither use additional semi-dense point information.
In the AV community 3D OBB detection is also investigated based on video and LiDAR points~\cite{yin2021center,focalsconv-chen}. LiDAR points sparsely but uniformly provide depth input in contrast to egocentric semidense points. The OBB distribution is also more constrained due to the fact that the car and objects move essentially in the same, locally-2D plane.
DETR3D~\cite{wang2022detr3d} and extensions such as PARQ~\cite{Xie_2023_ICCV} explore how to directly perform 3D OBB detection from multi-view image input. 
%
Cube R-CNN~\cite{Brazil_2023_CVPR} is following the Mask R-CNN-paradigm and directly regresses 3D OBBs form single frames. Similarly, MOLTR~\cite{li2021moltr} and ODAM~\cite{li2021odam} detect 3D OBBs per frame and additionally learn to track and fuse 3D OBBs.
Point-cloud based methods for 3D object detection such as VoteNet~\cite{qi2019votenet} and more recent Transformer-based 3DETR~\cite{misra2021end}, typically leverage either dense depth or fused RGB-D reconstructions as inputs. We find that the sparse, non-uniform nature of egocentric point clouds is challenging for this class of methods.
%


\noindent\textbf{3D Surface Estimation} Aside from the vast amount of traditional stereo methods~\cite{furukawa2009accurate,goesele2007multi}, learning-based surface reconstruction can be categorized into volumetric methods, depth-based methods, and scene-specific methods~\cite{mildenhall2021nerf,kerbl3Dgaussians}. 
Among depth-based approaches, the premise of monocular depth estimation is to learn strong priors about scenes and environments to regress depth without any multi-view information. ZoeDepth~\cite{bhat2023zoedepth} is a high-performing monocular depth estimator that is first trained on relative depth and then fine-tuned on datasets with metric depth. Typically mono-depth methods have problems with the scale ambiguity of natural images. 
ConsistentDepth~\cite{khan2023tcod} addresses this issue by leveraging the global sparse points to achieve temporally consistent depths. With multi-view inputs, MVSNet~\cite{yao2018mvsnet} builds a 3D cost volume via differentiable homography warping to regress depths. At the boundary between multi-view methods and mono-depth methods that infer depth from multiple views is SimpleRecon~\cite{sayed2022simplerecon}, which constructs a frustum cost volume but then squashes it into a 2D feature map for depth regression.
ATLAS~\cite{murez2020atlas} proposes the idea of unprojecting 2D image features into a 3D feature grid for semantic surface regression. and Lift-Splat-Shoot~\cite{philion2020lift} in the AV literature concurrently uses a similar idea to lift to a bird-eye-view (BEV) grid. 
NeuralRecon~\cite{sun2021neucon} expands on this idea by first regressing local sparse TSDF and then fusing local reconstructions into a global volume using a learned fusion module. As such it is the most similar method to the proposed \EFMmodel{} model.
%
None of these methods is designed for egocentric data modalities nor has been evaluated on egocentric data.

\section{Dataset Contributions}
\label{sec:dataset}


To facilitate novel research in 3D Egocentric Foundation Models, we release new data for both training and evaluation, spanning both synthetic and real egocentric Aria data. We summarize the three types of this new data below.

\noindent\textbf{Large-scale Synthetic OBB Metadata}. We release the 3D oriented bounding boxes (OBBs) corresponding to the \ASE{} dataset. This consists of approximately 3 million OBBs across 43 object classes. This includes visibility metadata for \ASE{} which enables image-based detector training and evaluation. We define the visibility of an OBB on a certain image based on its \textit{observability} and \textit{occlusion}.

\noindent\textbf{Small-scale Real world OBB Dataset}. We introduce a small real-world egocentric validation dataset, termed Aria Everyday Objects (\AOThree{}). It has high quality 3D OBB annotations to allow evaluation on real Project Aria sequences. 
One key aspect of this dataset is that the sequences were collected by non computer vision experts, without any specific guidance to scan the scene. This ensures realistic egocentric motion. 
%
%
The annotations are made by human annotators that use the semi-dense point depths, camera poses, calibrated multi-camera rig to label the 3D OBBs in a 3D viewer. 
%
%
The dataset contains 26 diverse scenes with 584 OBB instances across 9 classes: Chair, Table, Sofa, Bed, WallArt, Lamp, Plant, Window and Mirror. These semantic classes overlap with the ASE dataset and enable testing of sim-to-real generalization.

\noindent\textbf{Ground-truth Meshes}. For 3D reconstruction training the \ASE{} dataset already contains groundtruth depth images. We release the 3D groundtruth mesh for the simulated \ASE{} validation dataset as well as for the real-world \ADT{} dataset to enable benchmarking 3D reconstruction methods.



\section{Egocentric Voxel Lifting (\EFMmodel{})}

\begin{figure}
    \centering
    \includegraphics[width=\textwidth]{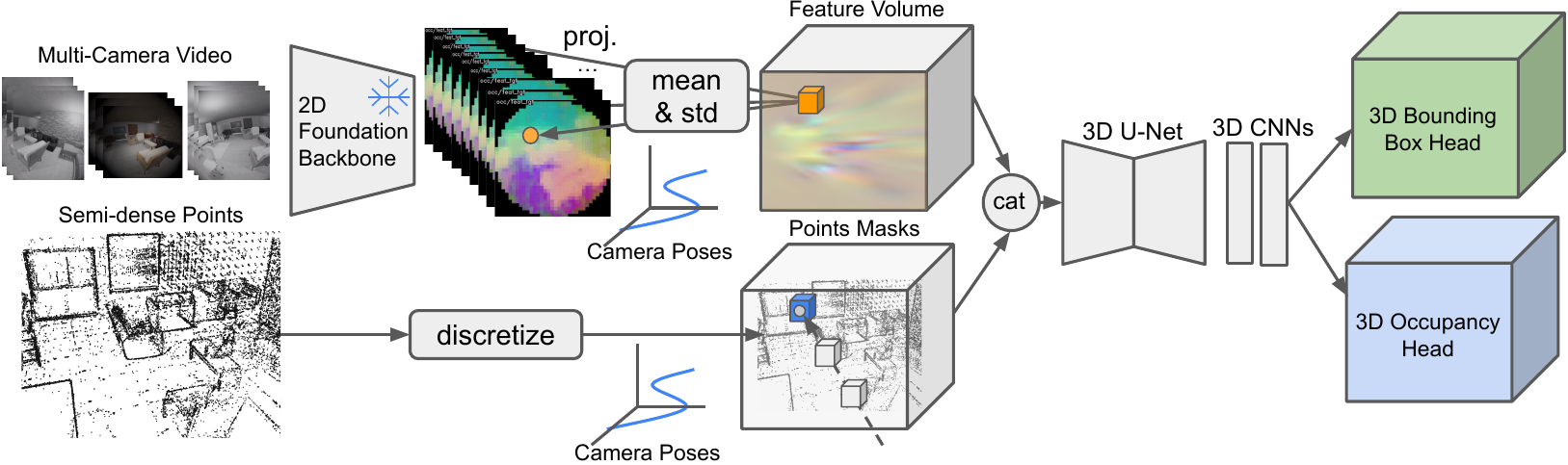}
    \caption{\EFMmodel{} lifts 2D features extracted from frozen image foundation models into an local gravity-aligned 3D voxel grid of features. After concatenating point masks a U-Net processes the 3D feature volume before running 3D CNN heads. These heads regress target parameters such as 3D OBBs and occupancy values.
    \label{fig:efm_model}}
\end{figure}

We design \EFMmodel{} as a universal 3D backbone that lifts 2D foundation features from posed and calibrated video streams into a 3D gravity-aligned voxel grid of features. Before processing this 3D feature volume with a 3D U-Net we concatenate masks derived from semi-dense points that indicate surface points as well as freespace. As shown in Fig.~\ref{fig:efm_model}, on this feature volume we can run different kinds of 3D-CNN heads to regress desired 3D quantities such as 3D OBB parameters and occupancy values.

\noindent\textbf{2D-3D Lifting Encoder} We run a 2D foundation model~\cite{oquab2023dinov2} with frozen weights on $T$ posed input images from each video stream. The resulting 2D foundation features are up-sampled to the full input image resolution using stacked 2D CNN \& upsample layers.  
We instantiate a local, gravity-aligned 3D voxel grid in front of the most recent Aria camera rig. We choose the last RGB camera pose and gravity align its rotation to obtain the anchor pose of the voxel grid.
%
Using known camera poses and calibrations, we project the voxel grid centers into each image and sample the 2D feature maps using bi-linear interpolation while respecting the valid radius of the distorted camera models.
This yields a feature volume of $T\times F \times D\times H\times W$ for every stream. 
We aggregate features across all streams and the time dimension using mean and standard deviation to obtain a lifted feature volume of shape $2F \times D\times H\times W$.
The metric size and resolution of the voxel grid can be defined freely. 
Note that this lifting formulation is camera model agnostic and supports online-calibrated video streams from Aria. Please refer to the supplemental for the experiments of \EFMmodel{} using different camera models.

\noindent\textbf{Semi-Dense Point Encoder}. We additionally input the semi-dense 3D point cloud provided by the Aria research toolkit~\cite{engel2023project} to provide a geometric prior for the lifting backbone. Together with the visibility information from where each point was observed this point cloud contains not only information about the surfaces in the scene but also the freespace. To provide this information to \EFMmodel{} we compute a point and a free-space mask tensor to concatenate to the lifted feature volume. 
The point mask indicates where there are surface points. It is computed by discretizing points into a binary mask sized $D\times H\times W$ corresponding to the voxel grid resolution.
The freespace mask indicates the voxel locations between the camera centers to the known surface points.
We compute it by sampling $S$ points along each ray within the bounds of the feature volume. 
We concatenate both masks to the lifted 3D features resulting in a 3D feature volume shaped $(F+2)\times D\times H\times W$.

\noindent\textbf{3D U-Net}. Once the 3D feature volume is instantiated, we run a 3D U-Net to allow 3D processing. The U-Net reduces the spatial resolution by a factor of 8x before upsampling back to the full 3D resolution.
See a more detailed model diagram in the supplemental.



\subsection{3D Bounding Box Detection}
We assume all object bounding boxes are gravity-aligned. We take inspiration from the ImVoxelNet~\cite{rukhovich2022imvoxelnet} to build a proposal-free and anchor-free 3D detection head. 
This head runs on top of the output of \EFMmodel{} and regresses the class distribution ($v^{cls}$) and eight values defining the geometric properties of a 3D bounding box. 
It first predicts a centerness score of $[0,1]$ for each voxel $v$. The centerness ($v^c$) of each voxel defines the probability of containing any object bounding box center.
Then the head regresses the seven parameters ($v^{obb}$) that describe the bounding box: three for the height, width and depth of the box, three that predict a small offset for the bounding box center to account for discretization of the voxel grid, and lastly a single yaw rotation parameter that determines the rotation of the bounding box around the gravity direction.
The predictions are then filtered with Non-Maximum Suppression (NMS) using 3D Intersection over Union (IoU) and the centerness confidence score. There are three losses used for training, one for each head. The centerness and the classifier heads are trained with focal loss (FL)~\cite{lin2017focal}, and the parameters head is trained using gravity oriented 3D IoU loss. With $N_v$ being the number of voxels:
\begin{align}\textstyle
    L_{obj} = \frac{1}{N_v}\sum^{N_v}_{n} & w_{c}\text{FL}(v_{n}^{c}, \widehat{v_{n}^{c}}) + w_{iou}\text{IoU}(v_{n}^{obb}, \widehat{v_{n}^{obb}}) + w_{cls}\text{FL}(v_{n}^{cls}, \widehat{v_{n}^{cls}}),
\end{align}

\subsection{3D Surface Regression}
To regress 3D surfaces in the local gravity-aligned voxel grid we regress an occupancy value for every voxel. 
We supervise occupancy using the GT depth maps. For each GT depth value we obtain three kinds of occupancy supervisions: one surface point, one free-space point, and one occupied point. We compute the surface point $p_\text{surf}$ by unprojecting the GT depth values using the camera calibrations. The free-space point $p_\text{free}$ is sampled in front of the surface, while the occupied point $p_\text{occ}$ is sampled behind the surface. Both $p_\text{free}$ and $p_\text{occ}$ are randomly sampled up to $\delta$ distance from the surface, which we choose as the length of a voxel.
At these locations we sample into the predicted occupancy grid via tri-linear interpolation to obtain predicted values for all three kinds of points. The final loss is then computed via the focal loss~\cite{lin2017focal} by defining the ground-truth probability of free-space point, surface point, and occupied point being $0.0$, $0.5$, and $1.0$, respectively:
\begin{align}
   L_{surf} = \textstyle\frac{1}{N}\sum^{N}_{n} \text{FL}(p_\text{free}^n, 0.0) + \text{FL}(p_\text{surf}^n, 0.5) + \text{FL}(p_\text{occ}^n, 1.0),
\end{align}
where $\text{FL}(\cdot)$ is the focal loss and $N$ is the number of unprojected points from the GT depth maps. Note that we use focal loss to make the optimization to focus less on the easy samples like flat surfaces while more on the hard samples like sharp object boundaries. We further add a total variation (TV) regularization loss on the predicted occupancy volume to encourage local smoothness. 

\subsection{Implementation and Training Details}

\noindent\textbf{Input Snippet Data Preparation}. We sample 1s snippets from the RGB and greyscale video streams of Aria running at 10Hz.
The input resolution for RGB and greyscale videos are $240\times 240\times3$ and $320\times240\times1$, which has approximately the same angular resolution per pixel.
We leave exploring other image sampling strategies like key-framing~\cite{Xie_2023_ICCV} for future research.
For the GT OBB labels, we use the OBBs which are visible inside the snippet and discretize the centers into the volume to obtain the centerness supervision signal. We further filter out the OBBs which are heavily occluded in the snippet as decribed in Sec.~\ref{sec:dataset}.

\noindent\textbf{2D Foundation Model Encoder}. While we could use other 2D foundation models like CLIP~\cite{radford2021learning} to encode the video streams, we find the base DinoV2.5~\cite{oquab2023dinov2,darcet2023vitneedreg} model works well for our tasks and is reasonably efficient. 

\noindent\textbf{Feature Volume} We choose a $4m^3$ voxel grid. We adapt the grid resolution to the task. For 3D object detection we find a $6.25cm$ voxel size ($D=H=W=64$) is sufficient and for surface regression we push to $4cm$ ($D=H=W=96$) to be comparable with related work~\cite{sun2021neucon}. 

\noindent\textbf{Hyperparameters}. The 3D Bounding Box Detection head is trained on ASE 10k sequences (with 526k snippets) for $10$ epochs and the 3D Surface Regression head is trained on the ASE 1k sequences (with 54k snippets). Both heads were trained with a base learning rate of $2e^{-4}$ using Adam~\cite{kingma2014adam} optimizer with cosine annealing scheduler. We set $w_{cent}=100$, $w_{iou} = 10$, and $w_{class} = 1.0$ for training 3D bounding box detection. For the inference, we use the centerness threshold of $0.2$ to select the candidate detections and performs simple 3D NMS with radius of $2$ voxels based on the 3D IoU. The focal loss paramters in the 3D Surface training is set to $\alpha=0.25$ and $\gamma=2.0$.


\section{The EFM3D Benchmark and Experiments}
In this section we describe the EFM3D benchmark, which consists of two egocentric tasks: 3D oriented bounding box detection and 3D surface regression. For the object detection task, we use the Aria Everyday Objects (\AEO{}) dataset. For the surface estimation task, we use the Aria Digital Twin (\ADT{}) dataset.
To demonstrate the value of this benchmark, we evaluate the proposed \EFMmodel{} model together with state-of-the-art-approaches on the 3D OBB detection and the 3D surface reconstruction tasks.
These tasks on egocentric data can leverage the video streams, poses and camera calibration as well as semi-dense point outputs extracted from the video streams via a SLAM~\cite{engel2023project} system. For both tasks we train on the large-scale simulated \ASE{} dataset and evaluate performance both in domain on \ASE{} as well as on real sequences with ground-truth annotations in \ADT{} and \AOThree{} recorded with Project Aria.

\begin{figure}[htbp]
    \centering
    \begin{subfigure}[b]{0.22\textwidth}
        \centering
        \includegraphics[width=\textwidth]{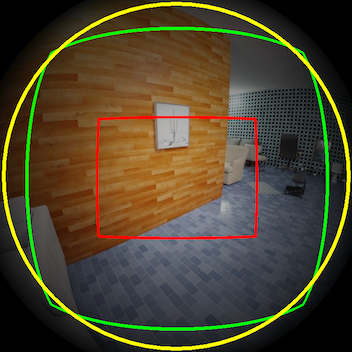}
        \caption{\tiny \textbf{ASE FoV}}
    \end{subfigure}
    \hfill
    \begin{subfigure}[b]{0.22\textwidth}
        \centering
        \includegraphics[width=\textwidth]{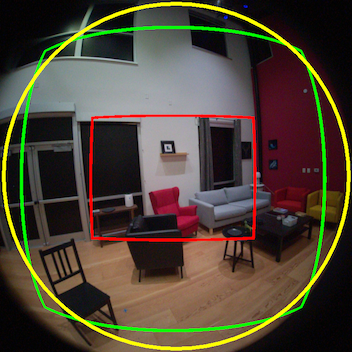}
        \caption{\tiny \textbf{ADT FoV}}
    \end{subfigure}
    \hfill
    \begin{subfigure}[b]{0.22 \textwidth}
        \centering
        \includegraphics[width=\textwidth]{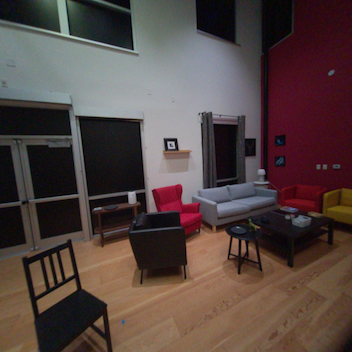}
        \caption{\tiny \textbf{Max-linear}}
    \end{subfigure}
    \hfill
    \begin{subfigure}[b]{0.292 \textwidth}
        \centering
        \includegraphics[width=\textwidth]{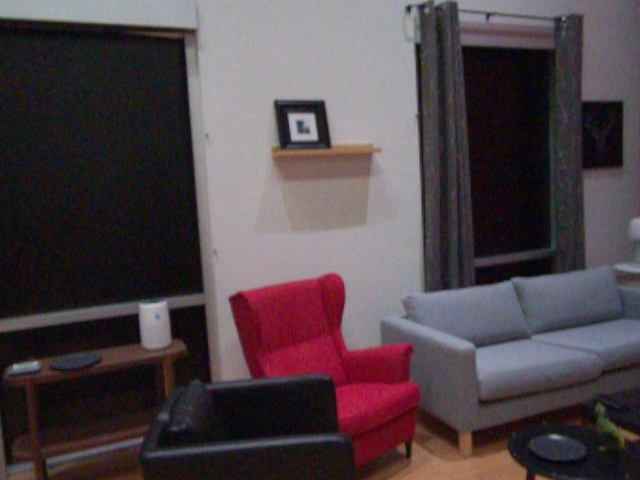}
        \caption{\tiny \textbf{ScanNet linear}}
    \end{subfigure}
    \caption{We visualize (c) maximum linear (region in {\color{darkgreen}green}) and (d) ScanNet linear (region in {\color{red}red}), as well as the original fisheye (with a valid region in {\color{darkyellow}yellow}) camera models for (a) ASE and (d) ADT.}
    \label{fig:camera_lin}
\end{figure}
\noindent \textbf{Field-of-view matters.} While \EFMmodel{} uses a non-linear camera model to leverage the full field of view (FoV) of the Aria cameras, various related methods~\cite{sun2021neucon,sayed2022simplerecon,rukhovich2022imvoxelnet,Brazil_2023_CVPR} rely on linear cameras. For the off-the-shelf (OTS) models trained on ScanNet, we undistort the images into the ScanNet intrinsics in order to provide the data in the format the models were trained with. For other OTS models~\cite{bhat2023zoedepth,khan2023tcod} that are not constrained by camera models, we provide the largest possible linear camera model (max-linear). As can be seen in Fig.~\ref{fig:camera_lin}, there is a substantial different in FoV between the two linear models. The max-linear model covers almost the same FoV as the distorted camera. As we show in the following experiments, larger FoV provides better coverage of the scene thus better performance.

\noindent\textbf{Persisting 3D Predictions}.
Inferring the 3D scene state from individual frame or snippet-level 3D inferences is not only useful for downstream systems but also enables comparing different algorithms and measuring how consistent these 3D inferences are across the full sequence. 
In the following experiments, we use tracking and fusion systems for both 3D OBBs as well as surface predictions to enable accumulation and improvement of 3D OBBs and 3D surfaces reconstructions. Predicted 3D OBBs are associated to the scene OBBs using a set of metrics including 3D intersection over union (IoU) and then fused via a running average over OBB parameters. For 3D reconstruction we use the implementation from~\cite{zeng20163dmatch} with minor improvements to fuse depth maps into an truncated signed distance field (TSDF)~\cite{newcombe2011kinectfusion}. The occupancy volume predicted by~\EFMmodel{} is fused using a per voxel running average in a global grid. The meshes are extracted using the marching cubes algorithm~\cite{lorensen1998marching} at the iso-level of $0$ for TSDF fusion and $0.5$ for occupancy fusion.
See supplemental for more details.

\subsection{3D Bounding Box Detection and Persistence}

Broadly speaking there are three different types of ML architectures that can run on these modalities: 2D CNNs running on individual frames~\cite{Brazil_2023_CVPR}, 3D CNNs running on 3D geometric quantities~\cite{rukhovich2022imvoxelnet} and Transformers that compute on the point cloud~\cite{misra2021end} or snippets~\cite{Xie_2023_ICCV}.
For Cube R-CNN~\cite{Brazil_2023_CVPR}, we rectify the images using the max-linear camera setting.
For ImVoxelNet~\cite{rukhovich2022imvoxelnet} we show performance of the base configuration following the training recipe for OBB training. We provide max-linear rectified RGB snippets. 
We freeze the ResNet50 image encoder to help sim-to-real transfer.
%
3DETR~\cite{misra2021end} is trained on the per snippet visible point cloud using the author's OBB training recipe and the random point-cloud crop augmentation from the paper.
%
%
All models are trained on 10k sequences of \ASE{} containing 600k snippets for 10 epochs except for the transformer-based models which are trained for 50 epochs, as we found its convergence to be slow. We report the mean Average Precision (mAP) averaged over IoU thresholds in the range $[0.0, 0.05, ..., 0.5]$ and semantic classes following Cube R-CNN~\cite{Brazil_2023_CVPR}. 

\begin{table}[]
    \centering
    \begin{tabular}{c|c|c|c| c|c|c|c|}
           & Train & Modality & Decoder & \ASE{} mAP & \ASE{} mAP & \AOThree{} mAP \\ 
           & Set   &          &         & Snippet  & Sequence  &  Sequence  \\ \hline    
        Cube R-CNN   &  OTS & frame & 2D CNN & 0.01 & 0.02 & 0.05 \\ \hline
        Cube R-CNN   &  ASE & frame & 2D CNN& 0.21 & 0.36 & 0.08 \\ 
        ImVoxelNet & ASE & snippet& 3D CNN  & 0.30 & 0.64 & 0.15 \\ 
        3DETR       &  ASE   & pts & Transformer & 0.24 & 0.33 & 0.16 \\
        \EFMmodel{} (ours)  &    ASE        & snip$+$pts & 3D CNN & \textbf{0.40} & \textbf{0.75} & \textbf{0.22} \\  
    \end{tabular}
    \caption{The OTS Cube R-CNN model does not generalize well to egocentric data. 
    Four models trained on \ASE{} training data are evaluated on the simulated \ASE{} validation dataset as well as real-world \AOThree{} dataset. }
    \label{tab:obj_det_perf}
    
\end{table}

We compare a representative set of related approaches that cover different subsets of modalities and architecture types in Tab.~\ref{tab:obj_det_perf} quantitatively.
Across all methods the OBB tracker increases the mAP---in most cases by almost a factor of $2\times$---when comparing the snippet to the sequence-level mAP on \ASE{}.  For the sequence results we perform a search over tracker instantiation thresholds for each model and report the best mAP.
%
Interestingly, when comparing performance on the challenging \AEO{} real-world dataset, the image based models suffer the strongest from the sim-to-real gap: Cube R-CNN (-32 mAP), ImVoxelNet (-49 mAP) and \EFMmodel{} (-48 mAP) . The points only method 3DETR has a smaller drop in mAP (-17 mAP). We attribute this to the fact that point-clouds are lower dimensional data than video and are less sensitive to photorealistic rendering artifacts in synthetic renderers. \EFMmodel{} performs best in both synthetic and real data scenarios.  We note that although the Cube R-CNN OTS (off-the-shelf) model is trained a diverse dataset it does not generalize well to egocentric data.

\begin{figure}[t!]
\centering
\begin{tabular}{cccc}
 & {\tiny GT} & {\tiny \EFMmodel{}} & {\tiny ImVoxelNet~\cite{rukhovich2022imvoxelnet}} \\
\begin{sideways}{\tiny ASE 100576}\end{sideways} & 
\includegraphics[width=0.3\textwidth]{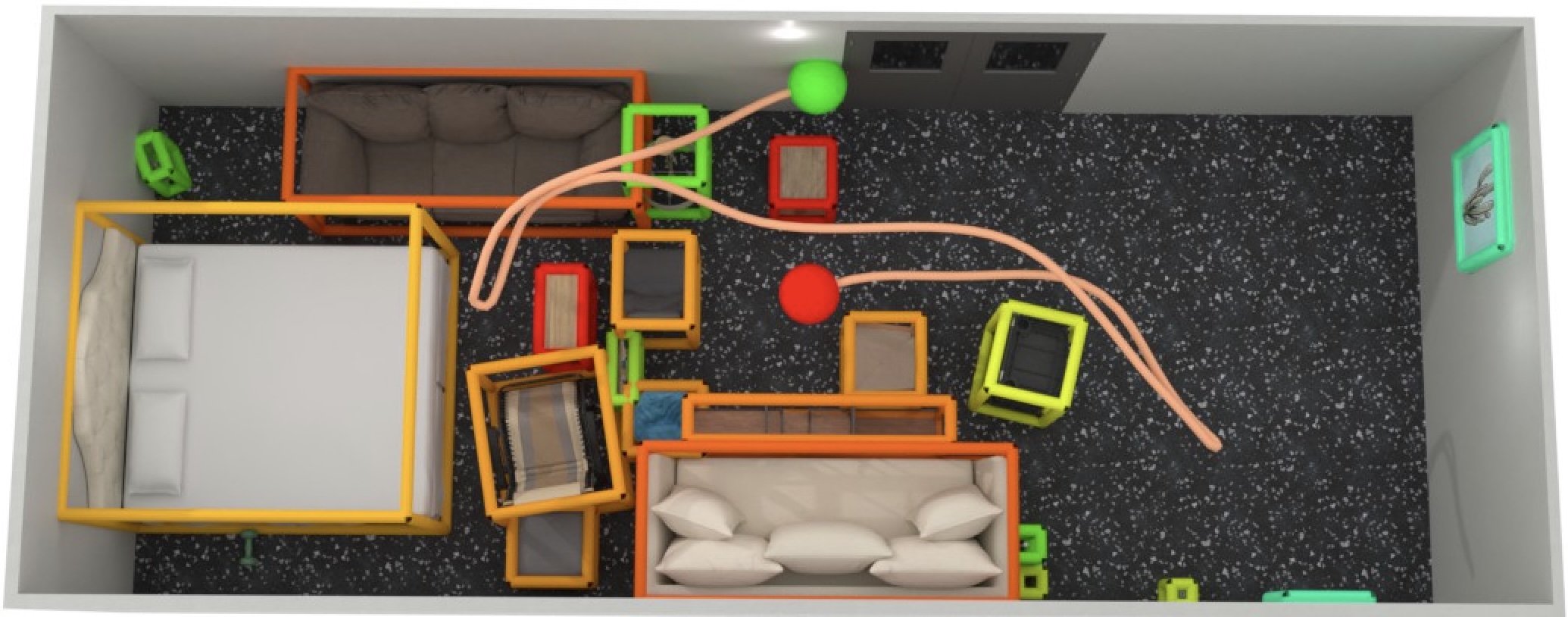}\hfill &
\includegraphics[width=0.3\textwidth]{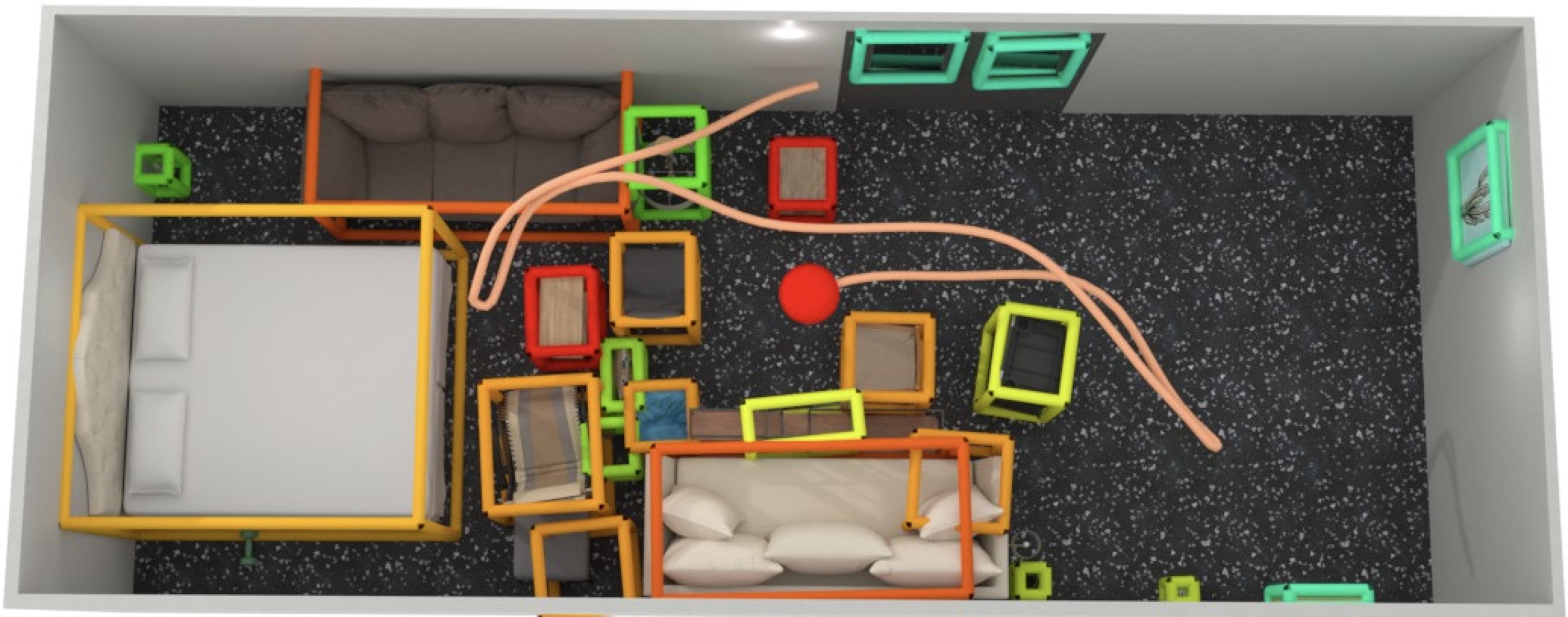}\hfill &
\includegraphics[width=0.3\textwidth]{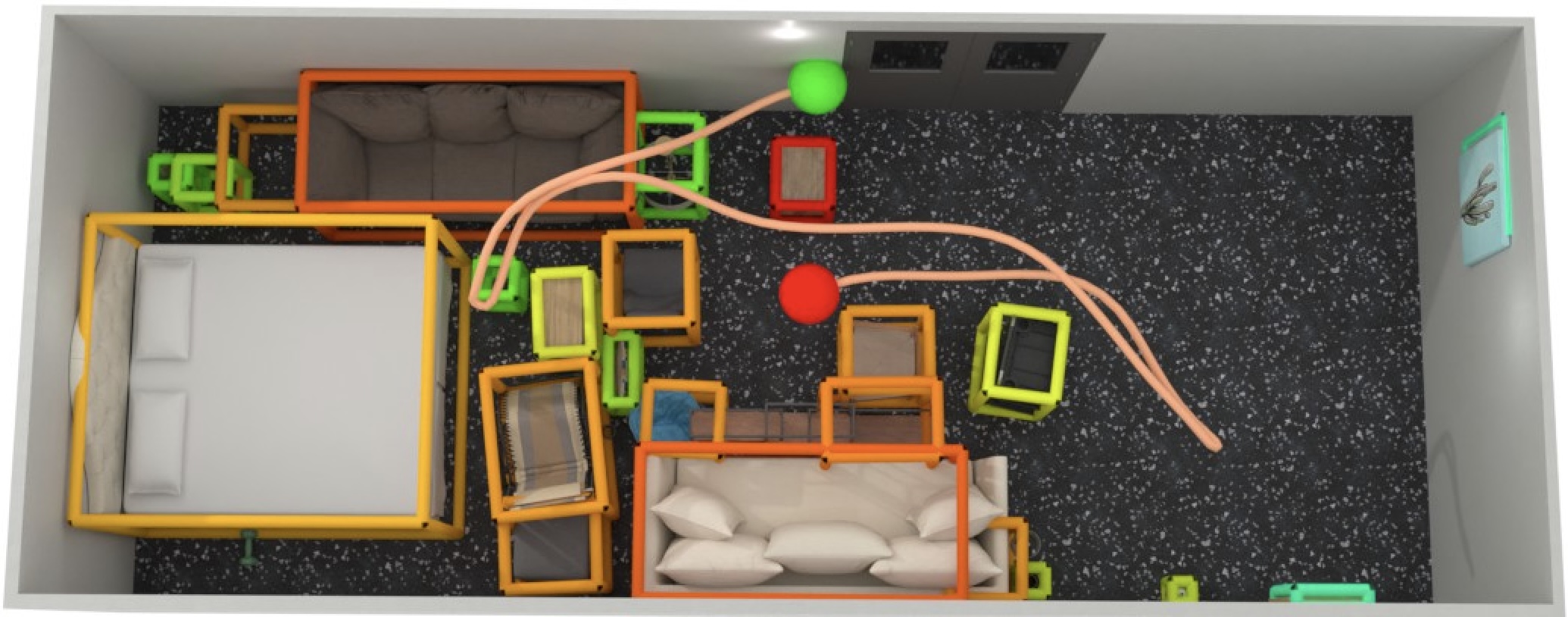}\hfill \\
\begin{sideways}{\tiny ASE 100303}\end{sideways} & 
\includegraphics[width=0.3\textwidth]{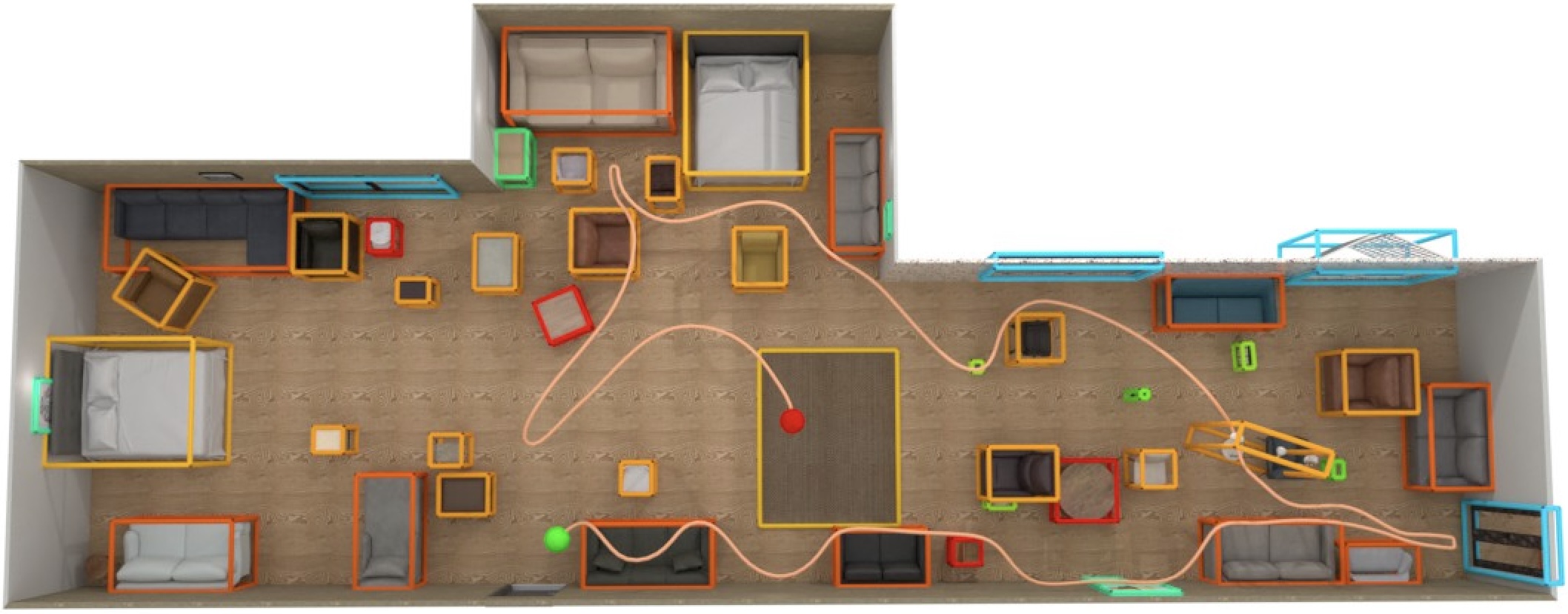}\hfill &
\includegraphics[width=0.3\textwidth]{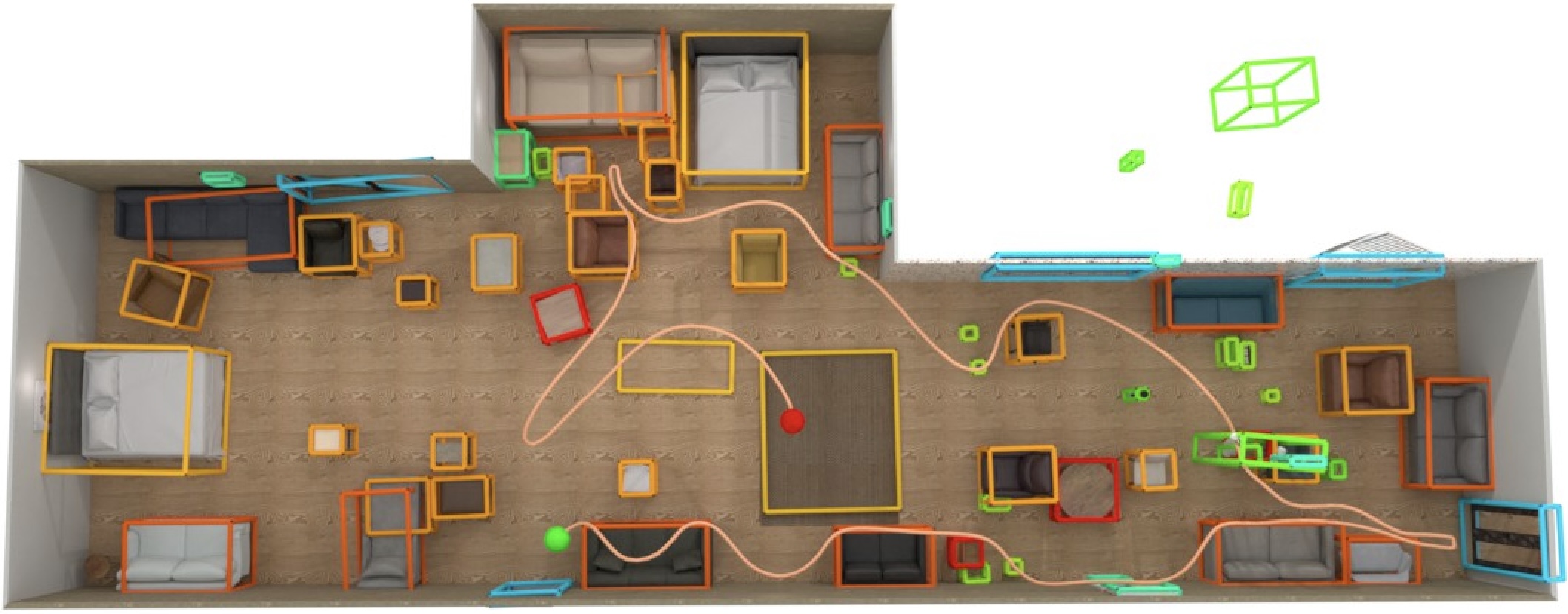}\hfill &
\includegraphics[width=0.3\textwidth]{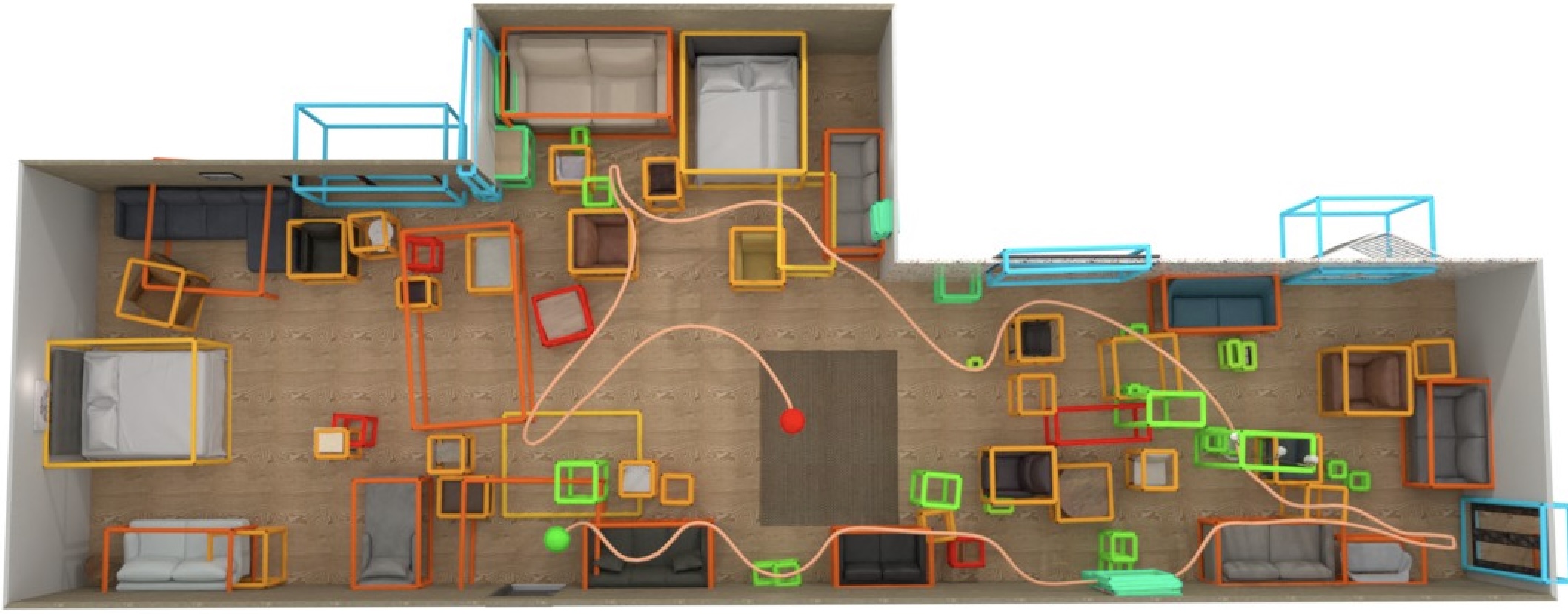}\hfill \\
\begin{sideways}{\tiny \ \ \ \ \ \ \ \ \  ASE 11132}\end{sideways} & 
\includegraphics[width=0.3\textwidth]{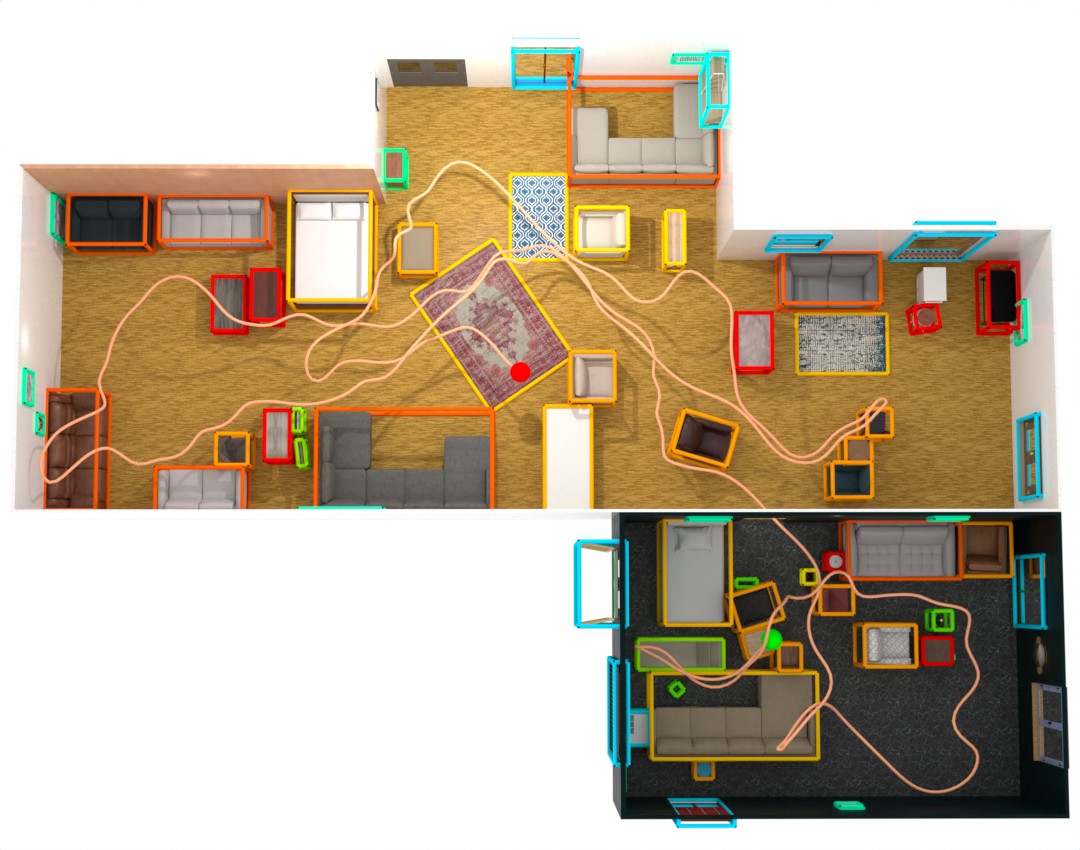}\hfill &
\includegraphics[width=0.3\textwidth]{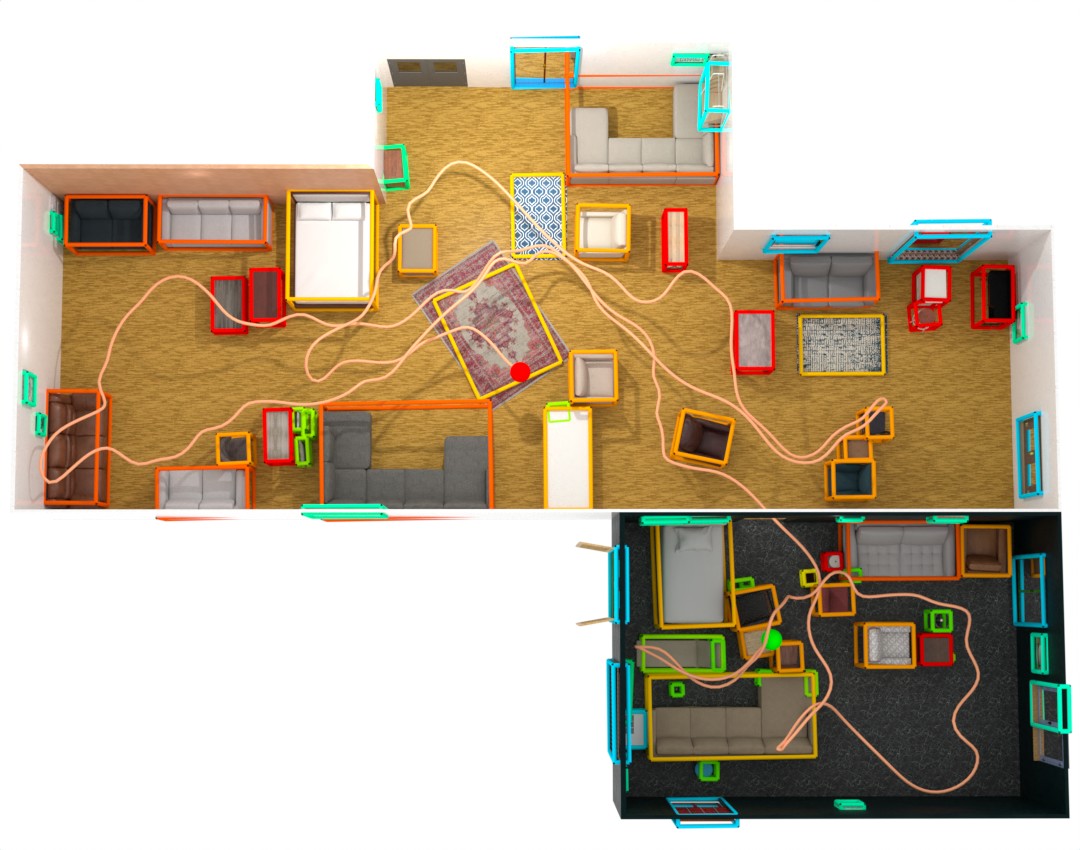}\hfill &
\includegraphics[width=0.3\textwidth]{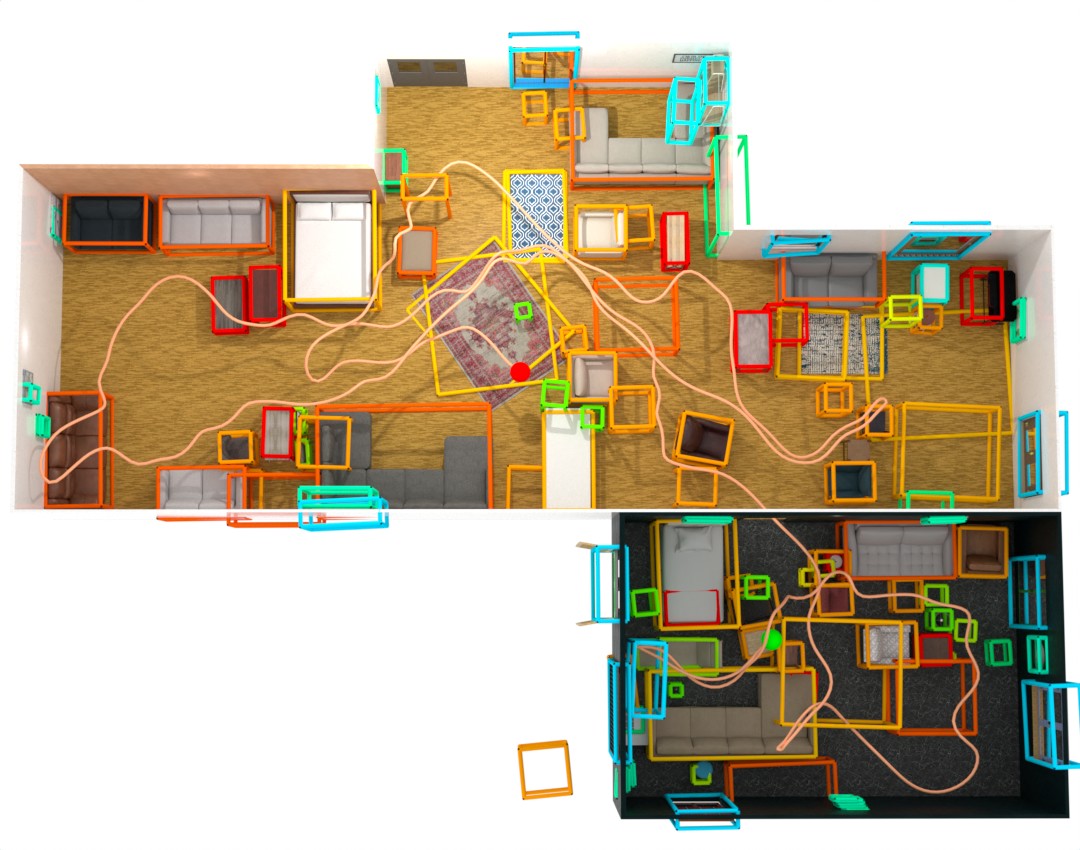}\hfill \\
\end{tabular}
\caption{\ASE{} validation scenes overlaid with camera trajectories and the 3D OBB predictions from \EFMmodel{} and ImVoxelNet~\cite{rukhovich2022imvoxelnet}.
\label{fig:obj_det_samples_ase}}
\end{figure}
We show overhead views of the sequence level OBB prediction results on held out synthetic scenes for the GT, \EFMmodel{} and ImVoxelNet in Fig.~\ref{fig:obj_det_samples_ase}. Both approaches perform relatively well on the synthetic data. ImVoxelNet has more false positives and the \EFMmodel{} has tighter bounding boxes. We attribute this gap mostly to the additional pointcloud input that the \EFMmodel{} has that provides a strong geometric cue for localizing the OBBs in 3D space.


\begin{figure}[t!]
\centering
\begin{tabular}{cccc}
 & {\tiny GT} & {\tiny \EFMmodel{}} & {\tiny ImVoxelNet} \\
\begin{sideways}{\tiny \ \ \ \ \ \ \ \ \ AEO Scene \#14}\end{sideways} & 
\includegraphics[width=0.3\textwidth]{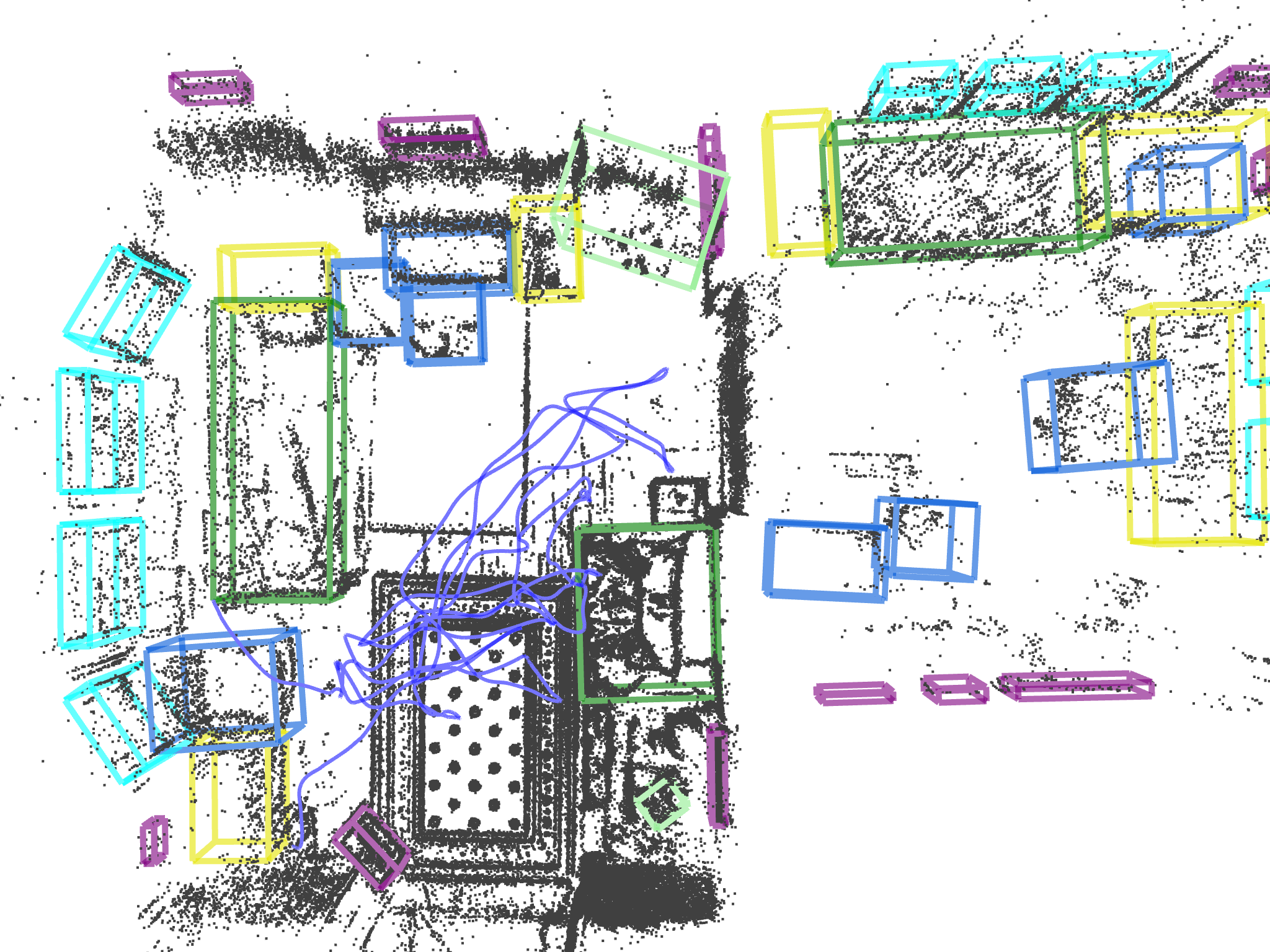}\hfill &
\includegraphics[width=0.3\textwidth]{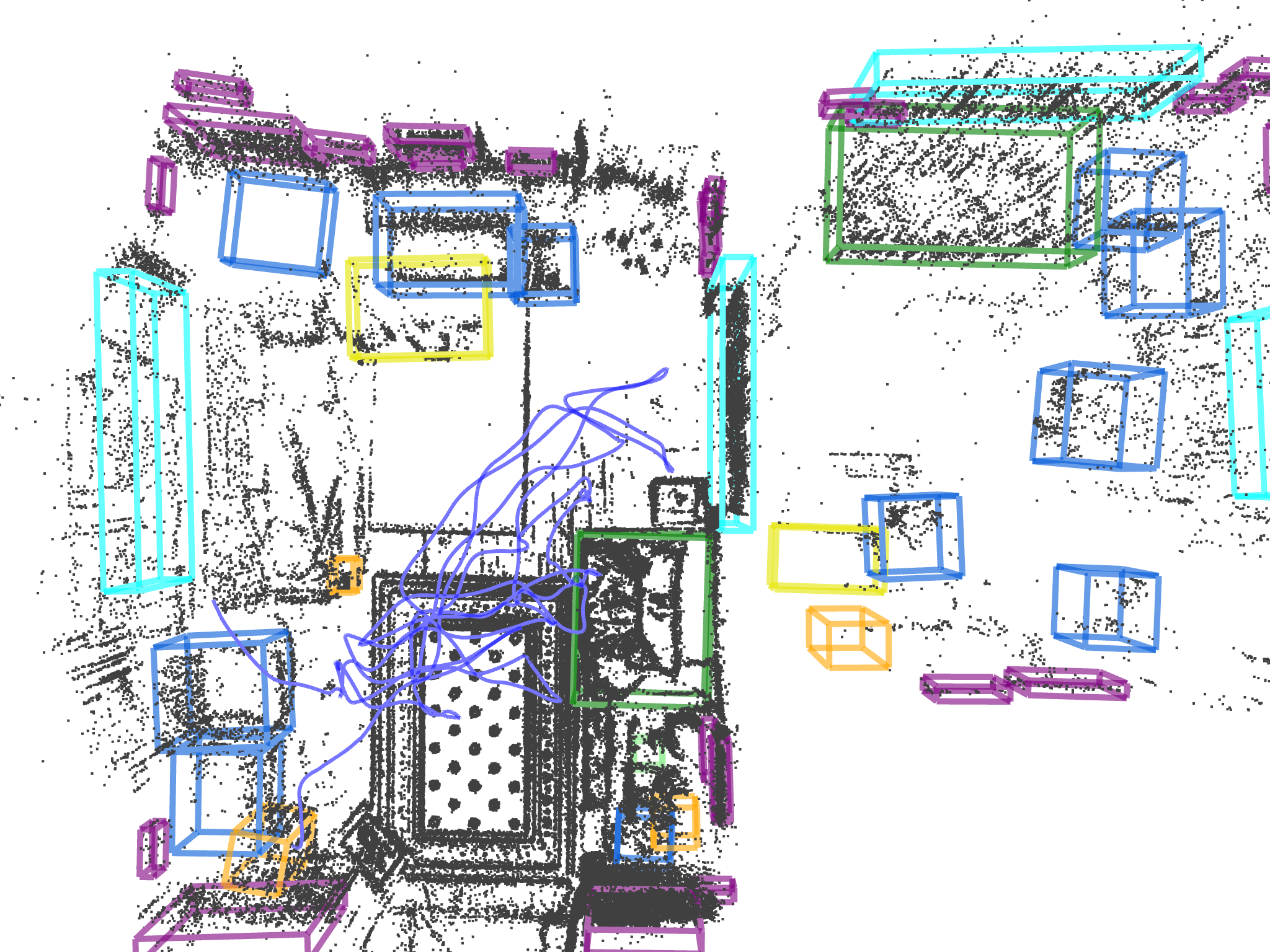}\hfill &
\includegraphics[width=0.3\textwidth]{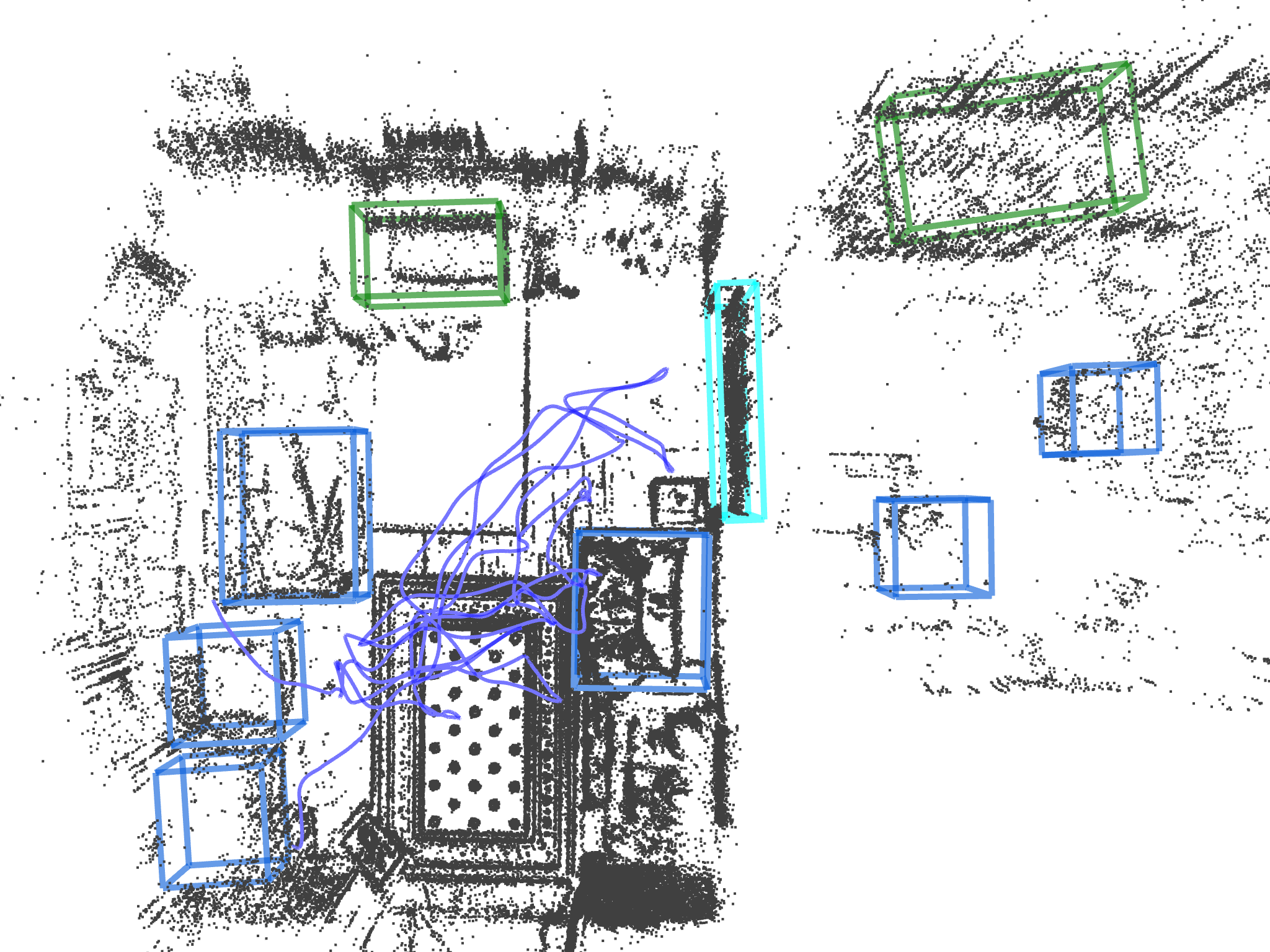}\hfill \\
\begin{sideways}{\tiny \ \ \ \ \ \ \ \ \ AEO Scene \#24}\end{sideways} & 
\includegraphics[width=0.3\textwidth]{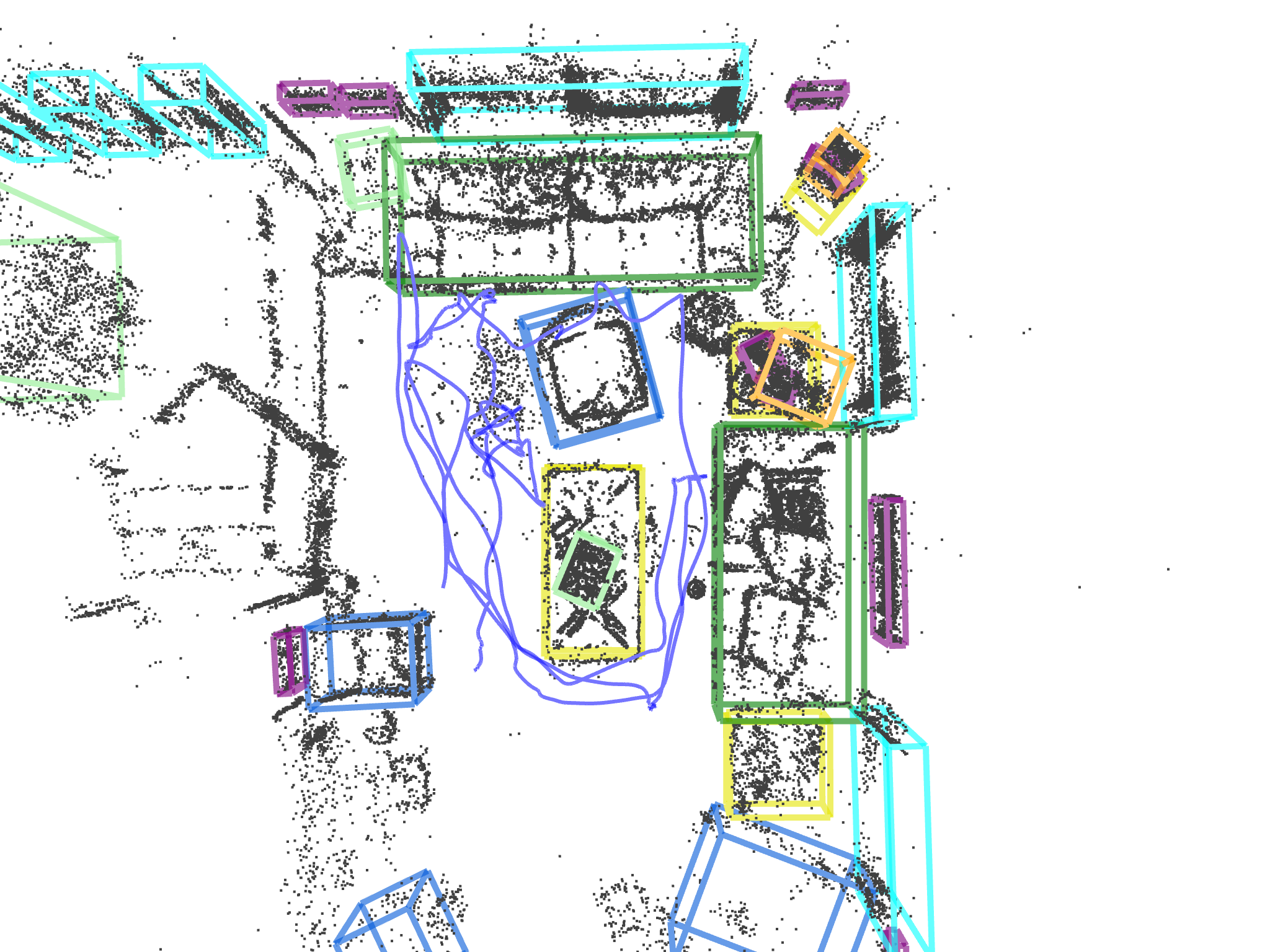}\hfill &
\includegraphics[width=0.3\textwidth]{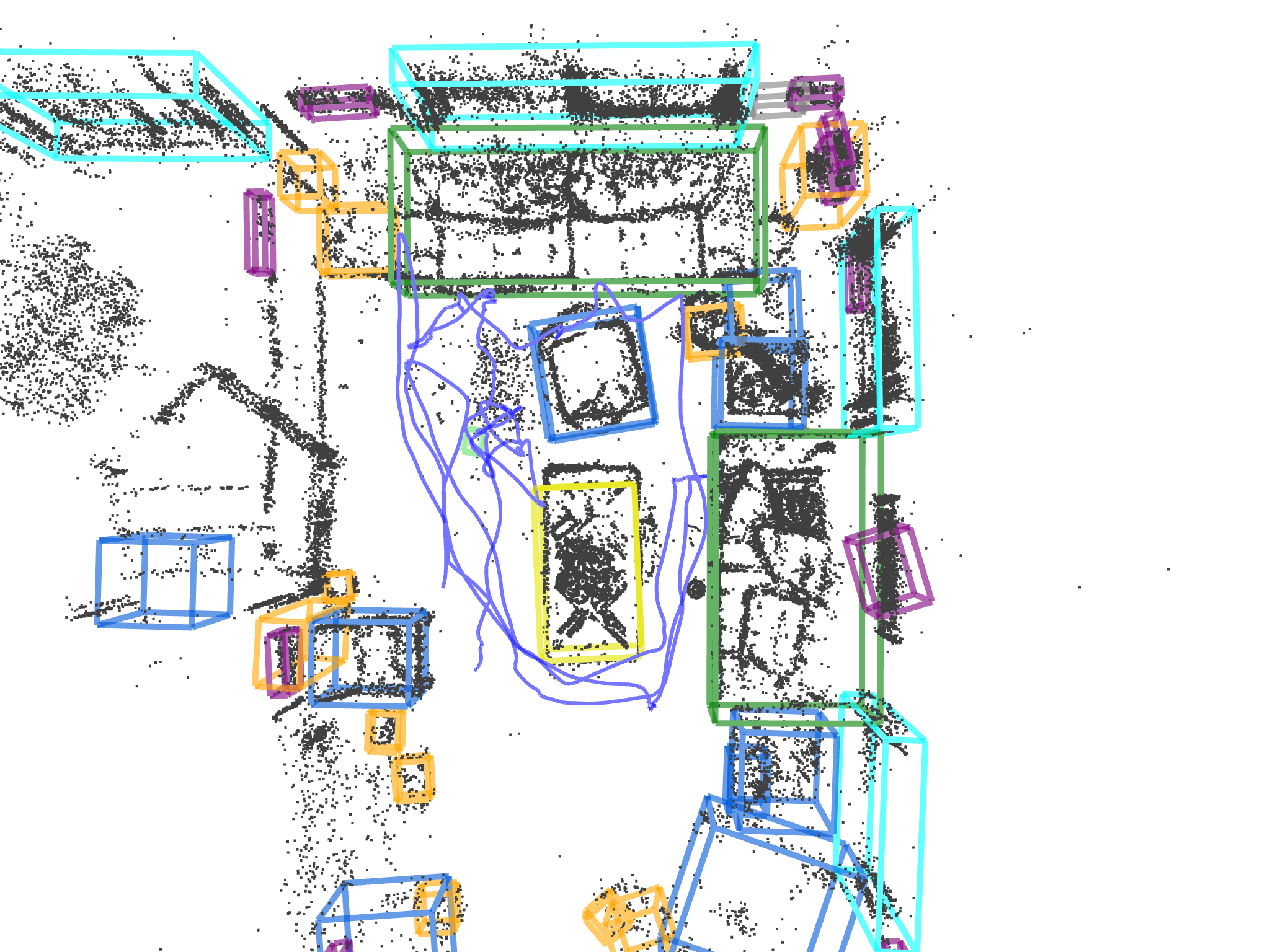}\hfill &
\includegraphics[width=0.3\textwidth]{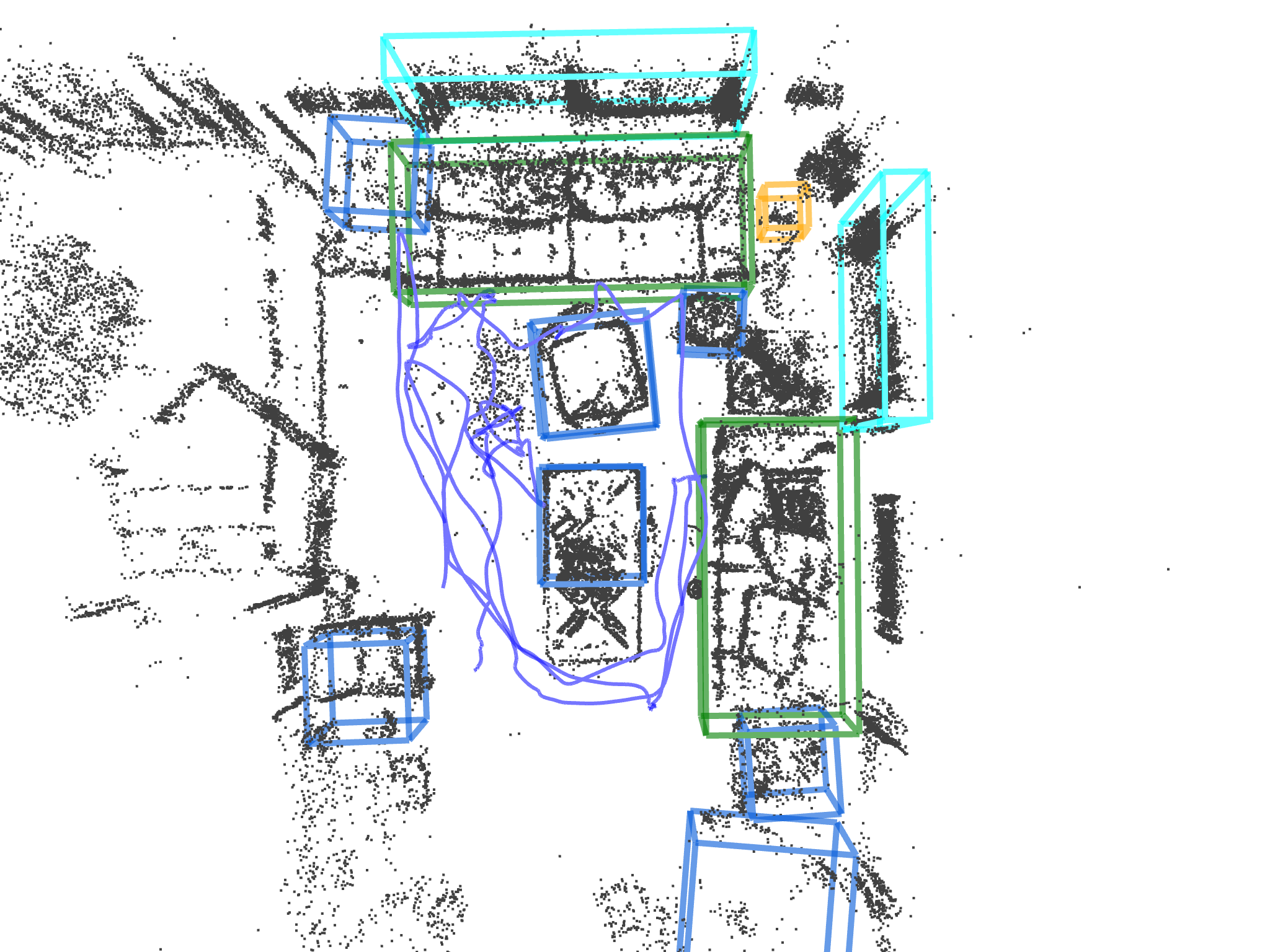}\hfill \\
\end{tabular}
\caption{Aria Everyday Object (\AEO{}) scenes overlaid with the camera trajectories (blue) and the colored 3D bounding boxes from Ground Truth (left), \EFMmodel{} (middle), and ImVoxelNet (right).
\label{fig:obj_det_samples_aeo}}
\end{figure}

In Fig.~\ref{fig:obj_det_samples_aeo} we show a similar comparison but on real data on the \AOThree{} dataset. Textured meshes are not available for this real data so we show the OBBs overlaid on top of the semi-dense point clouds. In this visualization the sim-to-real gap is more apparent. For example, there are never any bay windows in the \ASE{} dataset, but in the top row (AEO Scene $\#14$) shown in cyan there are four bay windows. \EFMmodel{} does a reasonable job to approximate them with a single window, whereas ImVoxelNet fails to detect any window . Another example of the sim-to-real gap is in the bottom-most row (AEO Scene $\#24$) there is a plant shown in a light green OBB on top of the table shown in a yellow OBB. The \ASE{} dataset does not have plants on top of other objects, so it is reasonable that both models fail to detect the plant on top of the table.

\begin{table}[]
    \centering
    \resizebox{\textwidth}{!}{%
    \subfloat[][\textbf{Augmentation:} photometric and geometric augmentations help generalization.]{
    \begin{tabular}{c|c|c|c|c|}
        geom.      & photo      & \ASE{} mAP & \ASE{} mAP  \\ 
        aug.       &  aug.      & Snippet  & Sequence    \\ \hline           
                   &            & 0.26 & 0.52  \\
        \cmark     &            & 0.38 & 0.67  \\
        \cmark     &    \cmark  & 0.38 & 0.68  \\
    \end{tabular}
    }\quad 
    \subfloat[][\textbf{Aggregation:} feature aggregation using mean and standard deviation is best.]{
    \begin{tabular}{c|c|c|c|c|c|}
         mean  & std    & \ASE{} mAP & \ASE{} mAP  \\ 
               &           & Snippet  & Sequence    \\ \hline           
        \cmark &           & 0.26 & 0.52  \\
               & \cmark    & 0.37 & 0.66  \\
        \cmark & \cmark    & 0.39 & 0.71  \\
    \end{tabular}
    }\quad 
    \subfloat[][\textbf{Usage of Points:} concatenating a points and freespace mask helps generalization.]{
    \begin{tabular}{c|c|c|c|c|c|}
         pts   & free    & \ASE{} mAP & \ASE{} mAP  \\ 
               &           & Snippet  & Sequence   \\ \hline   
        \cmark &           & 0.38 & 0.72  \\
               & \cmark    & 0.36 & 0.69  \\
        \cmark & \cmark    & 0.39 & 0.71  \\
    \end{tabular}
    }
    }
    \caption{We modify different hyperparameter settings based on the RGB stream \& points \EFMmodel{}. Turning on all augmentations, aggregating features in time using standard deviation (std) and mean, and concatenating both a points and freespace mask leads to the best model.}
    \vspace{-5mm}
    \label{tab:obb_ablation}
\end{table}

We perform a sensitivity analysis of \EFMmodel{} in Table~\ref{tab:obb_ablation}. Geometric augmentation at training time is more important than photometric augmentation. We find that standard deviation aggregation is more important than mean aggregation, but using both is best. Removing the freespace encoding has a smaller effect than remove the point mask, but using both is best.





\subsection{3D Surface Estimation and Reconstruction}

We evaluate the performance of the proposed \EFMmodel{} with occupancy output in comparison to a set of most relevant baseline methods on the final fused reconstructions. The representative methods are from different surface regression categories: monocular depth~\cite{bhat2023zoedepth} estimation, multi-view depth~\cite{sayed2022simplerecon,khan2023tcod} estimation, volumetric surface reconstruction~\cite{sun2021neucon}. For all depth-based methods, we adjust the depth values by aligning the depth output to the semidense points. We use the learned fusion module for NeuralRecon~\cite{sun2021neucon} instead of the aforementioned fusion system to generate the final outputs. Since NeuralRecon is the most similar method to \EFMmodel{}, we also re-train it on the same ASE training dataset. For other methods, we use off-the-shelf (OTS) models.
In addition, we also evaluate the GT depths and semi-dense points that are provided with the datasets.
%
%

\newcommand{\subfigwidth}{0.24}
\begin{figure}[htbp]
    \centering
    \begin{subfigure}[b]{\subfigwidth\textwidth}
        \centering
        \caption{\tiny \textbf{GT Mesh}}
        \includegraphics[width=\textwidth]{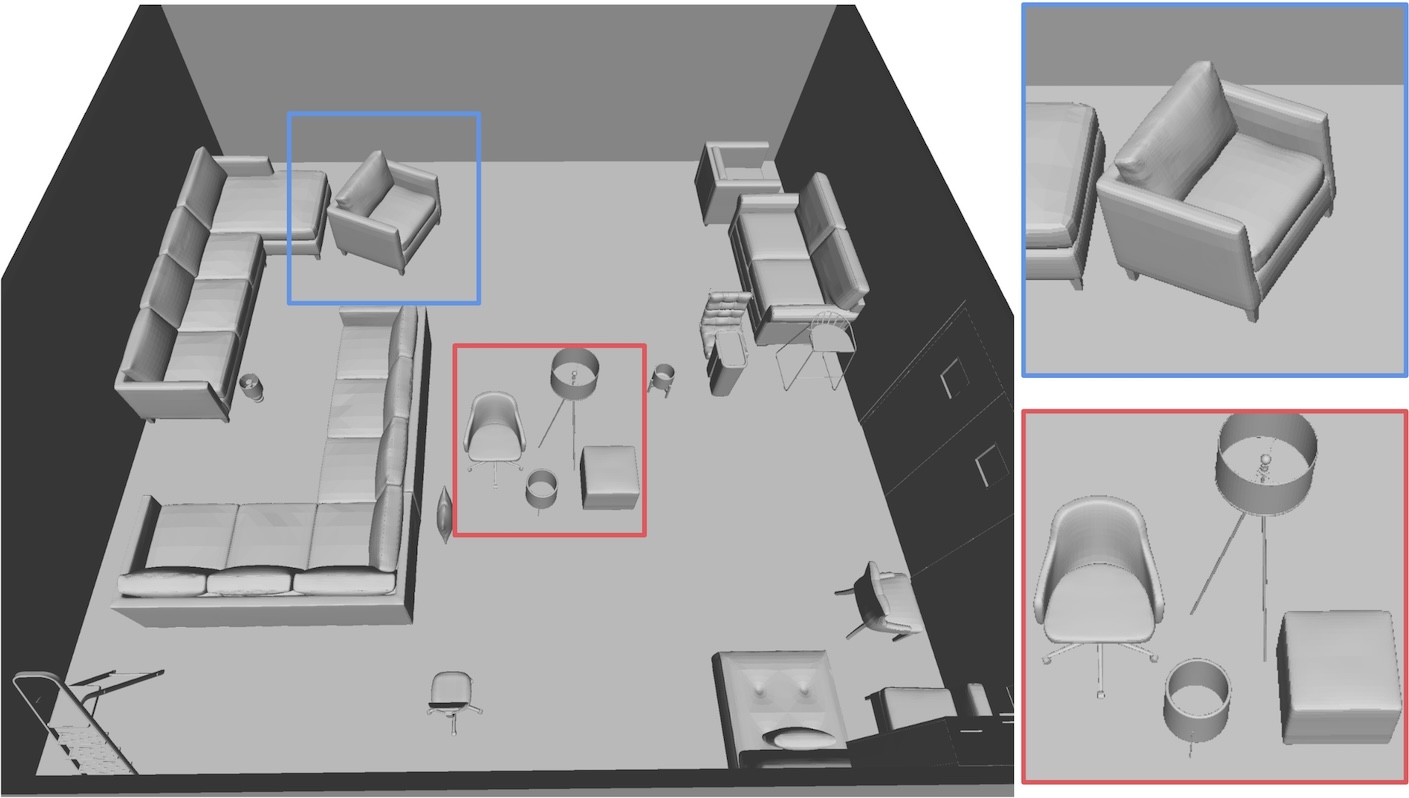}
    \end{subfigure}
    \hfill
    \begin{subfigure}[b]{\subfigwidth \textwidth}
        \centering
        \caption{\tiny \textbf{GT Depth Fusion}}
        \includegraphics[width=\textwidth]{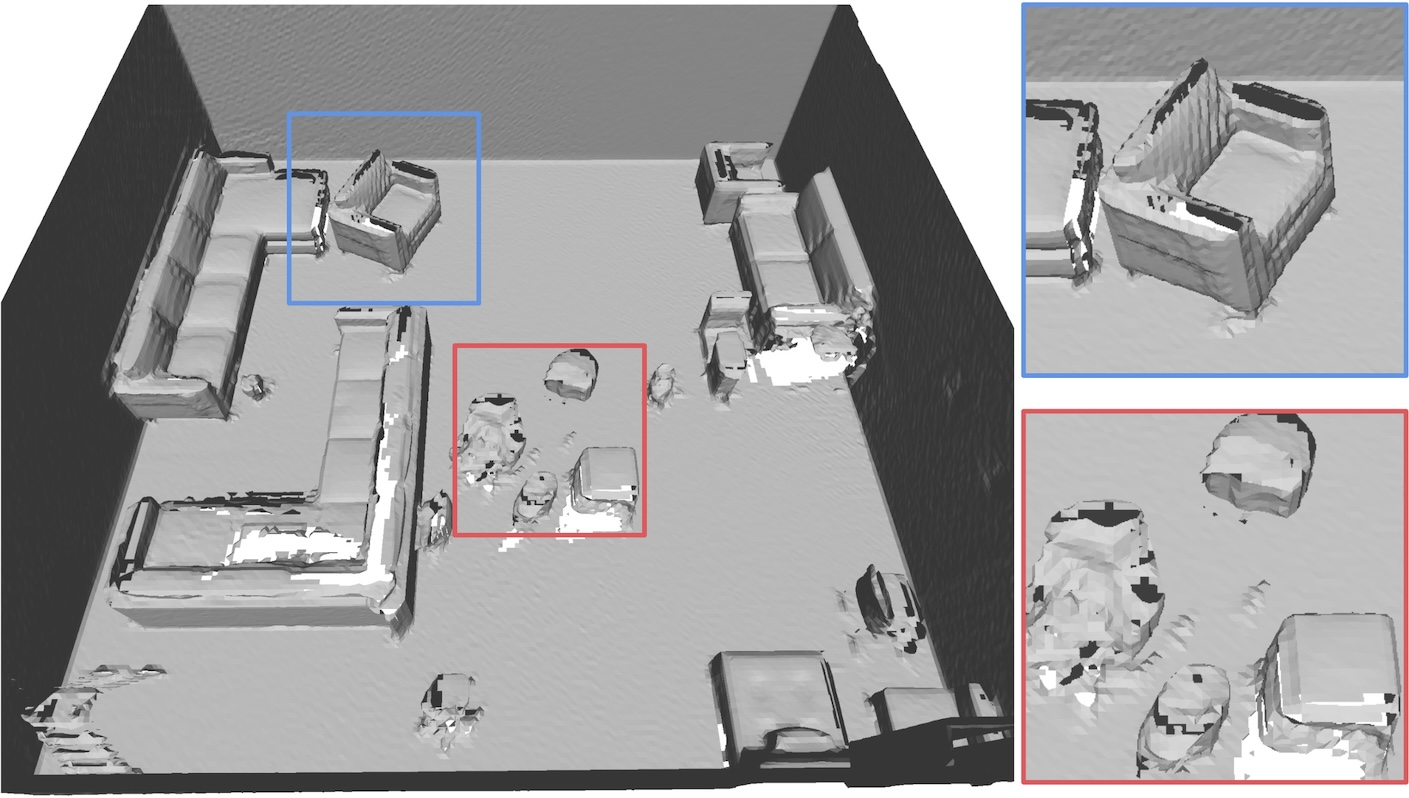}
    \end{subfigure}
    \hfill
    \begin{subfigure}[b]{\subfigwidth\textwidth}
        \centering
        \caption{\tiny \textbf{\EFMmodel{}}}
        \includegraphics[width=\textwidth]{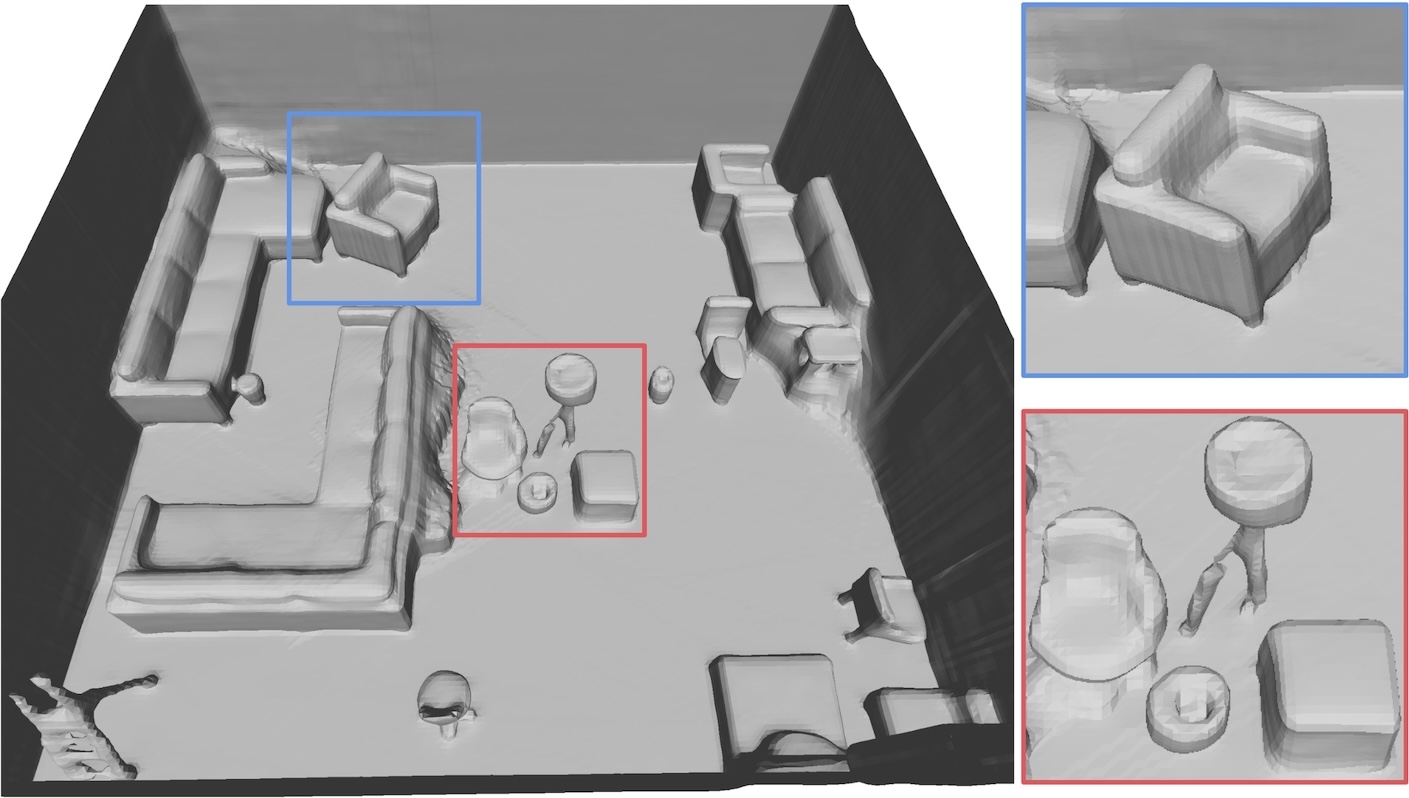}
    \end{subfigure}
    \hfill
    \begin{subfigure}[b]{\subfigwidth\textwidth}
        \centering
        \caption{\tiny \textbf{NeuralRecon}}
        \includegraphics[width=\textwidth]{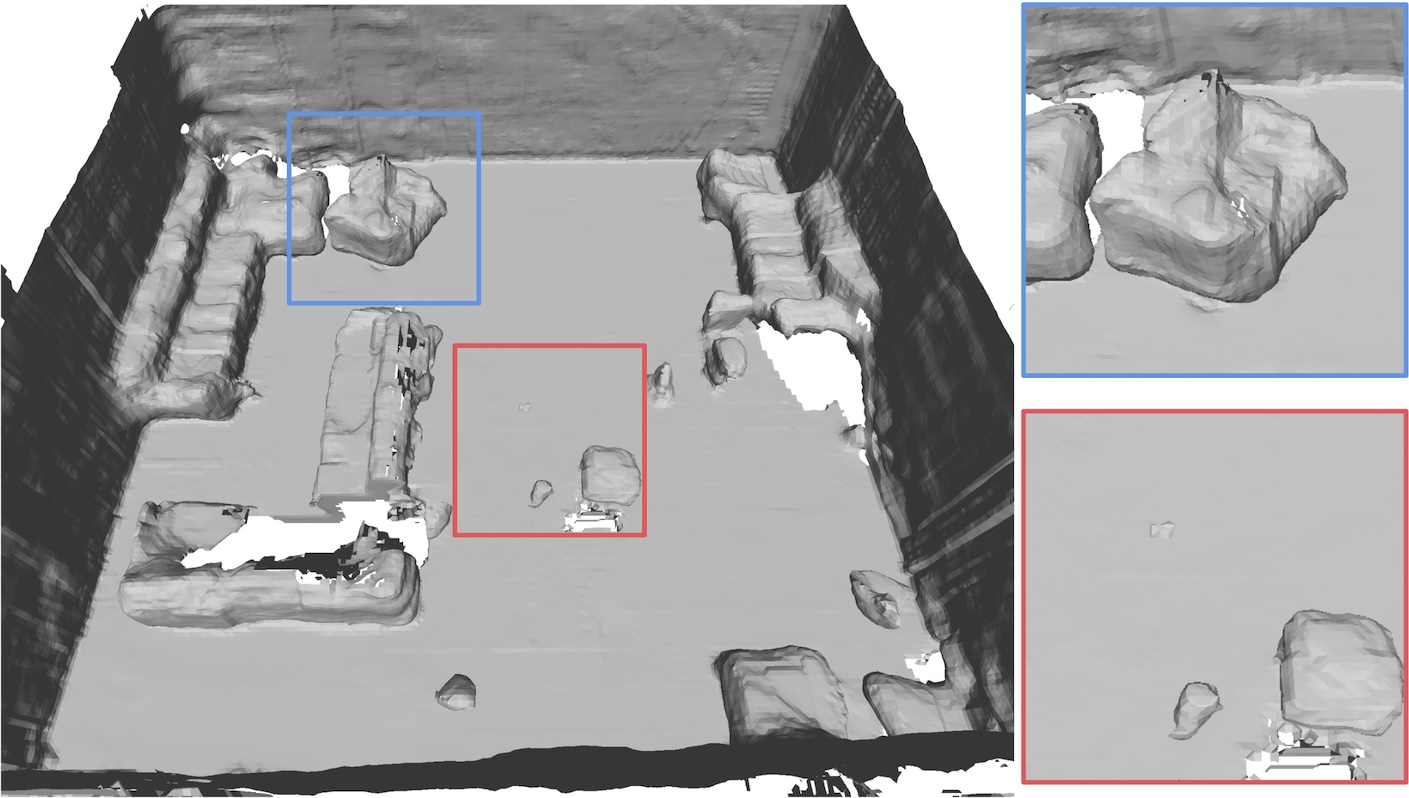}
    \end{subfigure}
    \hfill
    \begin{subfigure}[b]{\subfigwidth\textwidth}
        \centering
        \caption{\tiny \textbf{NeuralRecon OTS}}
        \includegraphics[width=\textwidth]{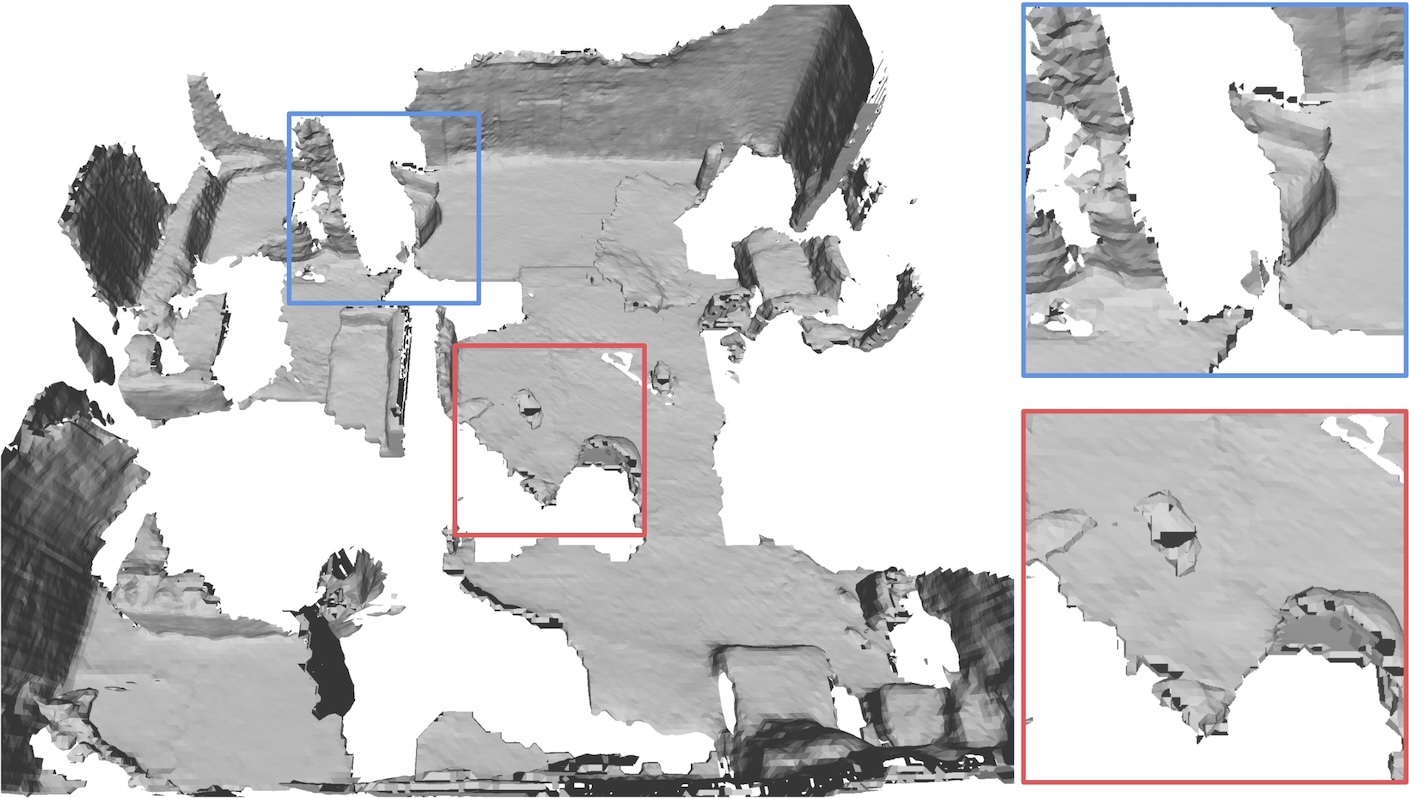}
    \end{subfigure}
    \hfill
    \begin{subfigure}[b]{\subfigwidth \textwidth}
        \centering
        \caption{\tiny \textbf{ZoeDepth}}
        \includegraphics[width=\textwidth]{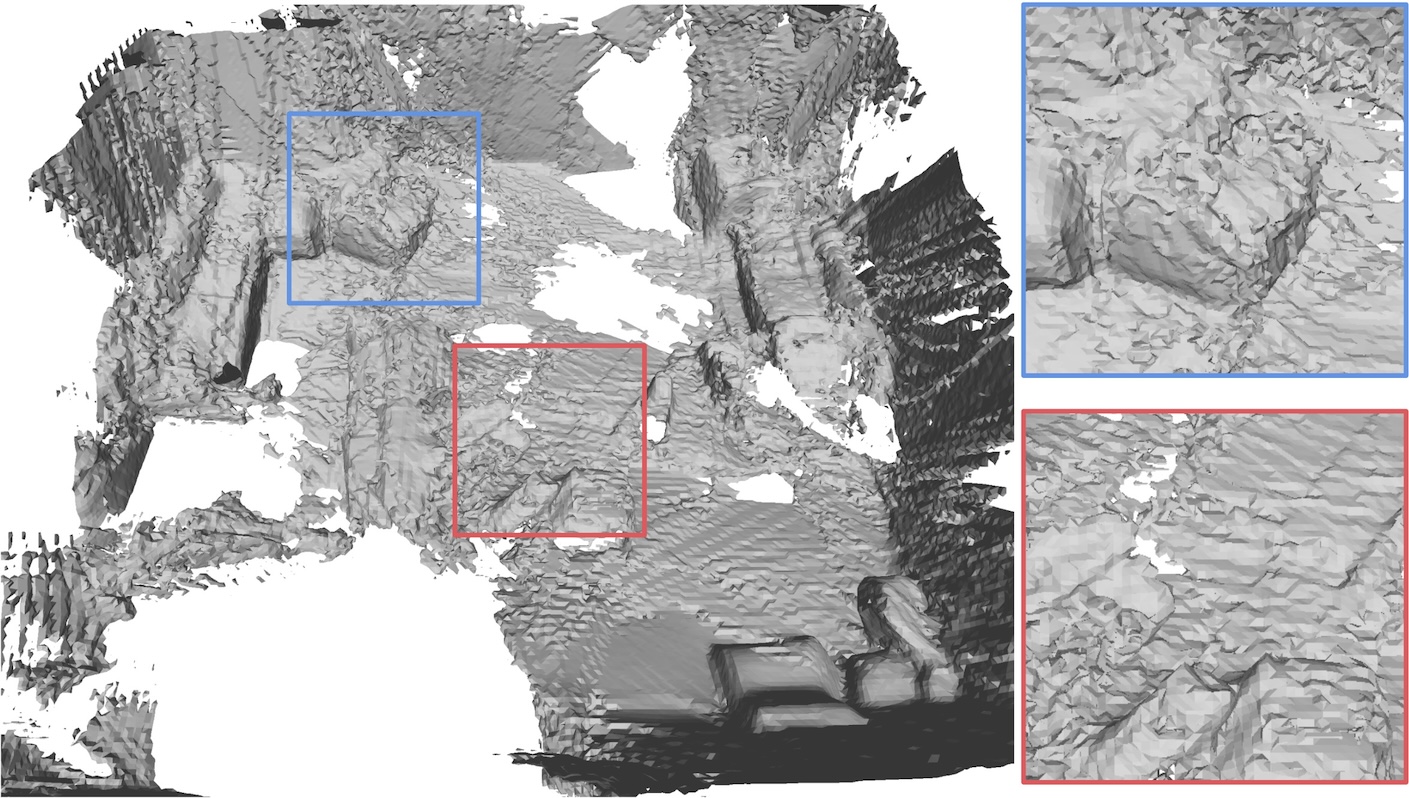}
    \end{subfigure}
    \hfill
    \begin{subfigure}[b]{\subfigwidth\textwidth}
        \centering
        \caption{\tiny \textbf{SimpleRecon}}
        \includegraphics[width=\textwidth]{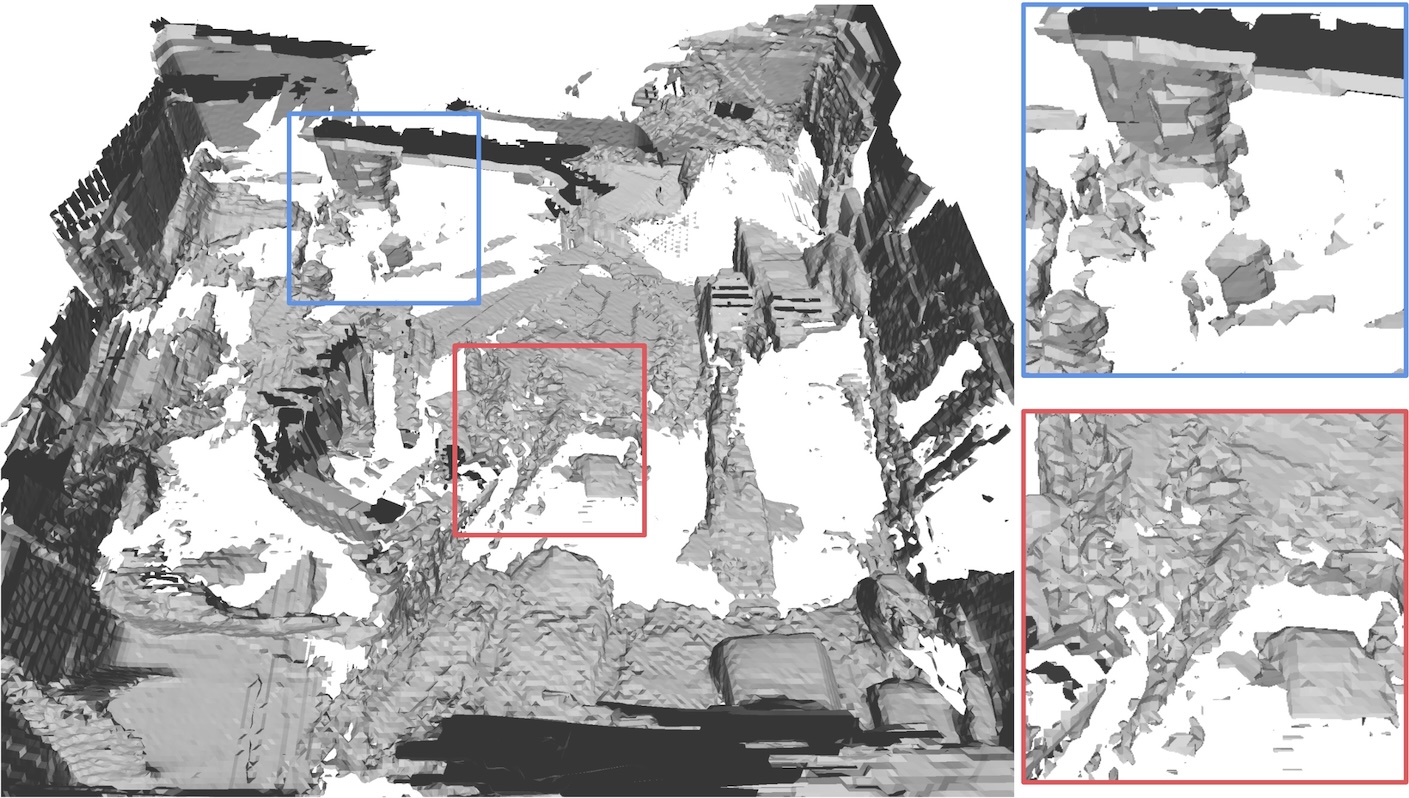}
    \end{subfigure}
    \hfill
    \begin{subfigure}[b]{\subfigwidth\textwidth}
        \centering
        \caption{\tiny \textbf{ConsistentDepth}}
        \includegraphics[width=\textwidth]{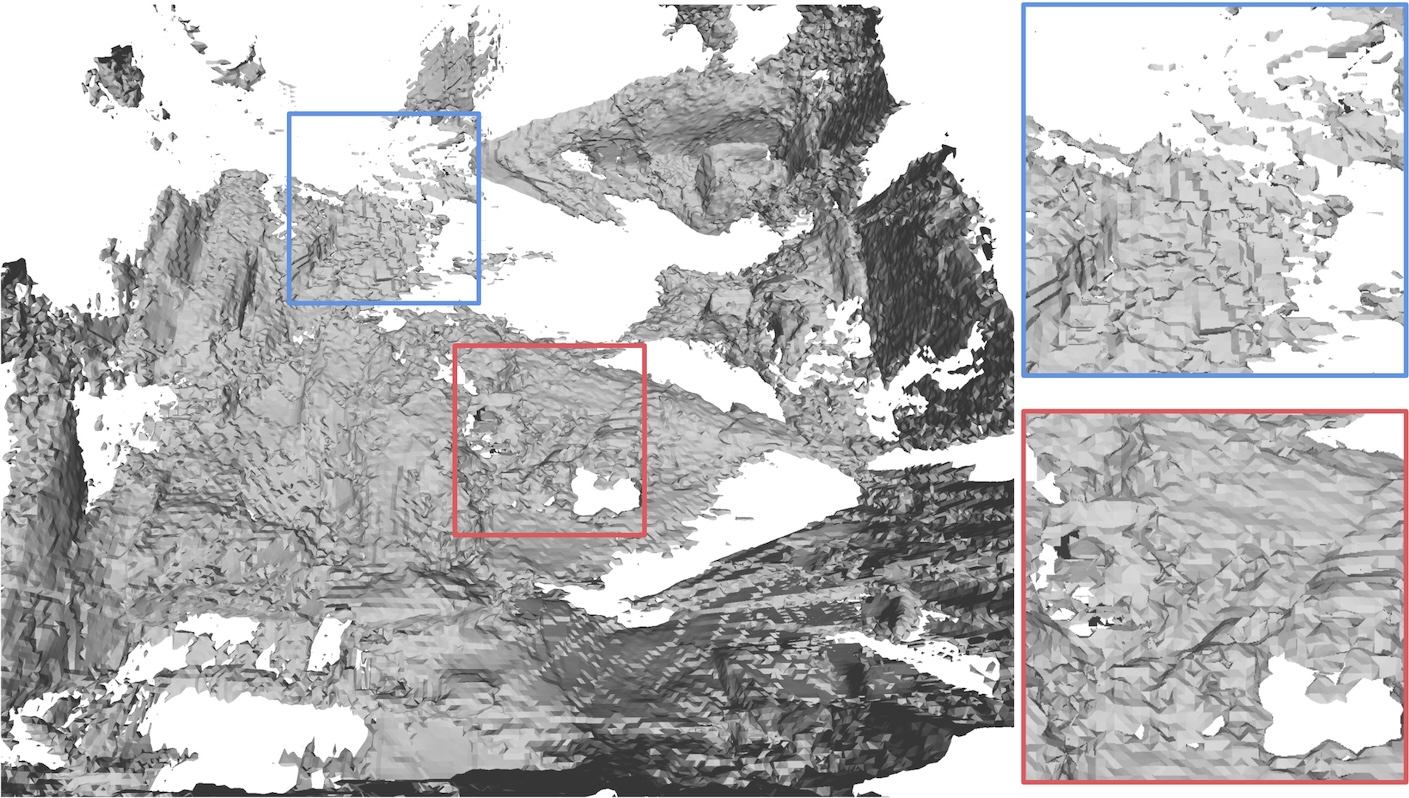}
    \end{subfigure}
    
    \caption{Surface reconstruction on ASE-100077. 3D volumetric methods like NeuralRecon and \EFMmodel{} reconstruct flat floors and orthogonal walls with explicit 3D priors. \EFMmodel{} delivers a more complete and detailed reconstruction. Depth-based methods fail to produce flat surfaces and result in major holes.}
    \label{fig:mesh_ase}
    \vspace{-5mm}
\end{figure}

Surface reconstructions are evaluated against GT meshes. We follow~\cite{jensen2014large} to compute mesh-to-mesh metrics as point-cloud-to-mesh metrics, with $Acc$ being the mean distance of points sampled from prediction to GT mesh triangles, and $Comp$ the  other way around (sampled points from GT to predicted surfaces). Please see the supplemental for a rigorous definition of the metrics.

Figure~\ref{fig:mesh_ase} and~\ref{fig:mesh_adt} show qualitative examples. The first three rows in Table~\ref{tab:surf_perf} demonstrate the quality of the GT depth maps, with a comparison of different camera models.

\renewcommand{\subfigwidth}{0.24}

\begin{figure}[htbp]
    \centering
    \begin{subfigure}[b]{\subfigwidth\textwidth}
        \centering
        \caption{\tiny \textbf{GT Mesh}}
        \includegraphics[width=\textwidth]{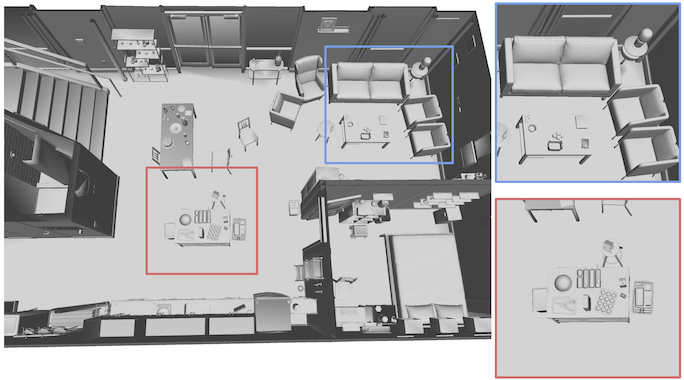}
    \end{subfigure}
    \hfill
    \begin{subfigure}[b]{\subfigwidth \textwidth}
        \centering
        \caption{\tiny \textbf{GT Depth Fusion}}
        \includegraphics[width=\textwidth]{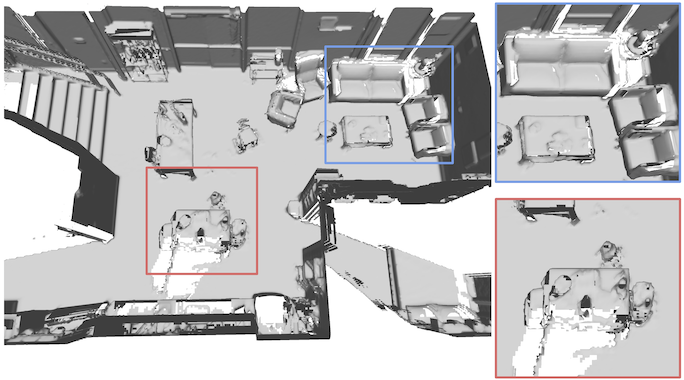}
    \end{subfigure}
    \hfill
    \begin{subfigure}[b]{\subfigwidth\textwidth}
        \centering
        \caption{\tiny \textbf{\EFMmodel{}}}
        \includegraphics[width=\textwidth]{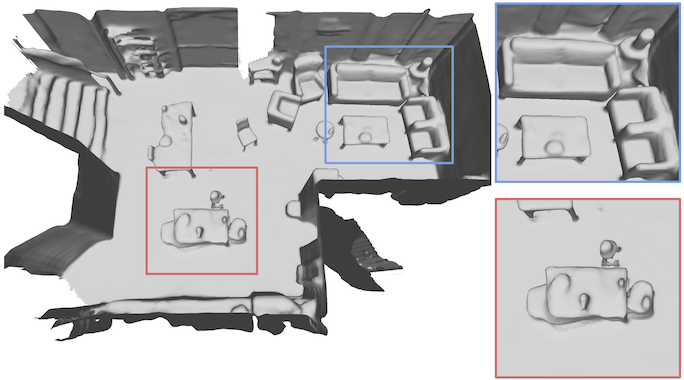}
    \end{subfigure}
    \hfill
    \begin{subfigure}[b]{\subfigwidth\textwidth}
        \centering
        \caption{\tiny \textbf{NeuralRecon}}
        \includegraphics[width=\textwidth]{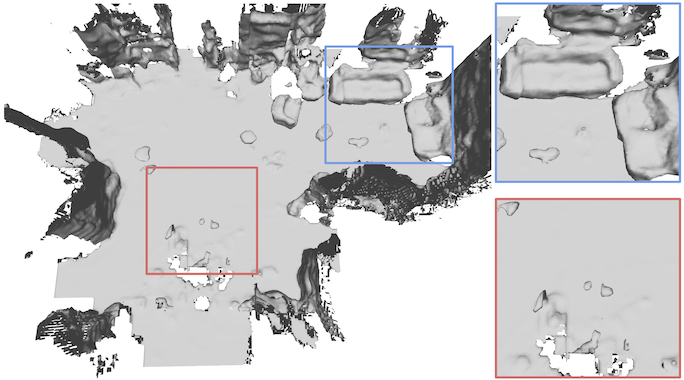}
    \end{subfigure}
    \hfill
    \begin{subfigure}[b]{\subfigwidth\textwidth}
        \centering
        \caption{\tiny \textbf{NeuralRecon OTS}}
        \includegraphics[width=\textwidth]{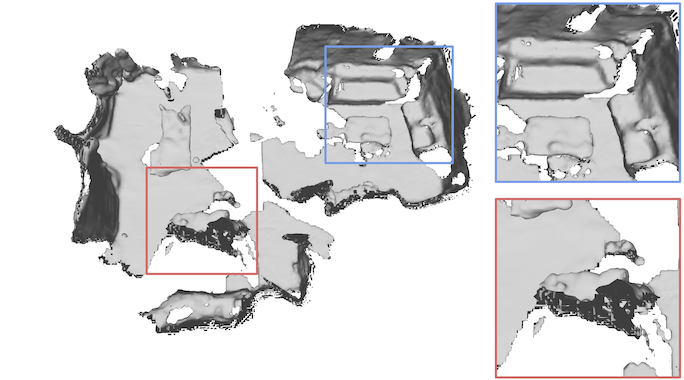}
    \end{subfigure}
    \hfill
    \begin{subfigure}[b]{\subfigwidth \textwidth}
        \centering
        \caption{\tiny \textbf{ZoeDepth}}
        \includegraphics[width=\textwidth]{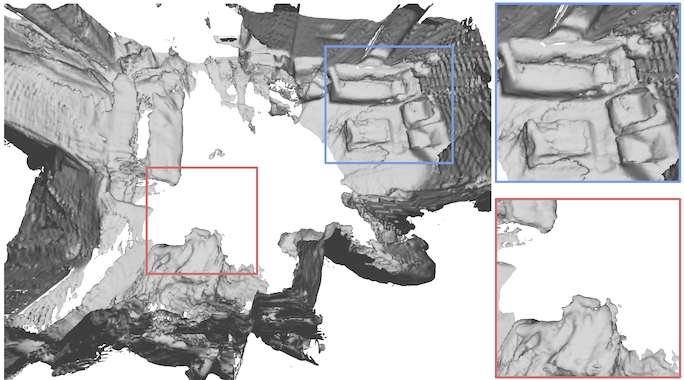}
    \end{subfigure}
    \hfill
    \begin{subfigure}[b]{\subfigwidth\textwidth}
        \centering
        \caption{\tiny \textbf{SimpleRecon}}
        \includegraphics[width=\textwidth]{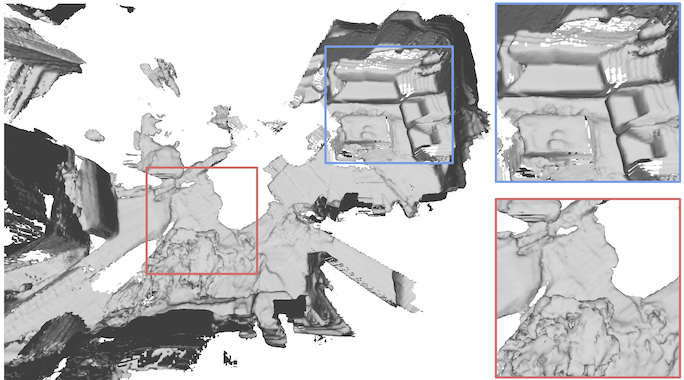}
    \end{subfigure}
    \hfill
    \begin{subfigure}[b]{\subfigwidth\textwidth}
        \centering
        \caption{\tiny \textbf{ConsistentDepth}}
        \includegraphics[width=\textwidth]{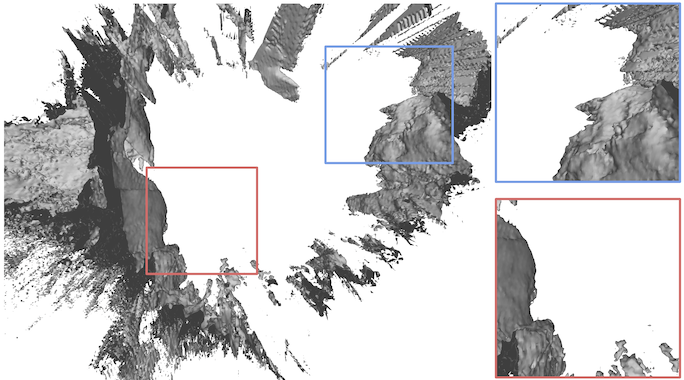}
    \end{subfigure}

    \caption{Surface reconstruction on ADT-106. It shows how models trained on simulation generalize to real data. \EFMmodel{} has the best generalization ability among all the methods.}
    \label{fig:mesh_adt}
    \vspace{-5mm}
\end{figure}

Using ScanNet-linear camera makes the fusion results much worse on completeness as it has limited FoV as shown in Fig.~\ref{fig:camera_lin}.
Fusing monocular depth methods generally does not work well as the scale of the per-frame depth maps can change due to inherent scale ambiguity which leads to incorrect fusion. 

Fusion of multi-view depth methods visually looks more consistent as shown in Fig.~\ref{fig:mesh_ase} and~\ref{fig:mesh_adt}, but does not have a good result quantitatively, mainly due to noisy depth estimation on low-textured regions such as walls. 
Volumetric methods generally perform better than depth-based methods. We test both the OTS NeuralRecon model trained on ScanNet, and the model re-trained on ASE dataset. Since the input to the OTS NeuralRecon model is ScanNet-linear images, the input has limited FoV and hence has worse completeness.

\begin{table}[]
    \centering
    \resizebox{\textwidth}{!}{%
    \begin{tabular}{c|c|c||c|c|c|c||c|c|c|c}
           & Model & Modality & ASE & ASE  & ASE & ASE & ADT & ADT  & ADT & ADT \\ 
           &   &      &  Acc $\downarrow$   & Comp $\downarrow$ & Prec $\uparrow$  &  Recal $\uparrow$ &  Acc $\downarrow$ & Comp $\downarrow$ & Prec $\uparrow$   &  Recal $\uparrow$ \\ \hline 
        \textit{GT-Depth-Fisheye} & - & frame & 0.011 & 0.473 & 0.945 & 0.568 & 0.060 & 1.392 & 0.793 & 0.360 \\ 
        \textit{GT-Depth-MaxLinear} & - & frame & 0.014 & 0.487 & 0.944 & 0.507 & 0.052 & 1.463 & 0.807 & 0.357 \\
        \textit{GT-Depth-ScanNet}  & - & frame & 0.015 & 1.183  & 0.952 & 0.386 & 0.020 & 2.744 & 0.907  & 0.178 \\ 
        \textit{Semidense Points} & - & pts & 0.034 & - & 0.943 & - & 0.122 & - & 0.720 & - \\ \hline
        ZoeDepth~\cite{bhat2023zoedepth} & OTS & frame & 0.368 & 1.225 & 0.290 & 0.130 & 0.417 & \textbf{2.127} & 0.200 &  0.076 \\
        ConsistentDepth~\cite{khan2023tcod} & OTS & frame & 0.349 & 1.304 & 0.277 & 0.125 & 0.603 & 2.759 & 0.145 & 0.045 \\
        SimpleRecon~\cite{sayed2022simplerecon} & OTS & snippet & 0.539 & 3.064 & 0.257 & 0.064 & 0.326 & 3.063 & 0.257 & 0.064 \\ 
        NeuralRecon~\cite{sun2021neucon}& OTS & snippet  & 0.110 & 1.952 &0.491 & 0.160 & 0.183 & 3.905& 0.371& 0.043\\
        NeuralRecon~\cite{sun2021neucon} & ASE & snippet  & 0.212 & 1.103 & 0.512 & 0.241 &0.307 &3.383 & 0.474 & 0.061\\
        \EFMmodel{} (ours) & ASE & snip$+$pts & \textbf{0.057} & \textbf{0.877} & \textbf{0.822} & \textbf{0.405} & \textbf{0.182} & 3.105 & \textbf{0.594} & \textbf{0.106} \\ 
    \end{tabular}
    }
    \caption{We show in-domain performance on ASE validation dataset as well as sim-to-real generalization performance on the ADT dataset. Accuracy and completeness are measured in meters. Precision and recall are given at a 5cm threshold. We train NeuralRecon and \EFMmodel{} on the ASE dataset. For other baseline methods we use the off-the-shelf (OTS) models. Depth-based methods such as ZoeDepth have better completeness on ADT as they are not limited by a 3D bounding volume.}
    \label{tab:surf_perf}
    \vspace{-5mm}
\end{table}


Both visually and quantitatively, \EFMmodel{} produces much better reconstruction with more smooth surfaces in contrast to the noisy fusion results from depth-based methods. Compared with NeuralRecon trained on the same data, \EFMmodel{} also adds more finer object details shown in zoom-in views.
We hypothesise that \EFMmodel{} performs better than NeuralRecon for two reasons. First, \EFMmodel{} takes advantages of the input semi-dense points, which provide a strong cue for surface prediction and have high accuracy as shown in the \textit{semi-dense points} row. Second, \EFMmodel{} uses DINOv2.5 as the 2D backbone.
%

\section{Limitations and Societal Impact}

The \EFMmodel{} 3D feature lifting mechanism and semi-dense point geometry inputs assume a mostly static world. Though the model is robust to some dynamics and scene changes (as they are present in the ADT and \AOThree{} real-world datasets), the model cannot handle large scene dynamics. Additionally, the \EFMmodel{} model has a limited 3D viewing frustum (by default we use $4m\times 4m\times 4m$), which cannot process far away scene geometry. We believe that with long enough egocentric data capture we will observe the majority of indoor scenes with this assumption.

All real-world Project Aria sequences go through an anonymization pipeline that blurs all personally identifying information. In addition all data collectors gave consent to the publishing of their data. We believe EFMs are poised to provide a profound impact for bringing the context of the physical world to egocentric devices such as AR glasses. While our datasets are collected in fully consented environments, the profound fidelity at which \EFMmodel{} already reconstructs objects and scenes reveals the pervasive nature of egocentric sensor data. It is important as EFMs continue to improve and evolve that strong privacy and consent models are incorporated into the models.

\section{Conclusion}

In this manuscript, we introduce the concept of 3D Egocentric Foundation Models that integrate egocentric sensor data for 3D scene understanding. We identify two core tasks for 3D EFMs---3D object detection and surface regression---and create a benchmarks for each task using high quality annotations of datasets captured with Project Aria glasses~\cite{engel2023project}. When evaluating these tasks over an entire sequence of egocentric data (as opposed to a single frame), existing methods exhibit poor 3D consistency in their predictions that leads to poor performance on the EFM3D benchmark. To address this, we design a simple but effective 3D backbone for egocentric data, \EFMmodel{}, that leverages semi-dense points and image features to produce a 3D voxel grid of features. \EFMmodel{} outperforms all other methods when evaluated on the proposed EFM3D benchmark.
The simplicity of this architecture underscores the effectiveness of the 3D inductive biases in \EVL{}. 
We encourage the development of more sophisticated models that can exploit the richness of egocentric 3D data even more effectively, including the incorporation of dynamic scene understanding and user interaction modeling. Such modeling advancements could improve the performance even further and extend the applicability of 3D EFMs to a wider range of real-world scenarios.

\section{Acknowledgements}

We would like to thank Suvam Patra, Armen Avetisyan, Samir Aroudj, Chris Xie, and Henry Howard-Jenkins for help with the ASE dataset. 
And Yang Lou, Kang Zheng, Shangyi Cheng, and Xiaqing Pan for helping with the dataset releases,
as well as Thomas Whelan for help with tuning the fusion system. 
We would like to thank Numair Khan for help with evaluating the TemporallyConsistentDepth pipeline and Campbell Orme for narrating the supplemental video.
Finally we would like to thank Dan Barnes, Raul Mur Artal, Lingni Ma, Austin Kukay, Rowan Postyeni, Abha Arora and Luis Pesqueira for help with the 3D OBB annotations. 


\clearpage

\bibliographystyle{splncs04}
\bibliography{main}

\clearpage

\appendix

In the following we provide additional details for the models we trained on ASE including \EVL{}, the ASE and AEO datasets we release,  more details for the EFM3D benchmark, and finally some additional experiments. 

\section{Model and Training Details}

We first describe more details for the \EVL{} model before detailing training procedures for all models.

\subsection{\EVL{} Model Details}

\noindent\textbf{\EVL{} Architecture Details}:
We use a frozen base DINOv2.5 backbone with an output feature dimension of 768. We upsample it using a simple 2D upsample network.
The 2D upsample network alternates 2D upsampling by a factor of $2\times$, $3\times3$ 2D convolution, ReLU activation, and batch-norm. In the upsampling we reduce the input feature dimension to an output dimension of 64. These upsampled 64-dimensional features are at the same resolution of the input image.
After resampling and aggregating these image features into the feature volume, we run a simple 3D U-Net to process the features in 3D.
The 3D U-Net downsamples the 3D volume three times down to 1/8th of the input resolution by iterating $3\times 3$ 3D convolutions, batch norm, ReLU activation and max pooling. The upsampling uses tri-linear $2\times$ upsampling layers that get combined with skip connections and then processed with a sequence of $3\times 3$ 3D convolutions, batch norm and ReLU. All intermediate feature dimensions are kept at 256. We currently keep the 3D object detection model and surface reconstruction model separate. 
The DINOv2.5 base model has 86.6M non-trainable parameters. 
On top of that the EVL model has 16.7M trainable parameters.  

\noindent\textbf{Computing Gravity-Aligned Voxel Grid}
We define a gravity-aligned voxel grid pose by aligning the camera pose to gravity as follows.
Given camera rotation $R_{wc} = [r_x, r_y, r_z]$ and unit-length gravity direction $g_w$ in world coordinates we compute the gravity aligned rotation $R_{wg}$ as:
\begin{align}
    R_{wg} &= [g_w, \lceil d_z \times g_w \rceil, \lceil d_z\rceil ] \quad \text{where}\quad
     d_z 
     = r_z - g_w^T r_z g_w \,,
\end{align}
and $\lceil x \rceil $ normalizes a vector $x$ to unit length.

\subsection{Model Training Details}


Overall the ASE training dataset contains 100k sequences. 
For OBB training we use a 10k sequence subset which contains 600k 1s snippets. 
For surface regression training we use a 1k sequence subset of 60k 1s snippets. We find these subsets are sufficiently large for solid performance at reasonable training times. For OBBs we find that the validation set performance saturates at 10k sequences with the current model architecture of \EVL{} (see Sec.~\ref{sup:scaling}). More details in Sec.~\ref{sup:dataset}.

\noindent\textbf{\EVL{} OBB training}: We train all configurations for 3D OBB regression for 10 epochs on the ASE 10k training dataset. We use a max learning rate of 5e-4 with an effective batch size of $160$. The learning rate is linear warmed up for $8\%$ and the cosine annealed. 
We set the loss weights to $w_c=100$, $w_{iou}=10$ and $w_{cls}=1$.

\noindent\textbf{\EVL{} Occupancy training}: We train all configurations for occupancy regression for 60 epochs on the ASE 1k training dataset. We use a max learning rate of 4e-4 with an effective batch size of 128. Similar to the OBB training the learning rate is linear warmed up until $8\%$ for the first 5 epochs. The surface loss by a factor of $1.0$ and the TV-loss by $0.01$.

\noindent\textbf{3DETR training}: We choose the base configuration of 3DETR targeted for SUN-RGBD with a GIoU matching cost of $5.0$ to train for oriented bounding box regression. We train for 50 epochs on the ASE training dataset and a cosine learning rate schedule with linear warmup. The max learning rate is 7e-4 and the effective batch size 320. We use the same point cloud cropping data augmentation as in the original implementation.

\noindent\textbf{ImVoxelNet training}: We choose the base configuration of ImVoxelNet with the SUB-RGBD config to train for oriented bounding box regression. We freeze the ResNet50 backbone to help with sim-to-real transfer. We train for 10 epochs on the ASE training dataset and a step learning rate schedule. The starting learning rate is $1e-4$ and gets reduced by a factor of 10x after $67\%$ and again after $83\%$ of training. The effective batch size is $320$. We provide max-linear rectified images.

\noindent\textbf{Cube R-CNN training}: We use the DLA backbone and train it on the ASE training dataset for 10 epochs. A max learning rate of 1e-4 is cosine rate scheduled. The effective batch size is 640. We train on max-linear rectified images.

\noindent\textbf{NeuralRecon training}: Among the surface reconstruction baseline methods, NeuralRecon is the only one we did the retraining. We train NeuralRecon using the default hyper-parameters on the ASE 1k training subset. It has a similar voxel configuration to \EVL{} Occupancy model, with a $[96, 96, 96]$ local volume and a 0.04m voxel size. We use a start learning rate of 2e-4. NeuralRecon has a two-phase training scheme, i.e. TSDF local volume regression and learned fusion training. We train each phase for 10 epochs.  


\section{Dataset Details}
\label{sup:dataset}
In the following we describe details of the Aria Synthetic Environment (ASE) training dataset as well as the real-world Aria Everyday Objects (AEO) 3D OBB validation dataset. 

\subsection{Aria Synthetic Environment (ASE) Details}

\textbf{Subselecting ASE Sequences}. As described earlier in Section \ref{sup:scaling}, we found that scaling the training beyond 10k sequences resulted in diminishing returns in terms of 3D Object Detection results on holdout ASE sequences. We thus report all OBB experimental results using a 10k sequence subset of the 100k sequences to reduce the engineering and infrastructure requirements for training.

\noindent\textbf{Subselecting ASE Classes}. We sub-select 27 total semantic classes in the ASE dataset by combining similar and filtering out certain classes from the original 42 semantic labels. We removed pure structural elements such as “Floor”, “Wall”, “Ceiling” and “Stairs”, because we assume that the full extent of the object can be reasonably visible within a short video snippet. We combined certain similar categories such as "Ottoman", “SittingChair” and “bench” into a single class called “Chair”. We also removed some vague classes such as “ElectricalCable” and “Decoration”. “Cutlery” and “BatteryCharger” are also removed because they are filtered by the absolute pixel counting. In the end we were left with 27 total classes.

\begin{figure}
    \includegraphics[width=1.0\textwidth]{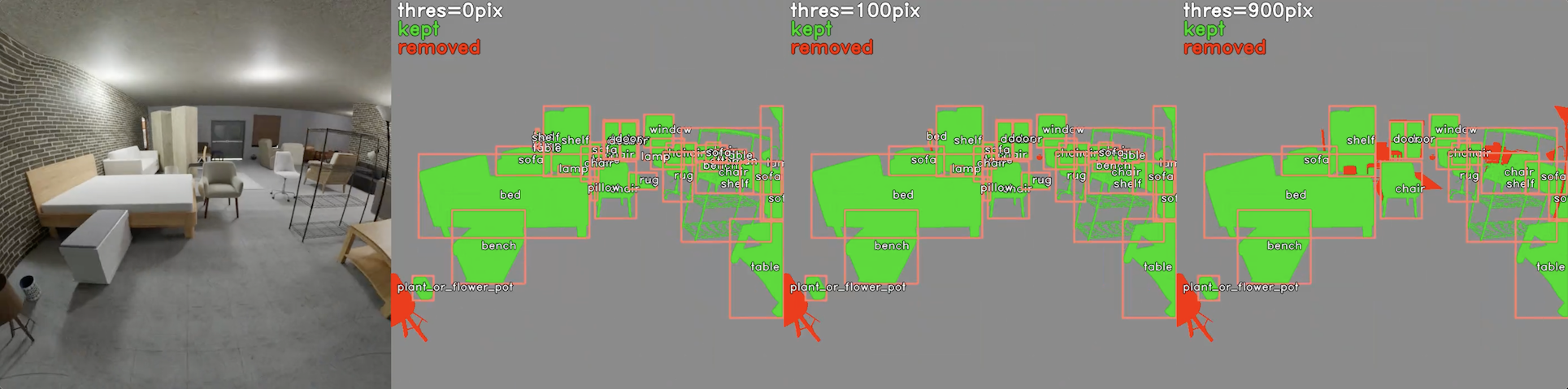}
    \caption{Heuristic for assigning 2D-3D observability. For a given RGB image (left), we visualize various thresholds for 2D-3D observability, going from more objects (left-middle) to fewer objects (right). We ultimately used the 100px threshold (middle-right).  \label{fig:ase_association}}
\end{figure}

\noindent\textbf{Visibility Annotation Details}. In order to train a 3D object detector that operates on short video snippets, it is necessary that we select which 3D objects to assign as "visible" within the video snippet. With a simulated dataset such as \ASE{}, we have per pixel observability information for each camera. We first associate the 3D OBBs with the 2D instance masks of the images to determine the observability. As no OBBs in \ASE{} intersects with each other, we achieve this by simply unprojecting the pixels inside an instance mask using ground-truth depth maps and the camera calibrations, then associate the mask with the OBB which contains the unprojected points. We further filter out the OBBs of which the mask contains too few pixels to account for the occlusion and very far away objects. In this paper, we use a simple heuristic which is to filter out the OBB observations with less than $100$ pixels in the 704x704 RGB camera image.

\begin{figure}
\centering
    \includegraphics[width=0.8\textwidth]{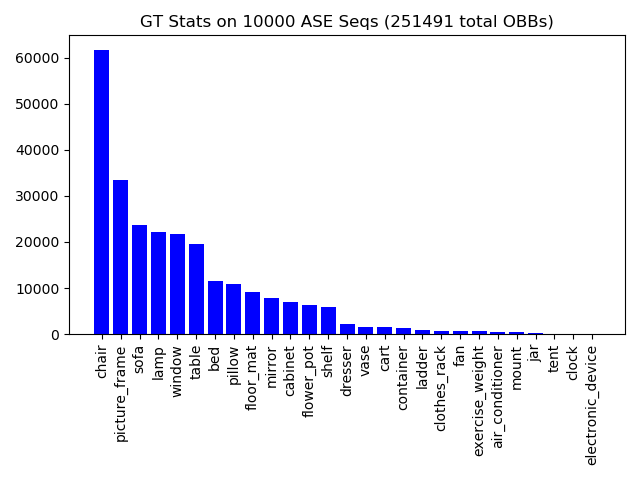}
    \caption{3D Object Bounding box per class frequency histogram for ASE.}
    \label{fig:class_hist_ase}
\end{figure}

\noindent\textbf{\ASE{} Per Class Statistics}. A histogram of the per-class 3d orientend bounding box counts are shown for 10k \ASE{} sequences in Fig.~\ref{fig:class_hist_ase}, for a total of about 251k OBBs.

\subsection{Aria Everyday Objects (AEO) Details}

\begin{figure}
\centering
    \includegraphics[width=0.88\textwidth]{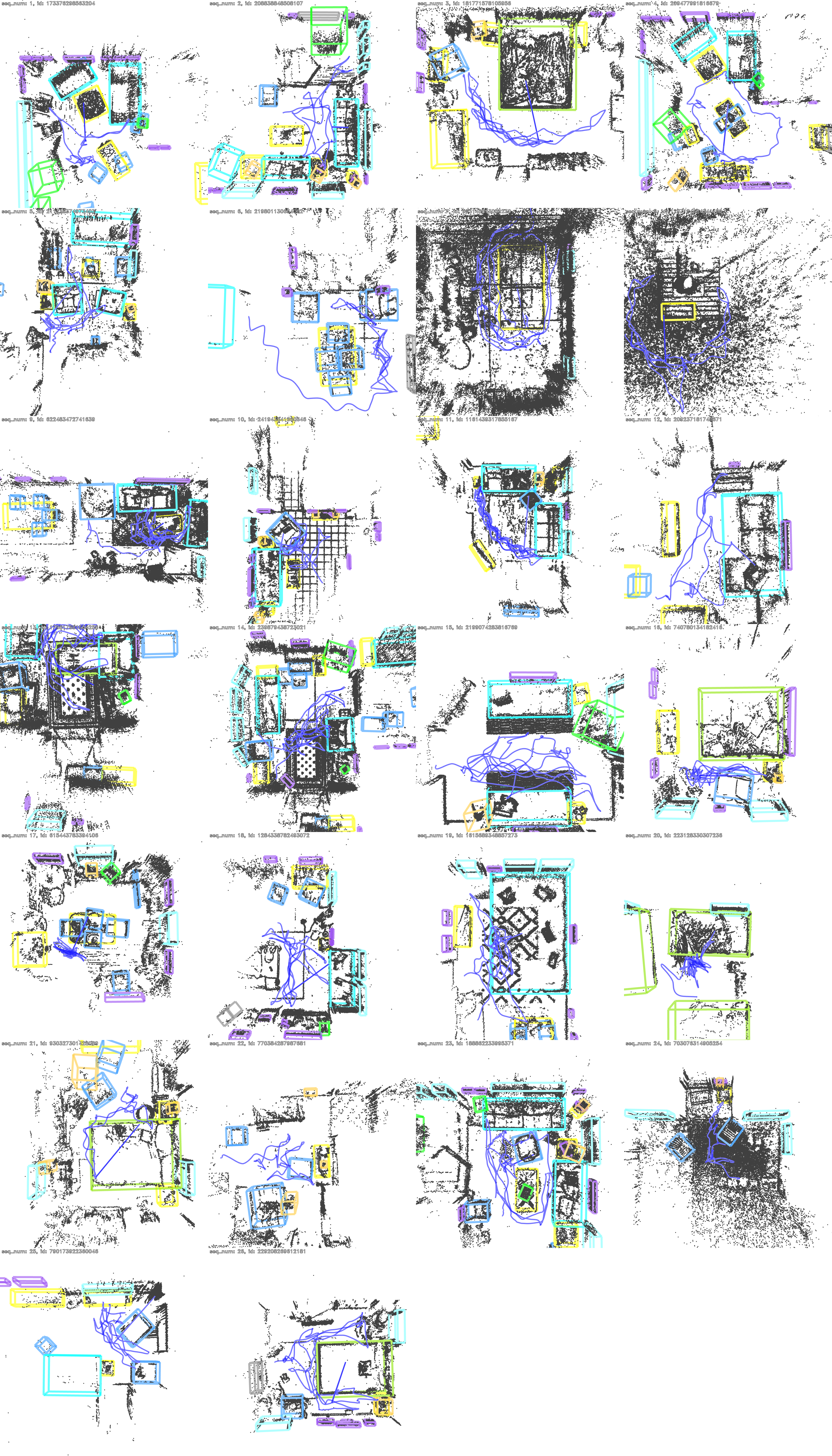}
    \caption{Top-down view of each of the 26 scenes from the AEO dataset show the OBBs (colored by class), point cloud (black) and Project Aria trajectory (blue). \label{fig:aeo_overhead}}
\end{figure}

\begin{figure}
\centering
    \includegraphics[width=0.88\textwidth]{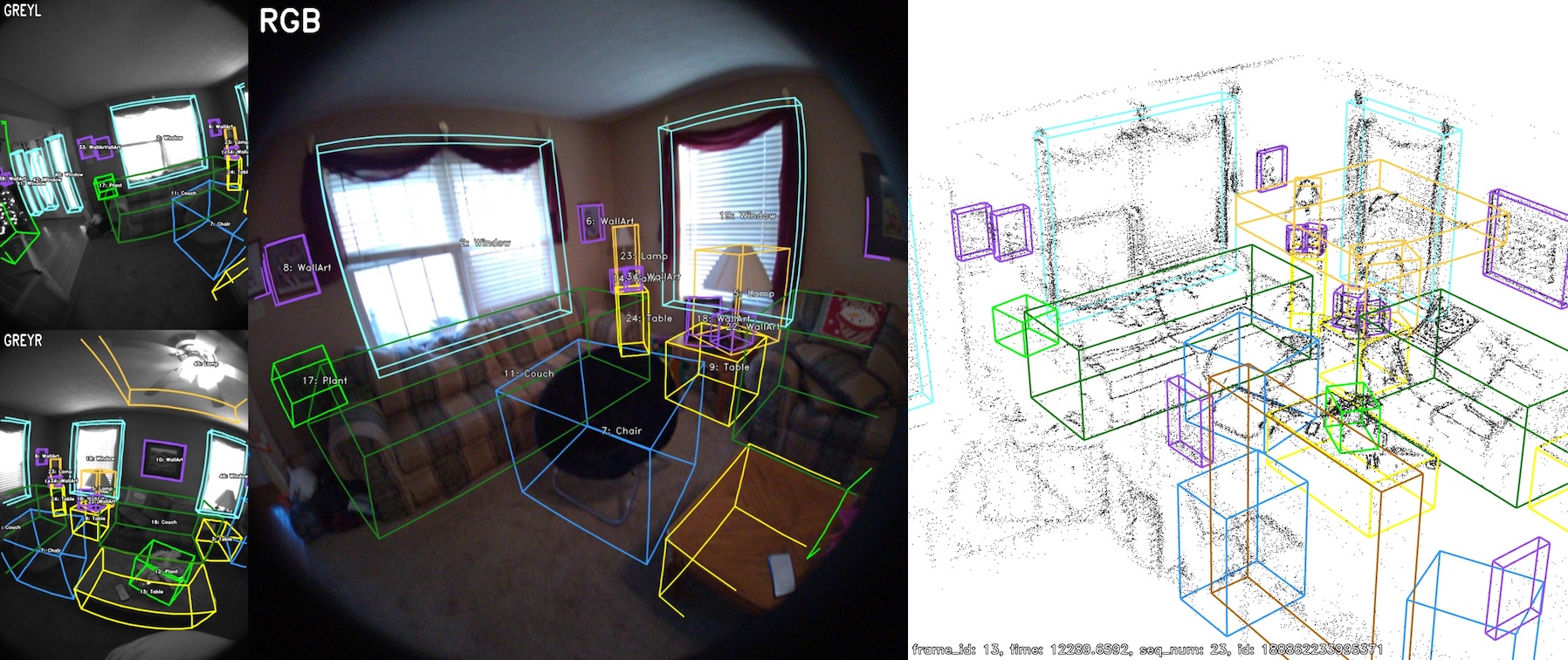}
    \includegraphics[width=0.88\textwidth]{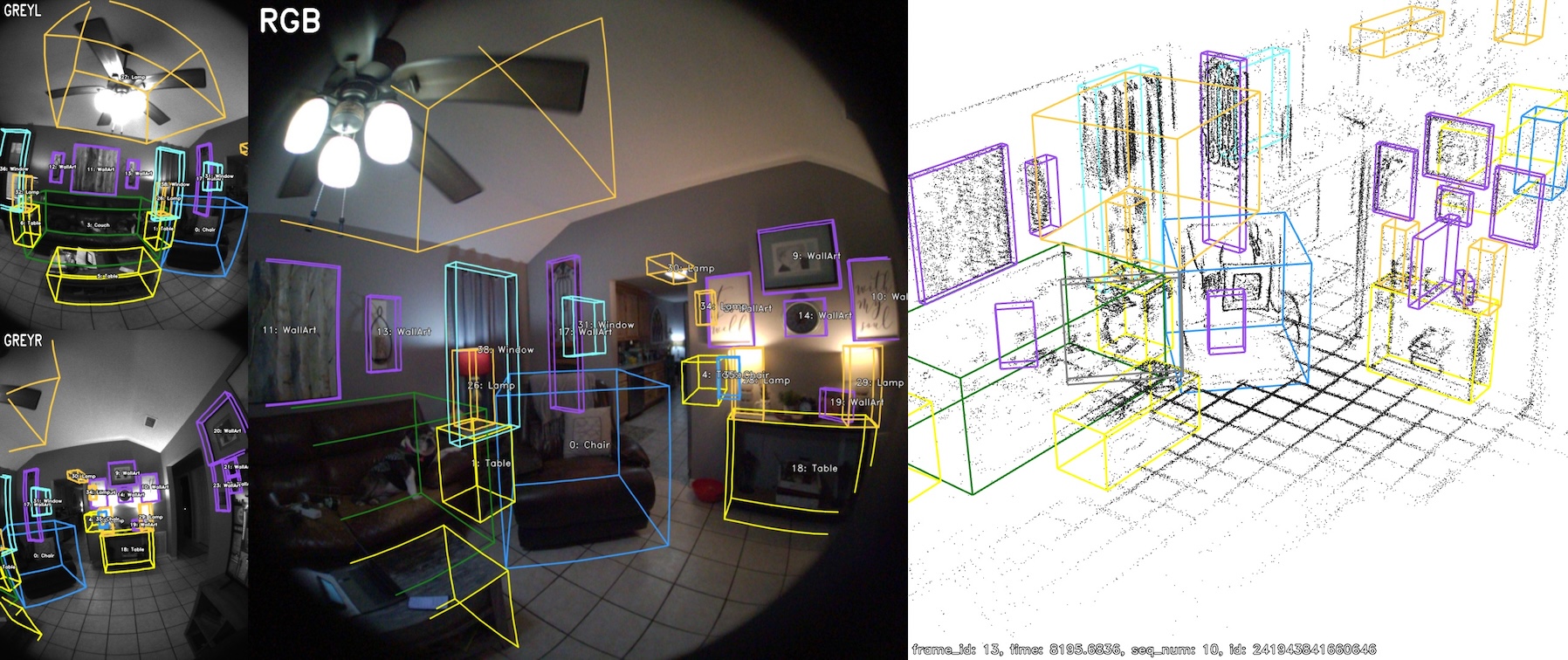}
    \includegraphics[width=0.88\textwidth]{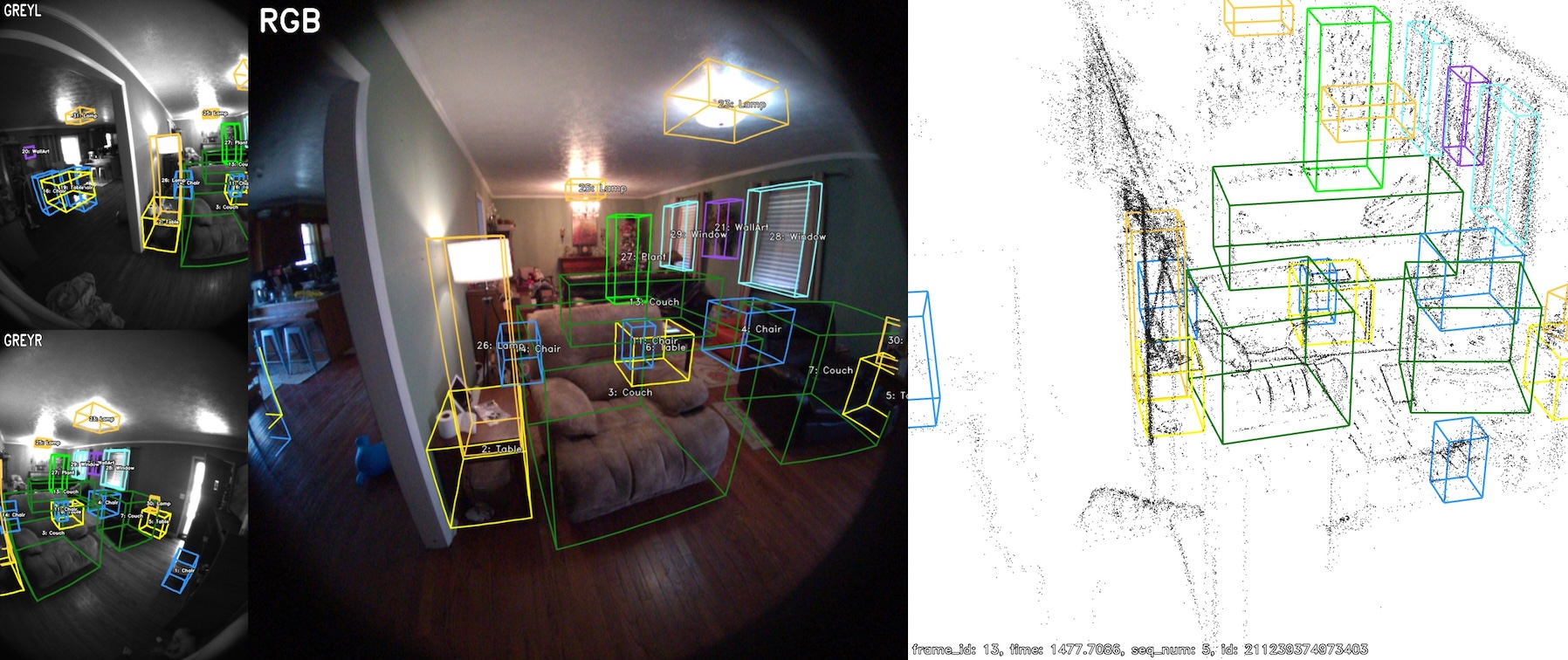}
    \includegraphics[width=0.88\textwidth]{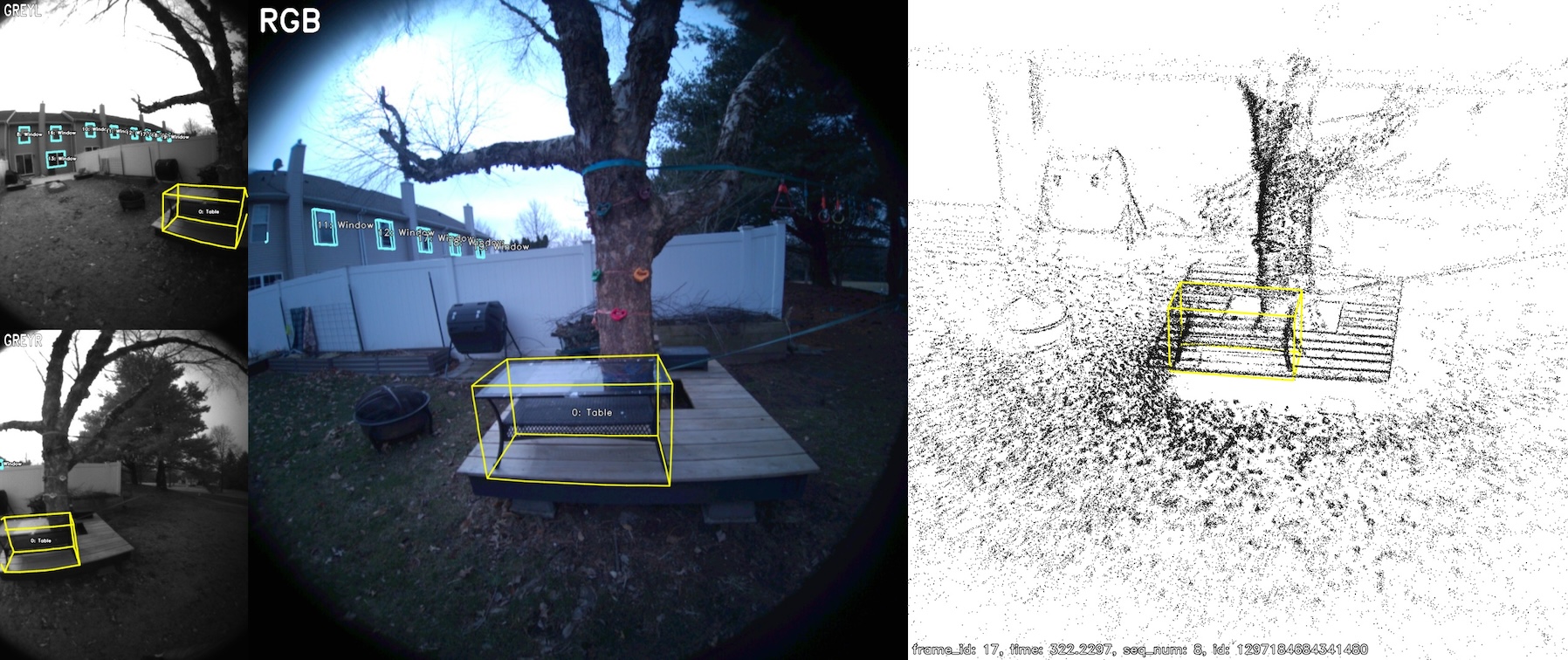}

    \caption{Example Ground Truth 3D Object Bounding Box Labels for AEO, each row is from a different sequence. The left two images show the OBBs visualized in the greyscale images, the middle shows the OBBs in the RGB image, and the right side shows a 3D visualization of the OBBs alongside the full scene point cloud \label{fig:aeo_examples}}
\end{figure}

\noindent\textbf{AEO More Details.} The AEO dataset consists of short (1-2 minute) recordings from Project Aria, collected by non-computer vision expert wearers. This results in trajectories that do not scan entire scenes but rather explore scenes in a more natural way, often focused on one or a handful of objects. The dataset consists of 20 unique scenes across 26 sequences with 584 total OBBs. Most sequences are collected indoors, though there are a few collected in backyards and a garage. Top down views of the dataset are visualized in Fig.~\ref{fig:aeo_overhead} and visualizations with the 3D OBBs projected onto the three Project Aria images are shown in Fig.~\ref{fig:aeo_examples}.

\noindent\textbf{Remapping ASE Classes to AEO}. The ASE taxonomy does not perfectly match the AEO taxonomy. The classes "Couch", "Table", "Bed", "Chair", "Window", "Mirror" and "Lamp" have the same name so we match those directly. The classes "PictureFrameOrPainting" is mapped to "wall art" and "PlantOrFlowerPot" is mapped to "plant". We note that the definition of each OBB differs slightly, for example "lamp" in AEO covers floor lamps, chandeliers and recessed overhead lighting for example, while in ASE it typically only covers floor lamp. These slight differences in taxonomy definition are another potential source of limitations for sim-to-real generalization. Per-class object detection performance differences shown in Sec.~\ref{supp:per_class_perf} indicate as much.
We leave open-taxonomy OBB detection to future work which should mitigate this issue.

\begin{figure}
\centering
    \includegraphics[width=0.7\textwidth]{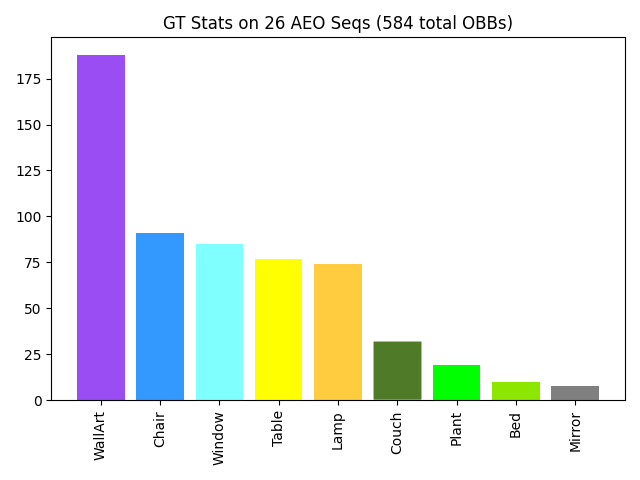}
    \caption{3D Object Bounding box per class frequency histogram for AEO. The colored bars for AEO correspond to the classes visualized throughout this paper.}
    \label{fig:class_hist_aeo}
\end{figure}

\noindent\textbf{\AEO{} Per Class Statistics}. A histogram of the per-class 3d orientend bounding box counts are shown for 26 \AEO{} sequences in Fig.~\ref{fig:class_hist_aeo}, for a total of 584 OBBs.

\section{EFM3D Benchmark Details}

We first provide more details for the tracking and fusion of 3D OBBs as well as scene surfaces before describing the surface error metrics in more detail.



\subsection{Tracking and Fusion of 3D Bounding Box Predictions}

Sequential object detectors predict an independent set of 3D OBBs per frame (e.g., CubeRCNN~\cite{Brazil_2023_CVPR}) or per snippet (e.g., ImVoxelNet~\cite{rukhovich2022imvoxelnet}). Directly appending all the independent sets of 3D OBBs to the scene state results in noisy and duplicated objects. Therefore, we propose a simple object tracker named \textit{ObbTracker} to track and update the objects in the scene from the sequential predictions.

ObbTracker gets the raw predictions as the inputs and maintains a scene state which consists of a set of tracked 3D OBBs. Note that the raw predictions from the object detector are transformed into the world coordinate system using the camera poses. After filtering out the low-confidence predictions, ObbTracker performs 3 key steps: Association, Update, and Removal, which we will explain in detail.

\noindent \textbf{Association}. We match the predictions that clear a matching score threshold of $p_{assoc}$ and the tracked OBBs in the scene using Hungarian Matching with the cost function:
\begin{align}
C_{assoc} = w_1 \cdot C_{class} + w_2 \cdot C_{bbox2} + w_3 \cdot C_{bbox3} + w_4 \cdot C_{iou2d} + w_5 \cdot C_{iou3d},
\end{align}
where $C_{class}$ measures the discrepancy between the classes of the predictions and the tracked 3D OBBs, which is implemented as the probability of tracked OBB’s class evaluated on the predicted class distribution, $C_{bbox2}$ is the distance between the 2D bounding box centers, $C_{bbox3}$ is the distance between the 3D bounding box centers, $C_{iou2d}$ is the 2D Intersection over Union (IoU) between the 2D bounding boxes, and $C_{ioui3d}$ is the 3D IoU between the 3D bounding boxes. 
Specifically we set $w_1=8$, $w_2=0$, $w_3=1$, $w_4=2$, $w_5=0$ for all experiments.
Then we further filter out the matches where the 2D and 3D IoUs are higher than predefined threshold of $0.2$ for both 2D and 3D IoU.
Those good matches will be used for updating the tracked OBBs. The remaining predictions are added to the scene as new candidates if they clear an instantiation score threshold $p_{inst}$. We tune the instantiation and association score thresholds but typically keep $p_{assoc} = p_{inst} - 0.05$. 

\noindent \textbf{Update}. The states of the tracked OBBs are updated with the associated predictions using running average. We define the states of an OBB as its scales $s \in \mathbb{R}^3$, pose in the world coordinate frame $T_{obj}^{w} \in SE(3)$, and the class probability distribution $c \in \mathbb{R}^k$ where $k$ is the number of total categories in the taxonomy. 
Given a predicted set of parameters $s'$, $c'$, and ${T_{obj}^{w}}'$, the running average of the object parameters is computed as
\begin{align}
    n' &= n+1 \\
    s &= (s n + s') / n'\\
    c &=  (c n + c') / n'\\
    T_{obj}^{w} &= T_{obj}^{w} \cdot \exp{({T_{obj}^{w}}' \boxminus T_{obj}^{w} / n')} \;,
\end{align}
where ${T_{obj}^{w}}'$ is the predicted object pose, $n$ is the number of observations of the tracked OBB, and $T_a \boxminus T_b = \log(T_a^{-1} T_b)$ is the difference between two poses on the pose manifold SE3 and $\exp{}$ is the exponential map to SE3. 


\noindent \textbf{Removal}. After the update step, we remove the tracked OBBs which do not get enough observations $w_{min}$ within a certain period of time $t_{inst}$ after it’s added to the scene. We set $n_{min}=2$ and $t_{inst} =1s$.  
We further suppress duplicate scene OBBs by using both 3D non-maximum suppression (NMS) with IoU threshold of $0.1$ and 2D NMS with IoU threshold of $0.5$.

\subsection{Fusion of Surface Predictions}
To persist local surface observations into a globally consistent estimation, a fusion system is needed for either depth-based or volumetric methods.

\noindent \textbf{Depth map fusion.} When evaluating depth-map-based surface reconstruction, we use a simple implementation from~\cite{zeng20163dmatch} to fuse per-frame observations into an TSDF volume~\cite{newcombe2011kinectfusion}. The method iteratively projects global volume to depth maps, and computes TSDF values by merging the current depth observations in a global averaging way. After getting the global TSDF volume, the meshes are extracted using the marching cubes algorithm~\cite{lorensen1998marching}. 

\noindent \textbf{Occupancy fusion.} For volumetric methods like \EFMmodel{}, a local occupancy volume is predicted per snippet. We fuse the local occupancy grids into a global grid. Each local occupancy grid is iteratively fused into the global volume, by transforming the global volume to local ones and sampling from them. Similar to depth map fusion, We extract mesh surfaces using marching cubes algorithm.

\noindent \textbf{Implementation details.} Both surface fusion algorithms follow an iterative style by feeding the current observations to a global TSDF (for depths) or occupancy (for \EFMmodel{}) volume. We use the iso-level of $0$ for TSDF fusion and $0.5$ for occupancy fusion. In surface extraction, we set a minimal number of observations for a voxel to be considered valid, which is $2$ for TSDF fusion and $5$ for occupancy fusion.


\subsection{Surface Error Metrics}
Unlikely the previous evaluations~\cite{sun2021neucon,murez2020atlas} that degenerates the metrics to point-cloud-to-point-cloud distance, we strictly evaluate mesh-to-mesh distance, since having GT meshes is one of the unique values of the proposed datasets. We follow Jensen \etal~\cite{jensen2014large} to compute accuracy and completeness errors, denoted as $Acc$ and $Comp$. Accuracy is the distance of the reconstruction (i.e. the input) to the ground truth (i.e. the reference); completeness is the distance from ground truth to reconstruction. When input and reference are meshes, both accuracy and completeness is computed. When the input is the semi-dense points, only accuracy is computed. 
To compute mesh-to-mesh distance, the implementation first samples a fixed number (=10k) of points on the input mesh, and then computes the distance of these points to the closest face of the reference mesh.
Like previous evaluations, we also compute $Precision@5cm$ ($Prec$) and $Recall@5cm$ ($Recal$) which are the ratio of point-to-mesh distances that are within $5cm$. These metrics are mathematically defined in Table.~\ref{tab:surf_metrics}, where $P$ and $P^{*}$ denote the sampled point sets from the predicted and GT mesh. $T$ and $T^{*}$ denote the triangles for the predicted and GT meshes. $\textbf{Dist}(p, f)$ is the orthogonal distance from a point $p$ to a triangle face $f$.  
\begin{table}[]
    \centering
    \begin{tabular}{c|c}
       \hline \hline
       Acc  &   $\text{mean}_{p\in P} (\min_{f \in T^{*}} \textbf{Dist}(p, f) )$ \\ \hline 
       Comp     &  $\text{mean}_{p\in P^{*}} (\min_{f \in T} \textbf{Dist}(p, f) )$  \\ \hline           
       Prec     &     $\text{mean}_{p\in P} (\min_{f \in T^{*}} \textbf{Dist}(p, f) < 0.05 )$        \\ \hline
       Recal     &      $\text{mean}_{p\in P^{*}} (\min_{f \in T} \textbf{Dist}(p, f) < 0.05)$         \\
       \hline \hline
    \end{tabular}
    \caption{Surface metric definitions.}
    \label{tab:surf_metrics}
\end{table}

\section{Additional Experiments and Details}

We provide more details for \EVL{} OBB detection performance by drilling into per-class mAPs, show more qualitative experiments on AEO, and illustrate some failure cases for OBB detection. 
Additionally, we provide additional sensitivity studies for surface regression with \EVL{} and show failure modes of Gaussian Splats for surface estimation on egocentric data. The later justifies why we did not compare against Gaussian Splats.


\subsection{EVL per-class Object Detection Performance} 
\label{supp:per_class_perf}
In Table~\ref{tab:obj_det_per_class_aeo} we show per-class mAPs for the four models on \AOThree{} and in Table~\ref{tab:obj_det_per_class_ase} on the \ASE{} validation dataset. Unlike on \ASE{}, no model does well on the Mirror, Lamp or Plant/FlowerPot class on \AOThree{}  which is likely due to substantial sim-to-real gap in the class examples in the ASE training dataset. For the Lamp and Plant/FlowerPot class the typical position on top of other objects in the real world may also contribute to the sim-to-real gap since in \ASE{} all objects are placed on the floor. 
Both ImVoxelNet and 3DETR struggle with the WallArt/Picture and Window classes which \EVL{} performs well on.
Larger, relatively well defined objects such as Bed, Chair, Sofa/Couch and Table are detected reasonably well by all detectors.

\begin{table}[]
    \centering
    \begin{tabular}{c| c|c|c|c| c|c|c|c| c|c|c|c|}
     &  Bed & Chair & Couch & Lamp & Mirror & Plant & Table & WallArt & Window \\ \hline
        Cube R-CNN    & 0.32 & 0.08 & 0.33 & 0.00 & 0.00 & 0.00 & 0.02 & 0.00 & 0.00 \\ 
        ImVoxelNet    & 0.49 & 0.22 & 0.46 & 0.05 & 0.00 & 0.00 & 0.08 & 0.01 & 0.05 \\ 
        3DETR         & 0.24 & 0.23 & 0.66 & 0.03 & 0.00 & 0.00 & 0.10 & 0.06 & 0.07 \\
        \EVL{} (ours) & 0.46 & 0.25 & 0.65 & 0.04 & 0.01 & 0.00 & 0.14 & 0.21 & 0.18 \\  
    \end{tabular}
    \caption{Four models trained on \ASE{} are evaluated the real-world \AOThree{} dataset. }
    \label{tab:obj_det_per_class_aeo}
\end{table}

\begin{table}[]
    \centering
    \begin{tabular}{c| c|c|c|c| c|c|c|c| c|c|c|c|}
     &  Bed & Chair & FlowerPot & Lamp & Mirror & Picture & Sofa & Table & Window \\ \hline
        Cube R-CNN    & 0.78 & 0.62 & 0.32 & 0.37 & 0.04 & 0.02 & 0.71 & 0.54 & 0.19 \\ 
        ImVoxelNet    & 0.94 & 0.92 & 0.68 & 0.83 & 0.19 & 0.11 & 0.95 & 0.91 & 0.61 \\ 
        3DETR         & 0.90 & 0.78 & 0.18 & 0.34 & 0.10 & 0.13 & 0.87 & 0.62 & 0.36 \\
        \EVL{} (ours) & 0.95 & 0.91 & 0.77 & 0.86 & 0.44 & 0.37 & 0.94 & 0.88 & 0.73 \\  
    \end{tabular}
    \caption{Four models trained on \ASE{} are evaluated on the \ASE{} validation dataset. }
    \label{tab:obj_det_per_class_ase}
\end{table}

\subsection{\EVL{} Object Detection Scaling Experiments}
\label{sup:scaling}

What training dataset size is big enough to saturate the current \EVL{} model? 
We train \EVL{} on increasingly large subsets of the ASE training dataset and measure detection performance on the same validation set. We find that \EVL{}s performance saturates at 10k sequences.
This is the foundation of our choice to train on the 600k 1s snippets of a subset of 10k sequences from ASE.
We expect that transformer-based methods would exhibit better scaling behavior and leave this investigation for future work.


\begin{figure}
    \centering
    \includegraphics[width=0.65\textwidth,trim=0 0 0 10,clip]{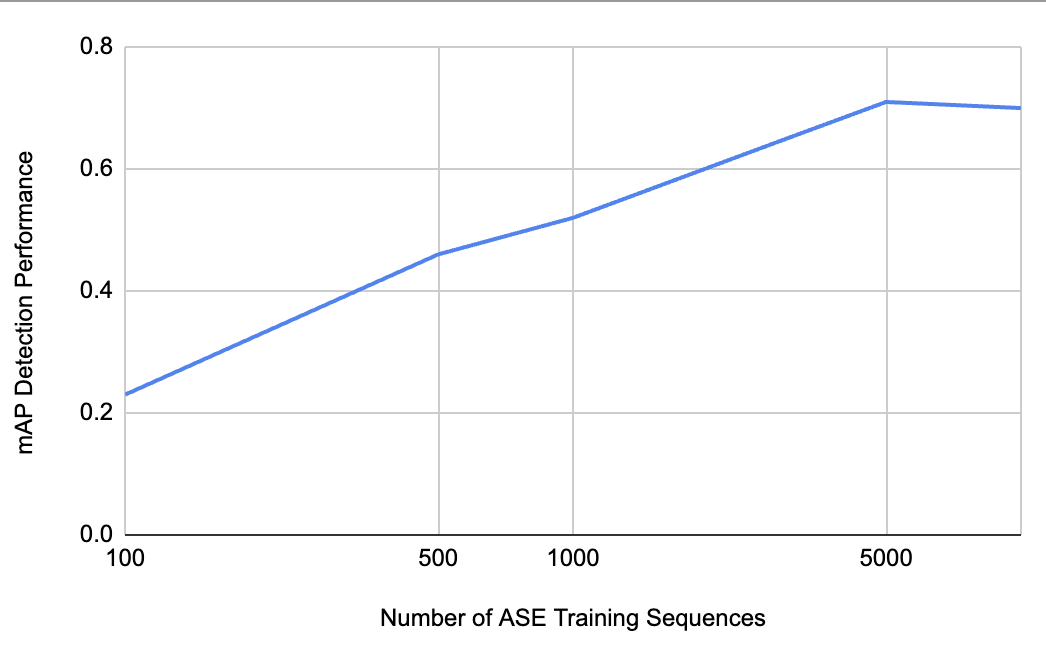}
    \caption{\EVL{} performance on the same validation set saturates at 10k ASE sequences (the last data point on the log-scaled x-axis).
    }
\end{figure}





\subsection{EVL Failure Cases and Limitations}

EVL has only been trained on the simulated ASE training dataset and therefore has only seen a limited set of scene variations. Particularly it has only seen indoor scenes with all objects located on the ground.

\begin{figure}
    \centering
    \includegraphics[width=\textwidth]{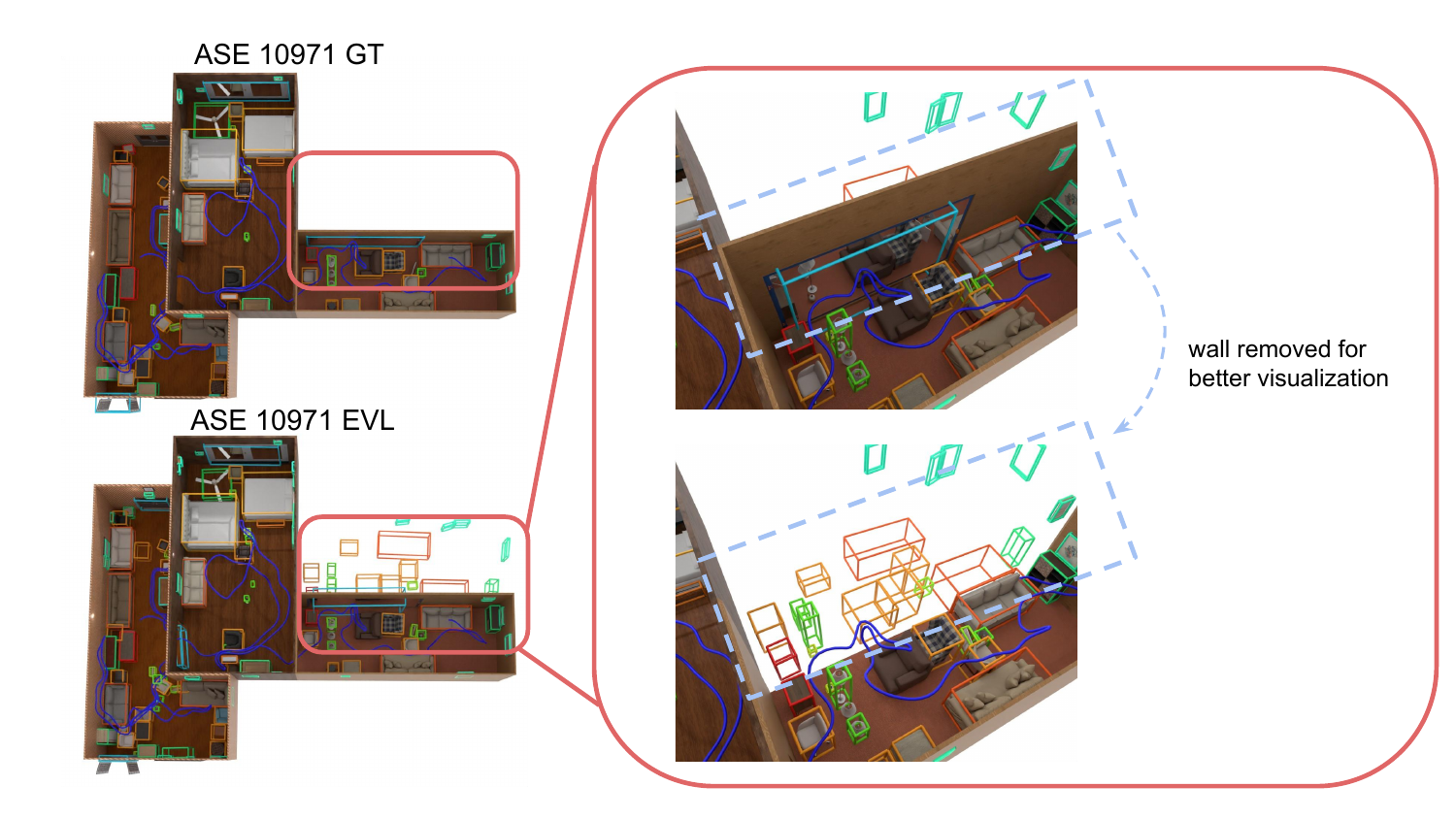}
    \caption{\EVL{} fails to reason about the reflections and predicts false-positives on the reflected image from the mirror on the wall. 
    \label{fig:sup_mirror}
    }
\end{figure}

\begin{figure}
    \centering
    \includegraphics[width=\textwidth]{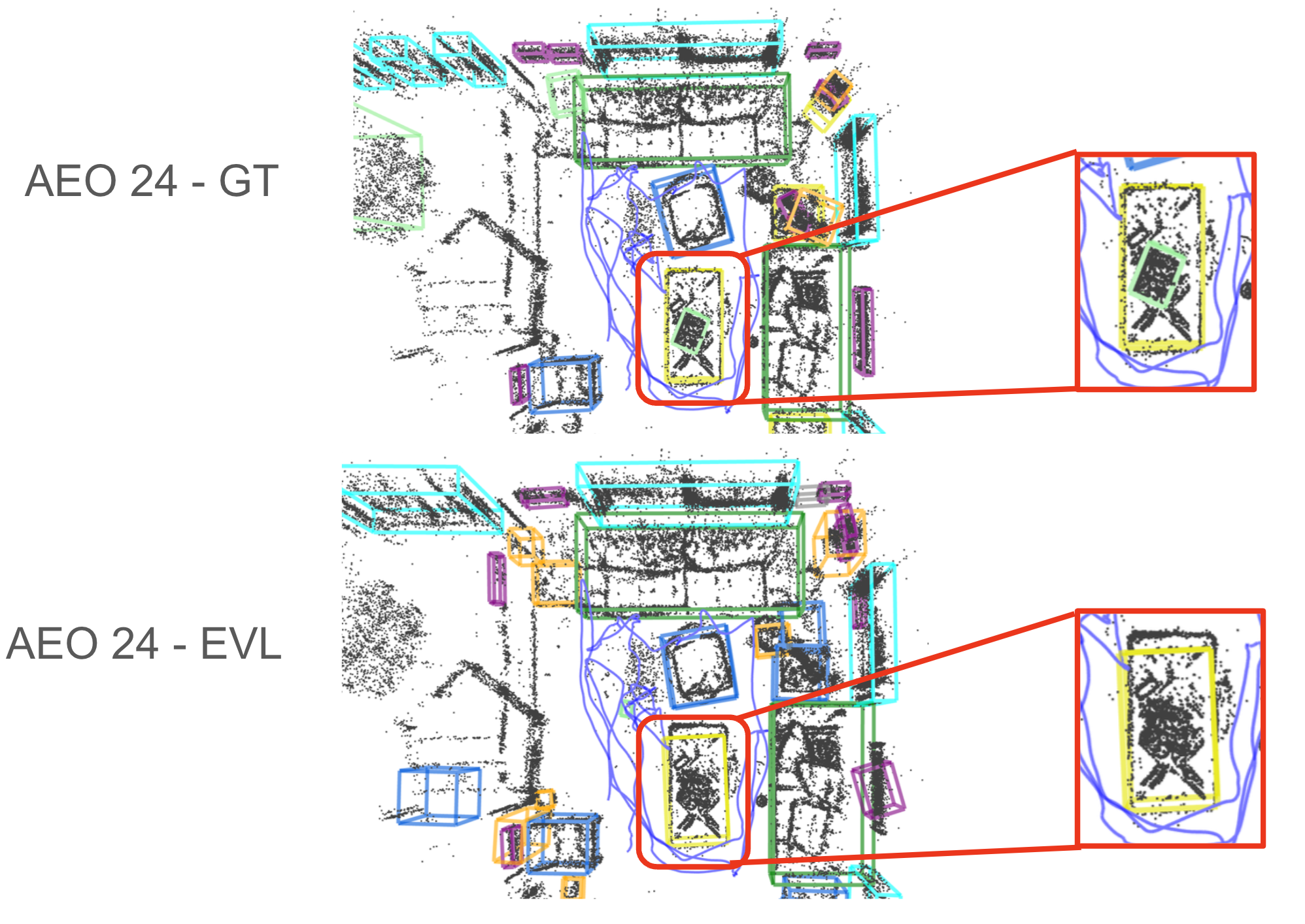}
    \caption{\EVL{} fails to detect the plant on the table (shown in the green OBB). This is due to a limitation in the training data where there are no objects on top of on another in \ASE{}.
    \label{fig:sup_object_on_top}
    }
\end{figure}

\noindent\textbf{Failure cases.}
In Fig.~\ref{fig:sup_object_on_top} we show that EVL has problems detecting objects on top of other objects. In this case the plant object on the table is not detected. This domain gap in relative object poses contributes to the substantially lower mAPs seen in objects that are typically not located on the floor (see~Sec.~\ref{supp:per_class_perf}).
In Fig.~\ref{fig:sup_mirror} we show that the presence of mirrors in the ASE dataset leads to erroneous reflected detections outside the room behind the mirrors. This shows that the model has not yet learned to correctly reason about reflections via mirrors.

\begin{figure}
    \centering
    \includegraphics[width=0.45\textwidth]{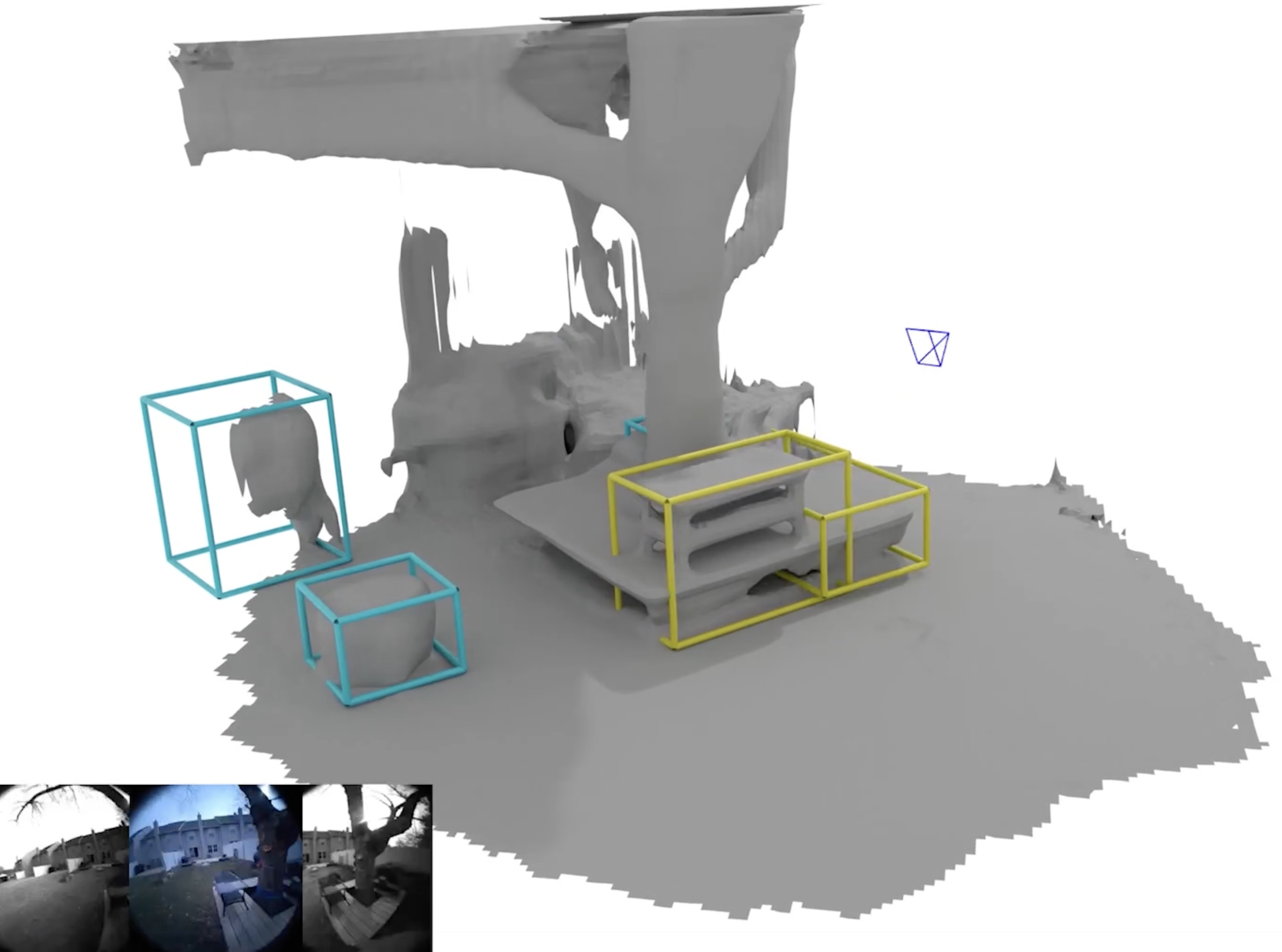}
    \includegraphics[width=0.45\textwidth]{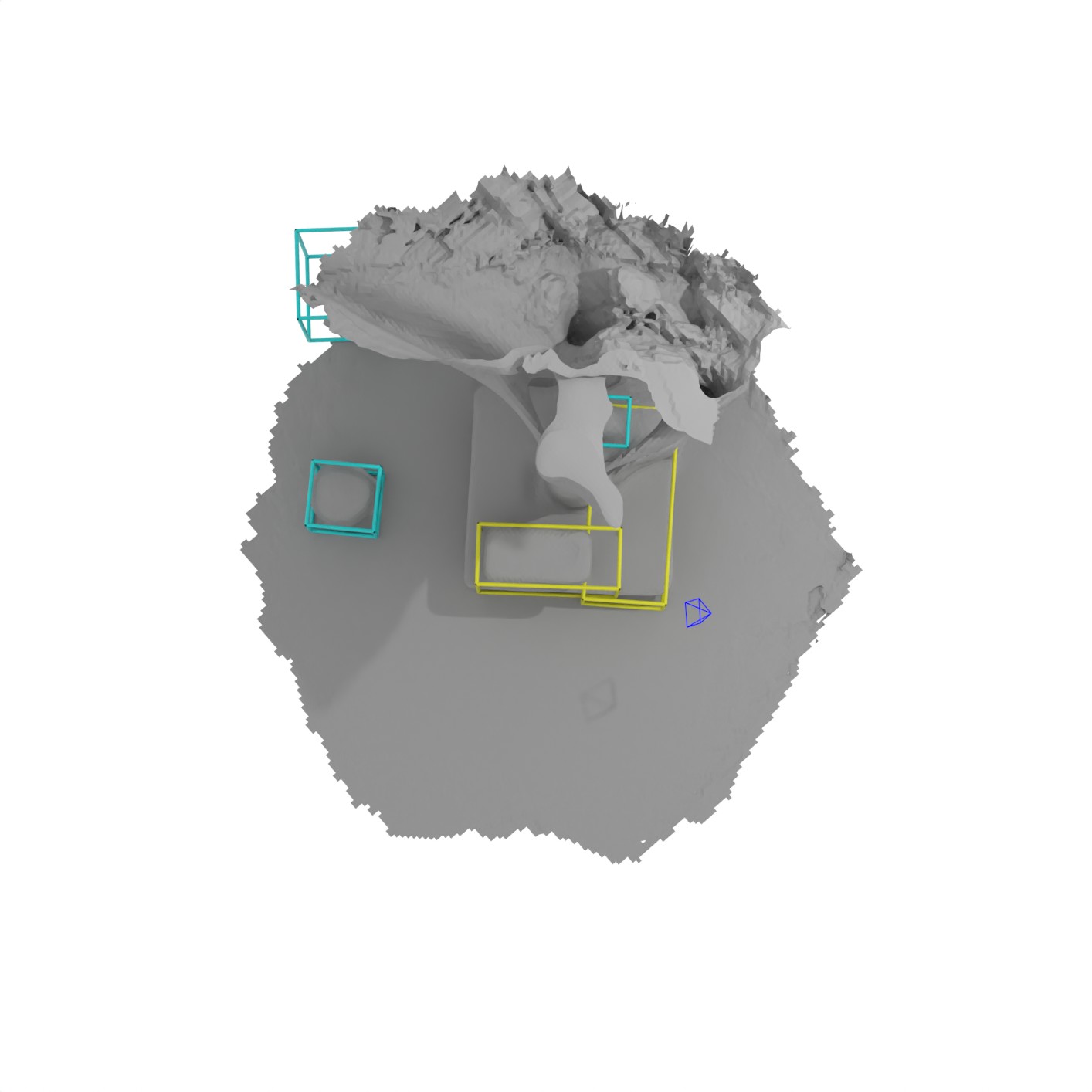}
    \caption{\EVL{} was trained only on indoor data from \ASE{} yet generalizes reasonably well on outdoor real Project Aria data. However, the limited FoV and lack of detail on tree branches remain problematic, and the 3D viewing frustum ($4m \times 4m \times 4m$) limits the range of reconstructions that \EVL{} can deliver. 
    \label{fig:evl_outdoor}
    }
\end{figure}

\noindent\textbf{Limitations.}
We show EVL running on an outdoor sequence from AEO in Fig.~\ref{fig:evl_outdoor} . 
As noted before EVL has only seen indoor scenes during training which means that trees and open sky are completely out of the training distribution. 
EVL reconstruction quality degrades gracefully on this sequence by still predicting smooth surfaces for the tree, table and grass, but struggles with the fine detail of the tree branches. Another drawback is that the EVL model has a limited 3D viewing frustum ($4m \times 4m \times 4m$ by default), which cannot process distant scene geometry which is more common outdoors.

\subsection{EVL Sensitivity Study for Surface Estimation}

\noindent\textbf{Camera Model Sensitivity:}
We test \EVL{} surface regression when changing the camera model. Specifically, we rectify the original Fisheye camera model to linear camera models and use the rectified images as the inputs to the \EVL{} model.
As shown in Table~\ref{tab:surf_cam_ablation}, the performance does not degrade substantially with different camera models. Camera models with smaller field of view like the ScanNet-linear camera model do lead to the largest degradation. Insensitivity to the particular camera model is a valuable property since it theoretically allows training the same model on different kinds of cameras. Likely the semi-dense points inputs are important in keeping performance relatively high independent of the camera model.
\begin{table}[]
    \centering
    \begin{tabular}{c|c|c|c|c}
         Camera Model & ASE Prec $\uparrow$ & ASE Recal $\uparrow$ & ADT Prec $\uparrow$ & ADT Recal $\uparrow$ \\ \hline \hline
         Fisheye & 0.73 & 0.35 & 0.44 & 0.077 \\ \hline
         Max. Linear & 0.73 & 0.35 & 0.42 & 0.076 \\ \hline
         ScanNet Linear & 0.69 & 0.33 & 0.41 & 0.070 \\ \hline
        
    \end{tabular}
    \caption{\EVL{} Surface regression performance when changing the camera model.}
    \label{tab:surf_cam_ablation}
\end{table}

\noindent\textbf{Ablation Studies:} We show the ablation studies of \EVL{} on Surface regression. Specifically, in Table~\ref{tab:surf_ablation} we show the effectiveness of using the points mask and the freespace mask. We can see that using point masks from the semi-dense points gives a large boost on the performance and by modeling the freespace it gets further improved.

\begin{table}[]
    \centering
    \begin{tabular}{c|c|c|c|c|c}
         points & freespace & ASE Prec $\uparrow$ & ASE Recal $\uparrow$ & ADT Prec $\uparrow$ & ADT Recal $\uparrow$ \\ \hline \hline
         \checkmark & \checkmark & 0.82 & 0.41 & 0.59 & 0.11\\ \hline
        \checkmark & & 0.73 & 0.35 & 0.44 & 0.077\\ \hline
             & & 0.47 & 0.18 & 0.14 & 0.023 \\ 
    \end{tabular}
    \caption{\EVL{} Surface regression ablation study. Adding point occupancy and freespace mask improves performance. }
    \label{tab:surf_ablation}
\end{table}

\subsection{Gaussian Splats Experiments}
We evaluated 3D Gaussian Splatting \cite{kerbl3Dgaussians} as a surface estimator  using NerfStudio's Splatfacto model \cite{nerfstudio}. The images from the RGB stream were undistorted from fisheye to linear cameras using the Max. Linear camera model before training and the Gaussian point cloud is optimized for 30K iterations with the default settings. 

\begin{figure}[htbp]
    \centering
    \begin{minipage}{\textwidth}
        \centering
        \includegraphics[width=0.75\textwidth]{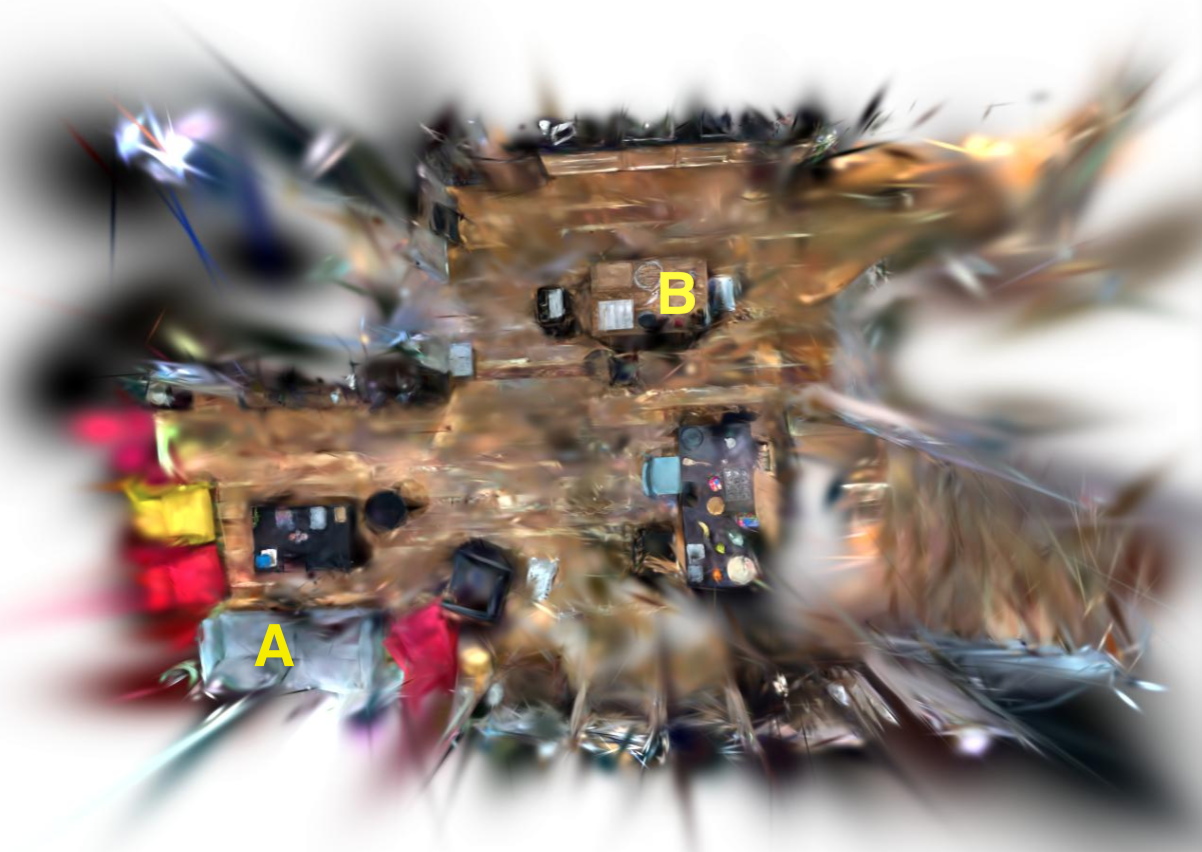} 
        \caption{Top-down view of ADT-136 "work" reconstructed using 3D Gaussian Splatting \cite{kerbl3Dgaussians}. Egocentric data contains sparse, dynamic observations of the user throughout the scene. The inability for 3D Gaussians to model these sparse dynamics leads to "fog" throughout the scene placed exactly at the user's trajectory\label{fig:gaussian_adt}. This "fog" degrades the accuracy of the rendered depth from Gaussian Splatting.}
    \end{minipage}
\end{figure}

\begin{figure}[htbp]
    \begin{minipage}{\textwidth}
        \centering
        \begin{subfigure}{0.32\textwidth}
            \includegraphics[angle=-90,width=\textwidth]{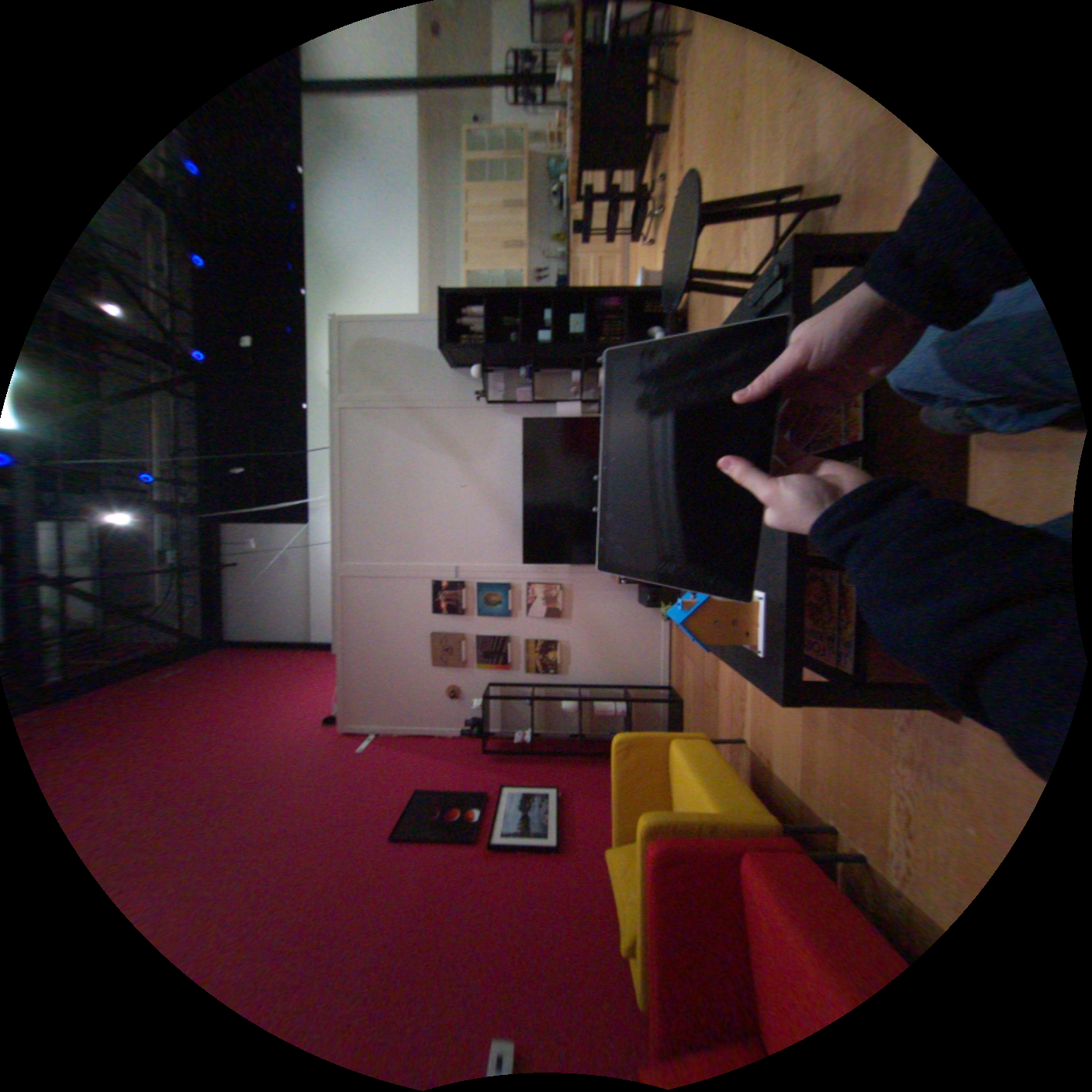}
        \end{subfigure}
        \hfill 
        \begin{subfigure}{0.32\textwidth}
            \includegraphics[angle=-90,width=\textwidth]{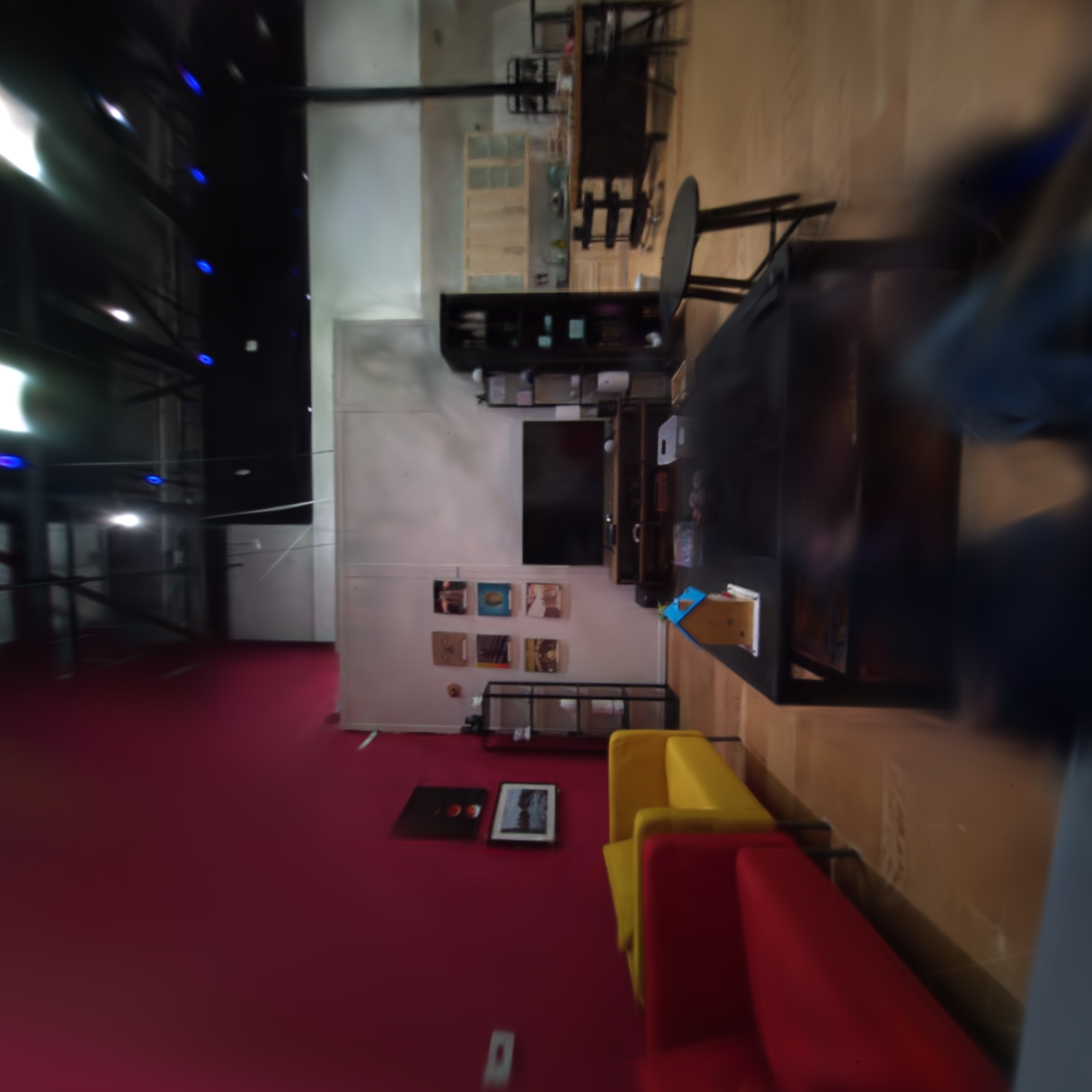}            
        \end{subfigure}
        \hfill 
        \begin{subfigure}{0.32\textwidth}
            \includegraphics[angle=-90,width=\textwidth]{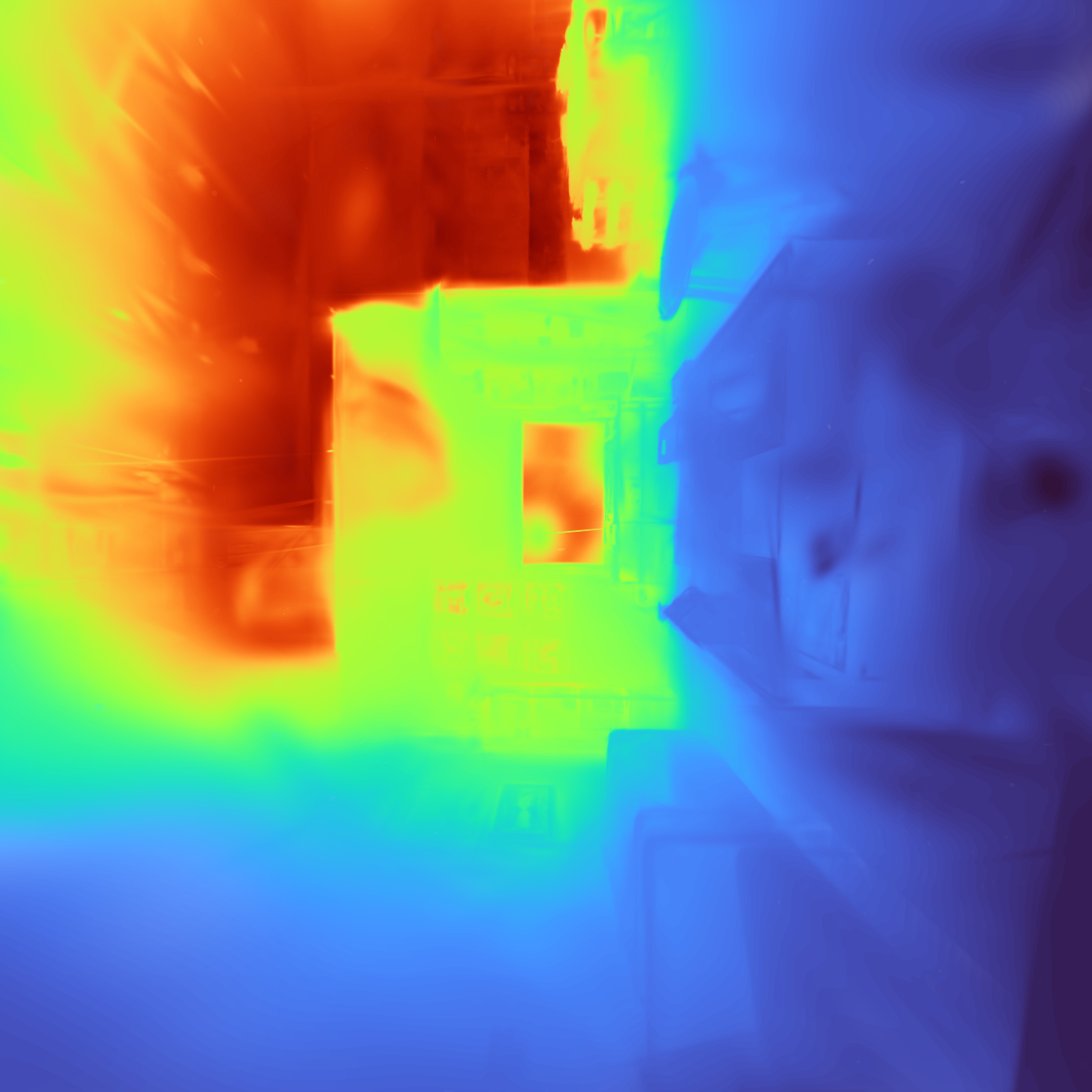}           
        \end{subfigure}

       \begin{subfigure}{0.32\textwidth}
            \includegraphics[angle=-90,width=\textwidth]{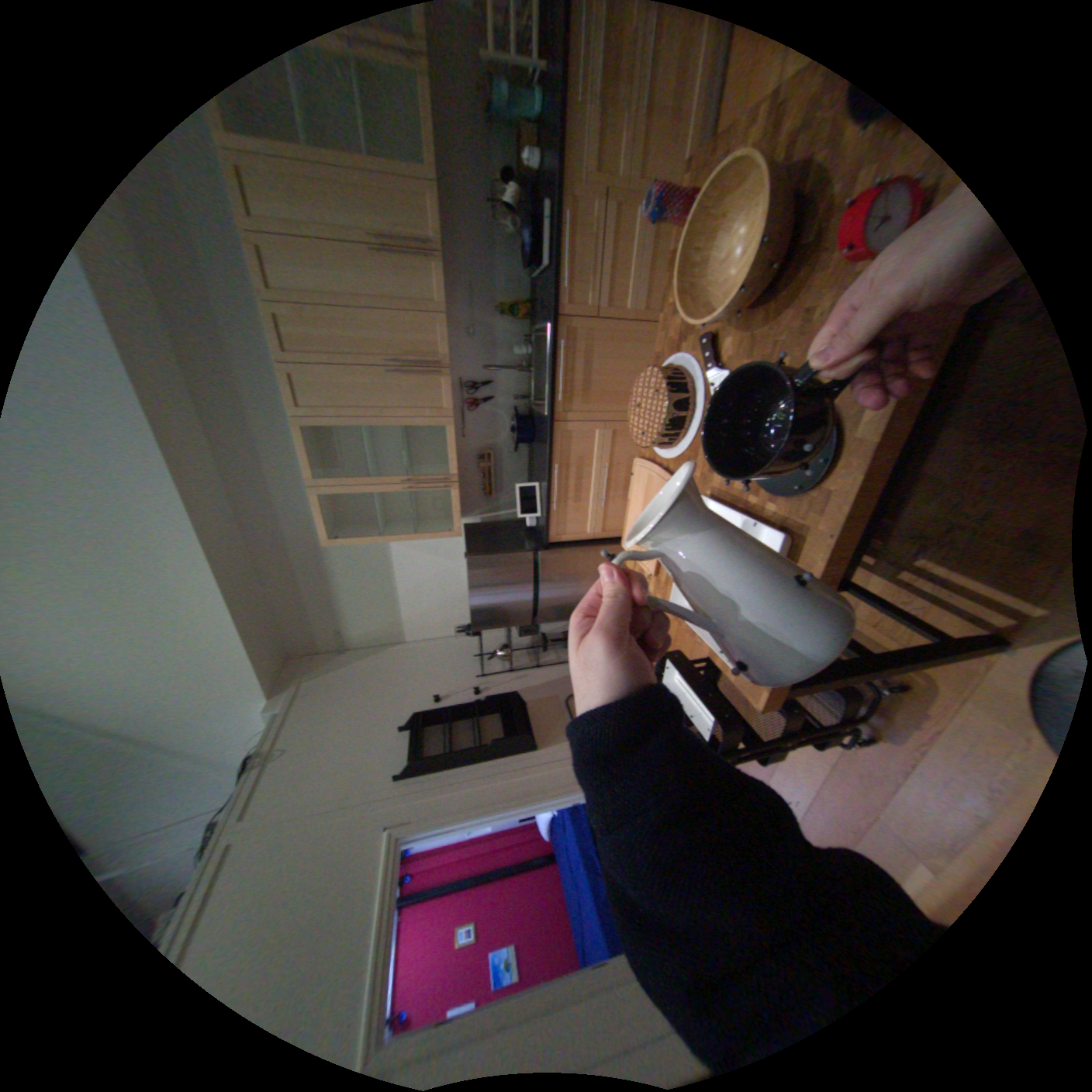}
        \end{subfigure} 
        \hfill 
        \begin{subfigure}{0.32\textwidth}
            \includegraphics[angle=-90,width=\textwidth]{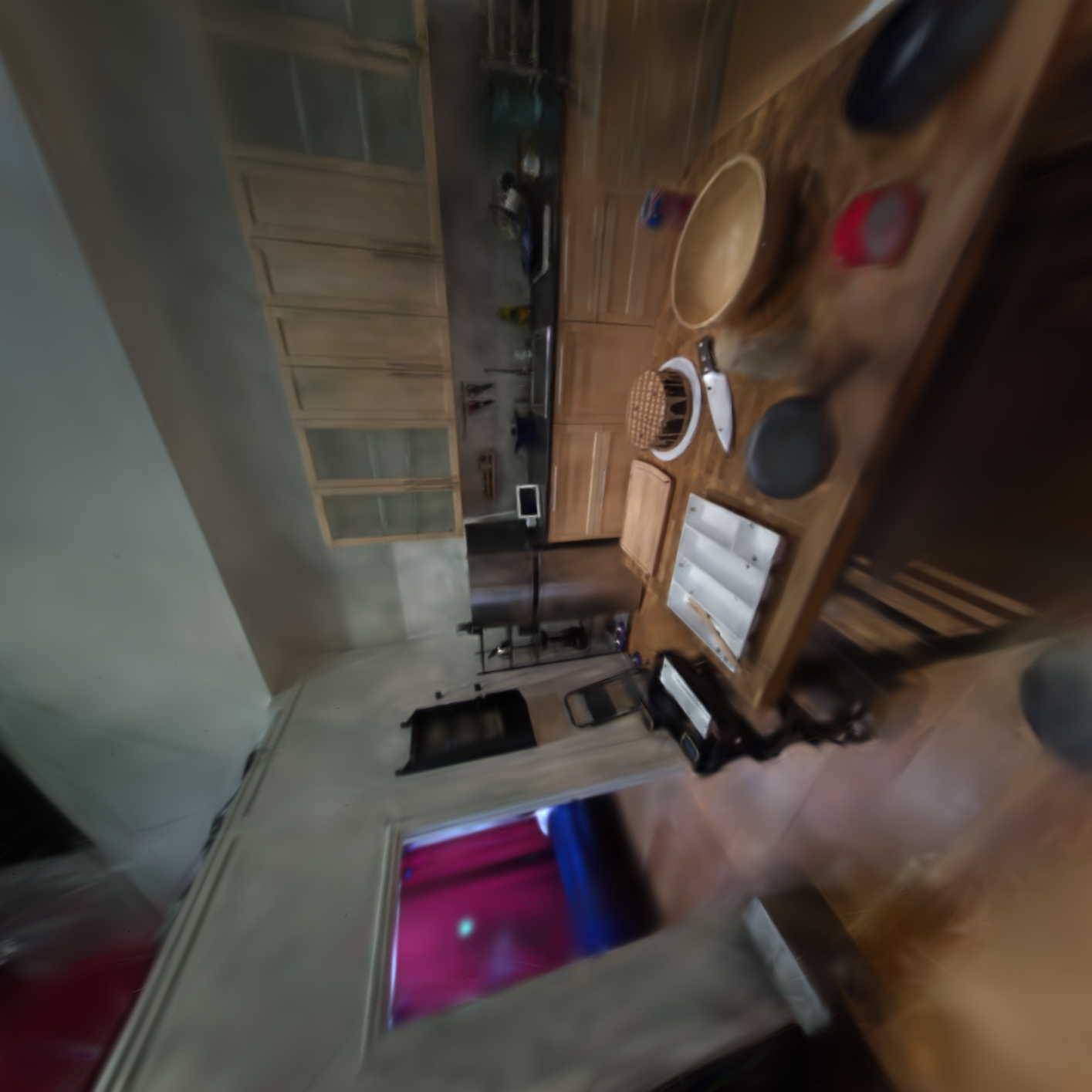}            
        \end{subfigure}      
        \hfill 
        \begin{subfigure}{0.32\textwidth}
            \includegraphics[angle=-90,width=\textwidth]{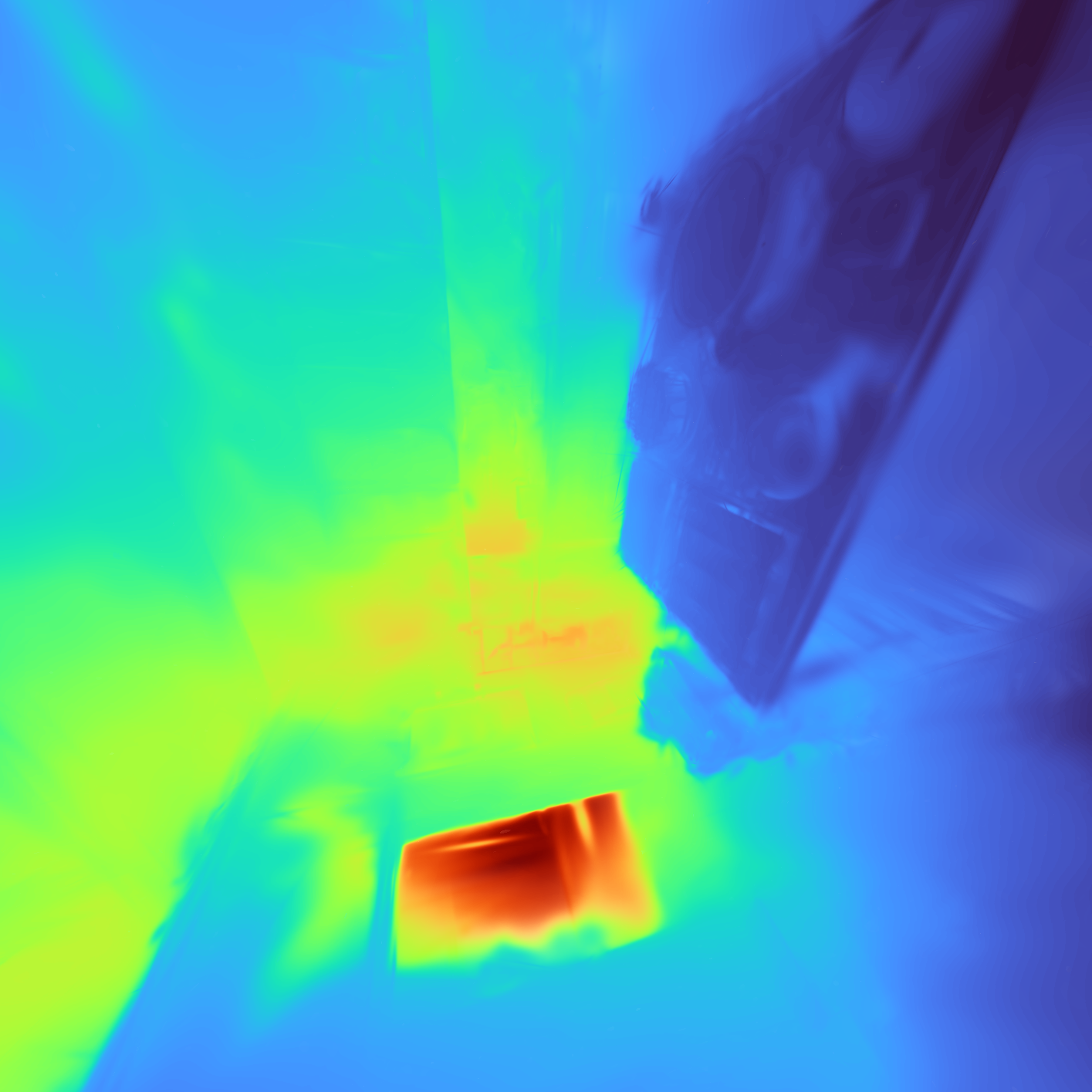}           
        \end{subfigure}

            \caption{\textbf{Left:} Two video frames from ADT-136 "work" containing observations of the user dynamically interacting with the scene. \textbf{Center:} The rendered RGB image of the same viewpoint using the optimized 3D Gaussian point cloud. \textbf{Right:} The rendered depth maps using the optimized 3D Gaussian point cloud.\label{fig:gaussian_breakdown}}
    \end{minipage}
\end{figure}

The rendered results of the optimized 3D Gaussian point cloud are shown in Fig.~\ref{fig:gaussian_adt}. Notice the inconsistent floor texture with many areas of dark patches or fog. These areas exactly correspond to the areas where the device-wearer moved while capturing the video. When the video used to train the 3D Gaussian point cloud contains observations of the wearer (as shown in Fig.~\ref{fig:gaussian_breakdown}) the model will attempt to create new Gaussians at the along the wearer's trajectory to explain those observations (incorrectly) as static Gaussians. This is particularly a problem when computing the depth via rendering because these artifacts create false surfaces or otherwise degrade the depth value output by the rasterization process (see Fig.~\ref{fig:gaussian_breakdown} right). The ADT dataset sequences contain significant dynamics of this nature, and so we did not evaluate Gaussian Splatting results for the surface estimation task. We expect that this rapidly evolving area will provide significant opportunity for future EFM algorithms.


\end{document}